\documentclass{article}

\usepackage{PRIMEarxiv}
\usepackage{placeins}  
\usepackage[utf8]{inputenc} 
\usepackage{float} 
\usepackage[T1]{fontenc}    
\usepackage{hyperref}       
\usepackage{url}            
\usepackage{booktabs}       
\usepackage{amsfonts}       
\usepackage{nicefrac}       
\usepackage{microtype}      
\usepackage{lipsum}
\usepackage{fancyhdr}       
\usepackage{graphicx}       
\graphicspath{{media/}}     
\usepackage{multirow}
\usepackage{gensymb}
\usepackage{amsmath}       
\usepackage{algorithm}     
\usepackage{algpseudocode} 
\usepackage{float}  
\usepackage{threeparttable}
\usepackage{caption}
\usepackage{tcolorbox}
\usepackage{authblk}
\usepackage{comment}           
\usepackage{natbib}
\usepackage{amsmath}

\title{Low-Latency Activation-Regularized Sparse Neural Operators with Distillation Assistance Towards Real-Time Neuromorphic Virtual Sensing}

\author[1]{William Howes}
\author[1]{Farid Ahmed}
\author[1,2]{Syed Bahauddin Alam\textsuperscript{*}}

\affil[1]{Grainger College of Engineering, Nuclear, Plasma \& Radiological Engineering Department, University of Illinois Urbana-Champaign, Urbana, IL, USA
}

\affil[2]{National Center for Supercomputing Applications, Urbana, IL, USA
}

\affil[*]{Corresponding author: \href{mailto:alams@illinois.edu}{alams@illinois.edu}}

\begin{document}
\maketitle
\begin{abstract}
Virtual sensing enables digital twins and safety-critical systems to reconstruct and forecast spatial-temporal physics in real time. However, conventional computational and data-driven methods often face challenges in generalization, latency, and energy efficiency for edge deployment. Neural operators offer a promising alternative but remain reliant on power-intensive hardware. Spiking neurons and neuromorphic computing can improve efficiency, yet surrogate-gradient training and multi-step spiking introduce convergence and latency challenges. We propose the Sparse-Activation-ReLU (SAR) layer, a single-step alternative that promotes activation sparsity without surrogate-gradient training while remaining compatible with event-based computing. Within a trunk-based NOMAD architecture, SAR achieves over a fivefold improvement in the combined Latency-Error-Energy (LEE) metric compared with Variable Spiking Neuron (VSN) and Leaky Integrate-and-Fire (LIF) implementations. We further analyze spiking entropy and feature usage and introduce synthetic knowledge distillation, reducing the LEE score by more than twofold. Finally, we improve VSN through a ReLU-based spiking loss and graph-neighbor thresholding. On the Heat Exchanger dataset, these approaches reduce L2 error by more than twofold and nearly sevenfold, respectively, while reducing spiking and spatial aggregation. Overall, the work presented is a step towards energy-efficient virtual sensing by providing an alternative framework that can be positioned towards neuromorphic or other edge device integration that can be a gold standard to compare latency, energy, and error performance for future efficient designs that are sparsity or brain-inspired spiking based.

\end{abstract}

\vspace{-1mm}
\section{Introduction}

With the rapid advancement of artificial intelligence, the unprecedented computational demands of modern datacenter infrastructure have significantly increased global electricity consumption, renewing interest in nuclear energy as a reliable, carbon-free, and scalable power source capable of supporting next-generation computing technologies \cite{epri2024powering, chen2025datacentres}. In particular, Small Modular Reactors (SMRs) and microreactors provide flexible, modular deployment strategies that enable efficient integration of nuclear power into distributed applications, including military installations, remote communities, industrial facilities, and other high-power environments where conventional large-scale nuclear reactors are impractical \cite{testoni2021review}. As these reactor systems continue to decrease in physical scale while increasing in deployment flexibility, there is a corresponding need to improve monitoring, control, and operational awareness to reduce outage time, enhance safety margins, and improve overall economic viability. Virtual sensing, specifically the real-time spatial-temporal reconstruction of governing physical fields through computational methods, offers a pathway toward achieving these goals by providing insight into system states that are either impossible or prohibitively expensive to measure directly. By mapping sparse physical boundary measurements, such as inlet flow sensors, to dense spatial-temporal fields including transient fluid flow, temperature, and pressure distributions, virtual sensing enables enhanced material degradation assessment, anomaly detection, predictive maintenance, and operational decision-making throughout the lifetime of a nuclear system. Existing approaches are primarily based on traditional numerical methods, such as the Finite Element Method (FEM), or machine learning frameworks including Physics-Informed Neural Networks (PINNs) \cite{NASER2026111704}. While these methodologies have demonstrated success across a variety of scientific applications, they often either struggle to satisfy the strict real-time requirements of online monitoring due to their computational complexity, limited ability to generalize to unseen operating conditions, or strict dependence on governing equations that may be unavailable or difficult to formulate for experimental systems \cite{NASER2026111704, kobayashi2024virtualsensingenablerealtime, howes2026realtimesensinginaccessiblephysical}. To overcome these limitations, neural operators have emerged as a powerful class of data-driven architectures capable of learning mappings between infinite-dimensional function spaces without requiring explicit equation integration, thereby enabling rapid inference for previously unseen boundary and initial conditions without retraining \cite{JMLR:v24:21-1524}. Among the earliest and most influential neural operator architectures are DeepONet \cite{Lu2021-ci}, which independently processes boundary conditions and evaluation coordinates through separate branch and trunk networks, and the Fourier Neural Operator (FNO) \cite{DBLP:journals/corr/abs-2010-08895}, which performs iterative spectral convolutions over structured computational grids. More recent neural operator designs have extended these capabilities to increasingly complex multiphysics systems and irregular computational meshes. Despite these significant advances in reconstruction accuracy and generalization, comparatively little attention has been devoted to computational efficiency. Most existing neural operator studies assume the availability of datacenter-class accelerators, such as NVIDIA A100 and H200 GPUs, whereas many practical virtual sensing applications require deployment on resource-constrained edge hardware with strict latency, memory, and power budgets \cite{10786287, 10.1007/s11554-024-01482-0}. Consequently, the development of energy-efficient neural operators capable of maintaining high reconstruction fidelity while substantially reducing computational cost remains an important and largely unexplored research direction.

Spiking Neural Networks (SNNs) \cite{doi:10.1142/S0129065709002002, WU20242909, brainsci12070863, Majumdar2025-vo} have emerged as one of the most promising paradigms for energy-efficient artificial intelligence due to their inherently sparse and event-driven mode of computation. Inspired by biological neural systems, SNNs temporally integrate input and communicate information only when neurons emit spikes, allowing computation to occur selectively rather than through the dense activation patterns characteristic of conventional artificial neural networks. This event-driven spatial-temporal behavior significantly reduces unnecessary computation by naturally exploiting the varying importance of individual neurons during inference. Among the available neuron models, the Leaky Integrate-and-Fire (LIF) neuron has become one of the most widely adopted because it provides an effective balance between biological realism and computational efficiency \cite{10242251}. However, conventional LIF neurons exhibit limited performance on regression-based tasks, including spatial-temporal field reconstruction mainly due to precision loss, motivating the development of the Variable Spiking Neuron (VSN), which handles directly encoded continuous-valued inputs, avoiding precision loss, while producing graded spike outputs on top of the memory/integration dynamics of the previous LIF neuron that preserve the computational advantages of event-driven communication \cite{garg2023neuroscienceinspiredscientificmachinepart2}. Current neural operator implementations incorporating VSNs are almost exclusively trained using surrogate gradients, introducing an inherent mismatch between the forward and backward passes during optimization that frequently leads to unstable training dynamics and degraded reconstruction performance \cite{10.3389/fnins.2026.1795946}. An alternative strategy commonly explored within the neuromorphic community is ANN-to-SNN conversion, although these methods sometimes require long spike time windows and depend upon spike-rate encoding and binary spike communication to accurately represent continuous information \cite{zhang2023artificialspikingneuralnetworks, pmlr-v202-jiang23a}. With the introduction of graded spike outputs in the VSN, which is shown to improve regression performance in neural operator applications, communication shifts away from rate-based coding and back toward the direct transmission of continuous-valued information, making it difficult to map the standard ANN-SNN conversion formulation to the VSN framework. Although VSN-based neural operators have demonstrated encouraging performance for virtual sensing applications \cite{howes2026neuroscienceinspiredgraphoperators, garg2023neuroscienceinspiredscientificmachinepart2, JAIN2025106152}, they remain fundamentally constrained by surrogate-gradient optimization. Increasing the number of spike timesteps can improve reconstruction accuracy, as demonstrated later in this work, but does so at the expense of increased inference latency, greater computational cost, higher overall spiking activity, and consequently increased energy consumption. 

For this paper, we chose to stay within the variable/continuous communication framework to allow for ideal regression performance. In addition, we focus on low-latency performance. The VSN has had initial success in terms of performance and potential efficiency and has even been demonstrated to perform within a single spike step \cite{garg2023neuroscienceinspiredscientificmachinepart2, howes2026neuroscienceinspiredgraphoperators} but this formulation completely removes the intended memory dynamics. With an interest to avoid the unstable surrogate training and stay within the one-step realm and remove the unnecessary memory integration, keeping the low-latency performance, we present a sparse-activation layer within the neural operator framework that allows variable communication similar to VSN and leverages the natural sparsity induced by the ReLU activation function to achieve improved control over the energy-accuracy tradeoff and latency performance. Essentially, our proposed Sparse-Activation-ReLU (SAR) layer applies activation-based sparsity regularization to the sparse output of the ReLU activation function, enabling explicit control over activation sparsity and consequently reducing neuron communication activity below that achieved by existing VSN-based approaches at low-latency while keeping variable communication key for regression neural operator performance and simultaneously maintaining superior reconstruction accuracy through surrogate-free optimization. The SAR layer framework can easily be mapped to an ANN-to-neuromorphic conversion formulation but with its removal of memory dynamics, our proposed work can also be directed towards any hardware/software that can leverage activation sparsity especially with the difficulty of neuromorphic implementation for densely connected neural networks. Another issue, relatively unexplored within spiking operator research which has focused on attaching spiking layers to model architectures, is the feasibility of neuromorphic integration or general edge deployment. To address this concern, we explore a synthetic neural operator distillation framework for edge-deployable virtual sensing that centers around improving the performance of weaker, neuromorphic-friendly or other general hardware portable and activation-sparse operator architectures with synthetic knowledge distillation from sophisticated operator models that are less natural for edge-deployment implementation, providing not only sparse communication and energy-efficiency but low memory/latency requirements. Knowledge distillation has had initial introduction within the neural operator field (\cite{wan2026spectralinspiredoperatorlearninglimitedistill1, Chen2026-nndistill2}), but we introduce the technique, specifically synthetic generation, with the unique perspective of neural operator edge deployment within neuromorphic/sparsity-exploiting hardware, utilizing sophisticated, less device-friendly architectures as teacher models.

Although the SAR layer is not intended to replace conventional spiking dynamics, which may ultimately provide richer feature representations and improved modeling of transient physical phenomena once more effective optimization techniques become available, it serves as a practical low-latency alternative for deployment in strict real-time environments while also establishing a strong benchmark against which future spiking neural operator developments can be evaluated. With this philosophy in mind and the interest of preserving more traditional spiking dynamics or flexibility in the variable spiking signal that is removed in the SAR layer, we also explore activation-ReLU-based regularization in the context of the Variable Spiking Neuron. We explore two opportunities to reduce the latency and improve the performance of the VSN despite its surrogate gradient. We first replace the original percentage based loss term with the activation-based output of the neuron memory before spiking minus the threshold which is fed through a ReLU function, providing a more enriched loss term that explores the strength of neuron spiking event for improved performance without the need for longer spike trains. In addition, we explore a ReLU-based thresholding technique within the gating mechanism of the spatial aggregation layer for the Variable Spiking Graph Neural Operator (VS-GNO) \cite{howes2026neuroscienceinspiredgraphoperators} that performs online filtering of a node's neighbors, essentially removing unnecessary edges from the overall aggregation and improving computational speed and latency. Although the original mechanism adapts the graph and forces gating weights towards zero it ultimately includes all neighbors in the overall computation. As a result, we attempt neighbor thresholding with a separate activation-based loss term to allow for a controllable accuracy-edge count tradeoff, resulting in improved spatial aggregation efficiency.

The contributions of this work are summarized as follows:

\begin{enumerate}
\item \textbf{Surrogate-free variable-signal neural operators towards neuromorphic/sparse sensing.} We introduce the Sparse-Activation-ReLU (SAR) framework, which provides sparse, event-based activation outputs for energy efficiency, rivaling traditional spiking models. The SAR layer exists with a direct ANN-to-neuromorphic conversion methodology that eliminates surrogate-gradient optimization while enabling explicit control over the energy-accuracy tradeoff for neural operator-based virtual sensing. With the absence of memory dynamics, SAR can also be applicable towards non-neuromorphic hardware/software that leverage the activation sparsity and reduce neuron-neuron communication.

\item \textbf{Comprehensive sparsity analysis and regularization term comparison.} We investigate two activation regularization techniques, the $L_1$ norm and Hoyer norm, within sparse neural operators and compare their performance against surrogate-gradient-based VSN and LIF implementations across multiple spike timestep configurations as well introduce a novel entropy-based characterization of sparse feature distributions and propose a unified evaluation metric that jointly balances reconstruction accuracy, inference latency, and energy efficiency.

\item \textbf{Synthetic knowledge distillation for assisting regression performance.} We present an initial synthetic distillation neural operator framework in which a sophisticated graph-based teacher neural operator transfers knowledge to an activation-regularized student model that is more natural to integrate on neuromorphic hardware or other edge devices that leverage sparsity. Although simplified in the present work, this framework establishes a foundation for future synthetic data generation techniques capable of reducing dependence on computationally expensive high-fidelity simulations.

\item \textbf{Activation regularization for traditional variable spiking neurons.} We extend the proposed activation-based sparsity analysis back to variable-spiking neural operators to improve performance without long spike trains and computational latency by replacing the original percentage-based VSN regularization objective with a SAR-inspired Hoyer-based activation loss, demonstrating improved optimization behavior while highlighting the remaining limitations imposed by surrogate-gradient training. We also attempt activation ReLU-based thresholding with VS-GNO's gating mechanism in order to remove unnecessary neighbors and improve the computational efficiencies of spatial layer in the variable spiking graph operator.

\end{enumerate}

Collectively, these contributions advance the development of energy-efficient, edge-deployable neural operators for scientific machine learning. Beyond introducing a surrogate-free reduction in activation communication that can map to neuromorphic, event-based hardware or other sparsity-leveraging device, this work provides new insight into activation-based sparsity as an effective regularization mechanism, establishes novel analysis tools for evaluating sparse neural representations and performance, demonstrates the potential of synthetic knowledge distillation for reducing data requirements with neural operators while bridging the gap between spiking algorithms and realistic hardware integration, and investigates the applicability of activation-based regularization within existing spiking neural operator architectures. Taken together, these developments represent a meaningful step toward practical, low-power virtual sensing systems capable of performing accurate real-time spatial-temporal reconstruction while reducing both computational cost and dependence on large-scale training datasets.

\section{Methods}

\subsection{Variable Spiking Neural Operators}
\label{sec:vsn}

To improve the regression capabilities of conventional LIF-based spiking neural operators, previous work introduced the Variable Spiking Neuron (VSN), a modification of the standard LIF neuron that preserves event-driven communication while transmitting continuous-valued signals more suitable for scientific machine learning and operator regression tasks \cite{garg2023neuroscienceinspiredscientificmachinepart2}. Rather than transmitting only a binary spike, the VSN gates a continuous signal using the binary spiking event, allowing information to propagate only when the neuron fires while maintaining higher representational fidelity than traditional LIF neurons. This formulation has previously been integrated into neural operator architectures, resulting in Variable Spiking Neural Operators capable of trading predictive accuracy for communication efficiency through sparse neuron activity \cite{garg2023neuroscienceinspiredscientificmachinepart2, howes2026neuroscienceinspiredgraphoperators}.

Given an input feature vector $z^{(t)}$ at spike time step $t$, the membrane potential evolves according to

\begin{equation}
\begin{gathered}
    M^{(t)} = \beta M^{(t-1)} + z^{(t)} \\
    \tilde{y}^{(t)} = 
    \left\{ 
        \begin{array}{lr} 
            1 ; \quad M^{(t)} \geq \Theta \\
            0 ; \quad M^{(t)} < \Theta 
        \end{array}
    \right\} \quad \text{if } \tilde{y}^{(t)}, M^{(t)} \gets 0\\
    y^{(t)} = \sigma(z^{(t)}\tilde{y}^{(t)}), \quad \text{where } \sigma(0) = 0, 
\end{gathered}    
\label{eq:vsn}
\end{equation}

where $\beta$ denotes the membrane leakage parameter, $\Theta$ represents the neuron threshold, $\tilde{y}^{(t)}$ is the binary spike event. When the membrane potential exceeds the threshold, a spike is generated, the membrane is reset, and the neuron transmits the continuous-valued signal $y^{(t)}=\sigma(z^{(t)})$. Otherwise, no communication occurs. In contrast to conventional LIF neurons, this continuous output formulation can significantly improve regression performance while preserving sparse communication behavior required for neuromorphic implementations \cite{garg2023neuroscienceinspiredscientificmachinepart2}.

To regulate communication activity throughout the network, previous Variable Spiking Neural Operators introduce a spike-percentage regularization term during training. Rather than directly penalizing activation values, this objective minimizes the average fraction of observed spike events to total possible spikes across all neuron layers. For a neural operator containing $L$ spiking layers, the overall optimization objective is written as

\begin{equation}
\mathcal{L}=\alpha\cdot\mathcal{L}_{L2}
+
\gamma \cdot
\sum_{l=1}^{L}
\mathcal{L}_{\mathrm{spike}}^{(l)},
\label{eq:vsn_loss}
\end{equation}

where $\mathcal{L}_{L2}$ denotes the reconstruction error, $\mathcal{L}_{\mathrm{spike}}^{(l)}$ is the average spike fraction measured at spiking layer $l$ (how many spikes occurred over how many spikes could have occurred) over all spike time steps (STS) and the input batch, $\alpha$ controls the reconstruction objective, and $\gamma$ determines the strength of the sparsity fraction-based regularization. $\mathcal{L}_{\mathrm{spike}}^{(l)}$ is essentially calculated by counting the binary output, $\tilde{y}^{(t)}$, at time step $t$. Increasing $\gamma$ encourages lower neuron firing rates, reducing communication and expected energy consumption while generally degrading reconstruction accuracy. This formulation therefore provides a controllable accuracy-efficiency tradeoff that has formed the basis of previous Variable Spiking Neural Operator architectures.

\begin{figure}[htbp]
    \centering
    \includegraphics[width=1.05\textwidth]{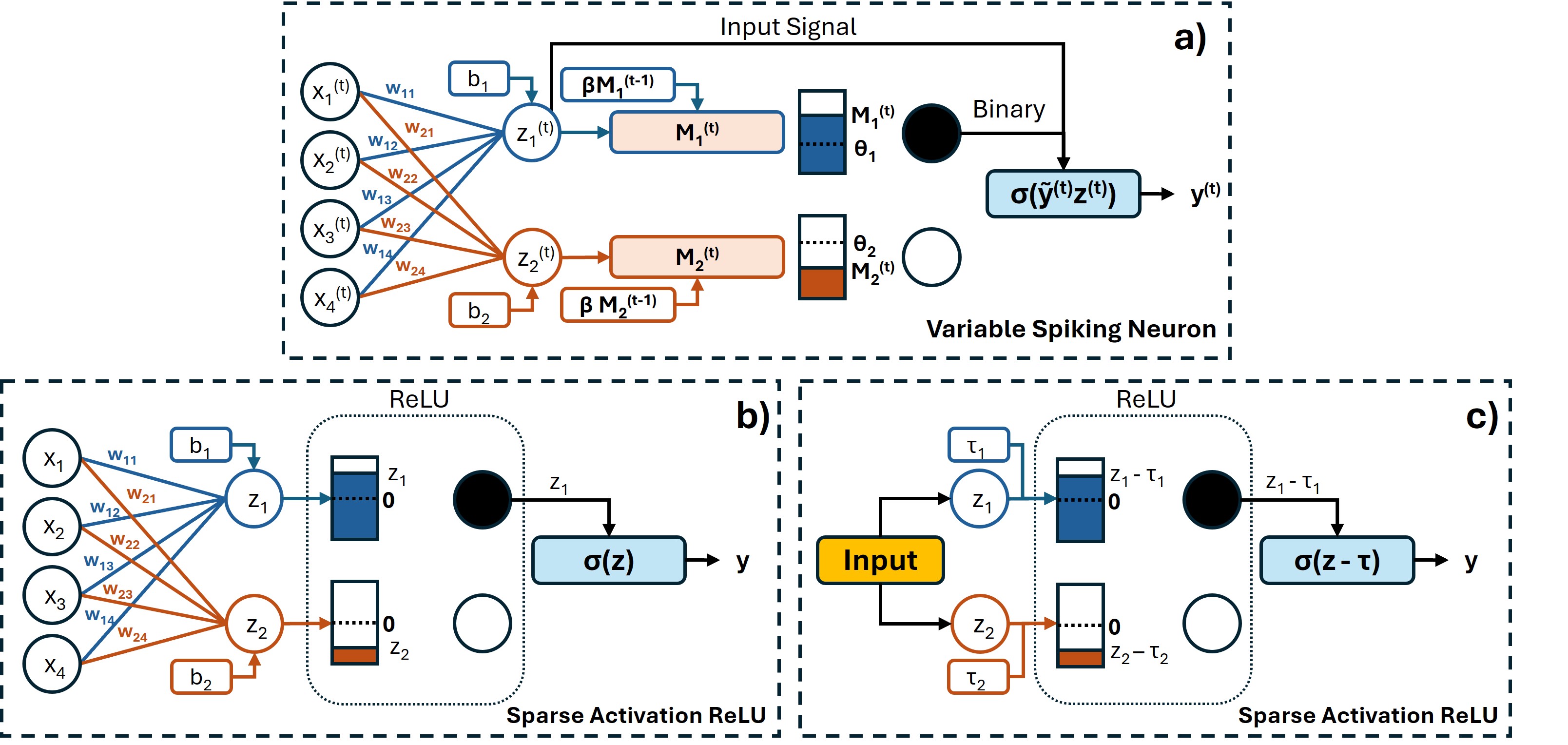}
    \caption{
    \textbf{Sparse-Activation-ReLU Model}
    \textbf{(a)} Visual demonstration of the Variable Spiking Neuron \cite{garg2023neuroscienceinspiredscientificmachinepart2} operating after a linear fully-connected layer. Synaptic weights and unique bias terms accumulate the input for each neuron at spike time step $t$ which is aggregated into the neuron's leaky memory, mirroring LIF dynamics. VSN replaces the binary spike output with an activation function operating on the neuron input.
    \textbf{(b)} Sparse-Activation-ReLU for a fully-connected linear layer. We apply a ReLU activation function to the linear layer output which only allows positive signals subsequently operated on by a chosen activation function. Such forward pass communication allows for surrogate-free ANN-to-neuromorphic conversion that allows for variable signals (unlike traditional binary spike rate to activation matching), included in the original VSN framework.
    \textbf{(c)} Sparse-Activation-ReLU for a generic input (spectral/spatial convolution or norm). We apply a ReLU activation function to the input subtracted by an optional threshold utilized for better spiking control. Similar to the linear layers, the ReLU only allows positive signals subsequently operated on by a chosen activation function.
    }
    \label{fig:sar_model}
\end{figure}

\subsection{Sparse-Activation-ReLU Alternative}
\label{sec:sparsification}
The Variable Spiking Neuron (VSN) was originally introduced as a sparse communication mechanism with variable communication that regulates information flow through thresholded spiking behavior that is controlled by a spiking percentage loss term. The VSN allows for improved regression performance compared to the binary LIF neuron \cite{garg2023neuroscienceinspiredscientificmachinepart2}. Previous attempts to implement VSN into operator architectures typically explore error and spiking behavior for a single spike time step (1 STS) \cite{howes2026neuroscienceinspiredgraphoperators, garg2023neuroscienceinspiredscientificmachinepart2}. Increasing spiking time steps adds increased memory and computational burden to training and risks increasing inference latency as well as energy consumption which can go against the real-time, edge constraints that dominate virtual sensing \cite{garg2023neuroscienceinspiredscientificmachinepart2}. As a result, it seems prudent, when exploring regression-based spiking neural operators for real-time virtual sensing, to reduce spiking time to one step. However, as a consequence, when operating in a single spike time step (1 STS), the temporal dynamics that traditionally motivate spiking neural networks no longer exist. Since only a single spike evaluation is performed, there is no meaningful membrane memory to maintain and the leakage dynamics become irrelevant. Consequently, the neuron functions primarily as a filtering operation, selectively allowing information to propagate when an activation exceeds a learned threshold. This transition in framework leaves the VSN design as underutilized. While a 1 STS VSN has shown promising effectiveness and improvement over LIF-based neural operators \cite{garg2023neuroscienceinspiredscientificmachinepart2}, training the VSN still requires surrogate-gradient approximations to back-propagate through the non-differentiable spike function. This introduces a mismatch between the forward and backward passes and can complicate optimization \cite{10.3389/fnins.2026.1795946}. Traditional ANN-to-SNN methods tend to have long latencies and even alternatives that shorten latency requirements \cite{pmlr-v202-jiang23a} are still based on binary communication, and for this paper we avoid such binary signaling and chose graded/variable spiking due to the previous shown improved regression performance that avoids precision loss. Motivated by the observation that the 1 STS VSN acts primarily as a sparsity-inducing filter, we replace the spiking neuron with a differentiable ReLU-based activation sparsity layer that can be trained directly using conventional gradient descent and runs in only a single step with variable communication. This layer does not rely on 
percentages but the direct sparsity-norm calculation of output activations of the ReLU layer.

Given an input feature vector $\mathbf{z}$ which is typically the result of a particular computational layer within the neural operator, the proposed Sparse-Activation-ReLU (SAR) Layer is simply defined as

\begin{equation}
\mathbf{y}=\sigma(\mathrm{ReLU}(\mathbf{z}-\boldsymbol{\tau})), \quad \text{where } \sigma(0) = 0.
\label{eq:relu_sparse}
\end{equation}

where $\boldsymbol{\tau}$ is an optional learnable subtraction parameter that can provide enriched learning dynamics towards sparse activations especially when $\mathbf{z}$ is not from a linear layer with a built in bias parameter. For any linear layer input, $\mathbf{z}=W\mathbf{x}+\mathbf{b}$, the included bias term acts as our thresholding, meaning we do not include $\boldsymbol{\tau}$ in our calculation, but when applying SAR to spectral-convolution, spatial aggregation layers, or after normalization layers without a clear bias term, we utilize $\boldsymbol{\tau}$ as a learnable filtering term analogous to the thresholding behavior of the original VSN to allow for better control over spiking and efficiency. The result of the ReLU function can then be fed into a subsequent activation function $\sigma$ such that $\sigma(0)=0$, allowing for more enhanced nonlinear computations if desired (e.g. softmax in transformer) that does not add to the efficiency of the spiking framework of VSN. This is where the variable communication seen in the VSN is partially preserved differing from traditional SNN methods that rely on binary communication. The resulting variable activations $\mathbf{y}$, which preserve the variable communication framework of VSN for improved performance, contain a level of sparsity (zero-count) due to the definition of the ReLU function and are then passed to the subsequent network layers. Regularization of the sparsity-level of $\mathbf{y}$ can help achieve a controllable and desired sparsity level. 

Previous sparsity inducing methodologies similarly exploit the ReLU function but typically rely on fixing $\boldsymbol{\tau}$ as a predefined value greater than 0, such as shifted ReLU \cite{price2024deepneuralnetworkinitialization}, or manipulating the ReLU dynamics to admit only values above a threshold, post-training, such as FATReLU \cite{pmlr-v119-kurtz20a}. Both methods treat thresholding within the ReLU dynamics as a fixed hyperparameter while SAR aligns closer with the VSN by optimizing the threshold parameter. These methods are also applied towards standard LIF/binary communication and have not been approached towards variable communication such as SAR. Techniques such as shifted ReLU or FATReLU can be applied post training to further induce sparsity but risk severe degradation.

Following training, the SAR layer can be incorporated into the neuromorphic systems with the following procedure: define synapse weights with the multiplicative weights utilized in computational layers within neural operators, combine the existing bias or the optional subtraction parameters if needed ($\mathbf{b}/\boldsymbol{\tau}$), and simply set a zero-threshold spiking neuron that corresponds to the ReLU function designed to operate in one step (no stored memory/leakage). The resulting signal is simply a positive filter with a value, for a specific neuron/dimension $i$, that is either zero (no signal) or $\sigma(z_i-\tau_i)/\sigma(z_i)$ if greater than zero, essentially meaning $\mathbf{y}$ from Equation \ref{eq:relu_sparse} is the signal sent aligning naturally with the original $\sigma$-ReLU-based ANN computation. Figure \ref{fig:sar_model} visualizes the neuromorphic implementation of the SAR layer as well as the Variable Spiking Neuron. It is important to note that because the ReLU is converted to the neuromorphic thresholding, our $\sigma$ function does not add additional computation compared to the VSN preserving the original efficiency characteristics. This provides a simple ANN-to-neuromorphic conversion mechanism without requiring surrogate-gradient optimization during training. Within this framework, activation sparsity becomes directly analogous to spiking activity, allowing standard sparsification techniques to regulate the communication behavior of the network. In addition, the SAR layer operates only in a single step similar to the previously trained 1 STS VSN models. As a result, SAR-based neural operators provide a low latency, efficient alternative that can provide better power consumption and reconstruction performance due to the lack of surrogate gradient training. The summarized dynamics of the SAR spiking layer for neuron/dimension $i$ is below where $y_i=0$ corresponds to no signal sent and no further downstream computation:

\begin{equation}
\begin{gathered}
    y_i = \sigma(z_i-\tau_i) \quad \text{if} \quad z_i-\tau_i > 0 \quad \text{otherwise} \quad y_i = 0.
\end{gathered}    
\label{eq:sar}
\end{equation}

Alternatively, hardware that supports sparse activations and reduced neuron-neuron communication without memory dynamics unlike the existing neuromorphic devices (e.g. Loihi 2 \cite{9605018}) are another alternative for the SAR implementation and potentially a preferable one since neuromorphic hardware does not naturally match with the dense connectivity and high memory requirements of neural operators.

A potential limitation of this formulation arises from its reliance on the ReLU activation which allows for the zero threshold neuromorphic conversion. In the original VSN formulation, the spike gate is applied to the incoming signal with an optimized threshold that is isolated from the signal sent and can potentially allow negative values. The resulting signal can also be operated on any subsequent activation function (that is zero at zero input) with a positive or negative value depending on the neuron input and threshold. Our sparsity formulation includes a subsequent general activation function, but the output of Eq.~\ref{eq:relu_sparse} is strictly non-negative, restricting all subsequent activations to positive-valued inputs which could limit the expressive capabilities of the ReLU-sparsity layer in terms of optimization. If the thresholds employed by the original VSN are predominantly positive, this behavior may resemble the original VSN formulation, however, this is just speculation and may not hold at all and represents a potential limitation of the proposed approach.

Consequently, future work should investigate alternative sparsity-inducing layers capable of preserving both positive and negative signal information that also avoiding surrogate-gradient optimization. Such approaches may provide a closer approximation to the original VSN formulation while retaining the optimization benefits of conventional artificial neural networks. Despite this limitation, the results presented in this work demonstrate improved or similar predictive performance relative to the original VSN framework across multiple activation functions, suggesting that the practical impact of this restriction, relative to the gradient mismatch issue, remains unclear and warrants further investigation.

To regulate activation sparsity, previous methods explore initialization \cite{yang2024neuralnetworkssparseactivation} as well as the previously mentioned fixed thresholding with shifted ReLU \cite{price2024deepneuralnetworkinitialization} and FATReLU \cite{pmlr-v119-kurtz20a}, but to allow guaranteed and optimal control over the sparsification/efficiency level without hyperparameter tuning, regularization-based training in the form of a sparsity loss term, similar to the VSN loss objection, is most effective. We replace the average spike-percentage regularization term utilized by the original VSN framework with norm-based activation regularization applied directly to the output of Eq.~\ref{eq:relu_sparse}. When applied post-ReLU, the norm-based terms will attempt to force vector elements towards zero which improves sparsity. Two activation terms, previous utilized for activation sparsity \cite{pmlr-v119-kurtz20a}, are explored: $L_1$ and Hoyer norm. For a feature vector $\mathbf{v}(\mathbf{x}_i) \in \mathbb{R}^{d}$ corresponding to node or evaluation point $\mathbf{x}_i$, the $L_1$ activation sparsity term is defined as

\begin{equation}
\mathcal{L}_{L1}^{(i)} = |\mathbf{v}(\mathbf{x}_i)|_1 = \sum_{j=1}^{d}|v_j(\mathbf{x}_i)|,
\end{equation}

which encourages sparse activations by penalizing activation magnitude. We additionally investigate the Hoyer sparsity measure,

\begin{equation}
\mathcal{L}_{H}^{(i)} = \frac{|\mathbf{v}(\mathbf{x}_i)|_1^2}{|\mathbf{v}(\mathbf{x}_i)|_2^2+\epsilon}
=\frac{\left(\sum_{j=1}^{d}|v_j(\mathbf{x}_i)|\right)^2}
{\sum_{j=1}^{d}v_j(\mathbf{x}_i)^{2}+\epsilon},
\end{equation}

where $\epsilon$ is a small constant introduced for numerical stability. With the loss term defined for a single node $\mathbf{x}_i$, we can subsequently average or sum over the provided grid to provide the sparsity loss contribution for the specific layer of compuation. Both regularization strategies have previously been explored for activation sparsification \cite{pmlr-v119-kurtz20a}. However, the Hoyer formulation provides improved control over sparsity due to its scale-invariant nature, whereas the $L_1$ norm primarily suppresses activation magnitude across the network. In the following section, we present the complete ANN-to-neuromorphic training methodology.

\begin{figure}[htbp]
    \centering
    \includegraphics[width=0.7\textwidth]{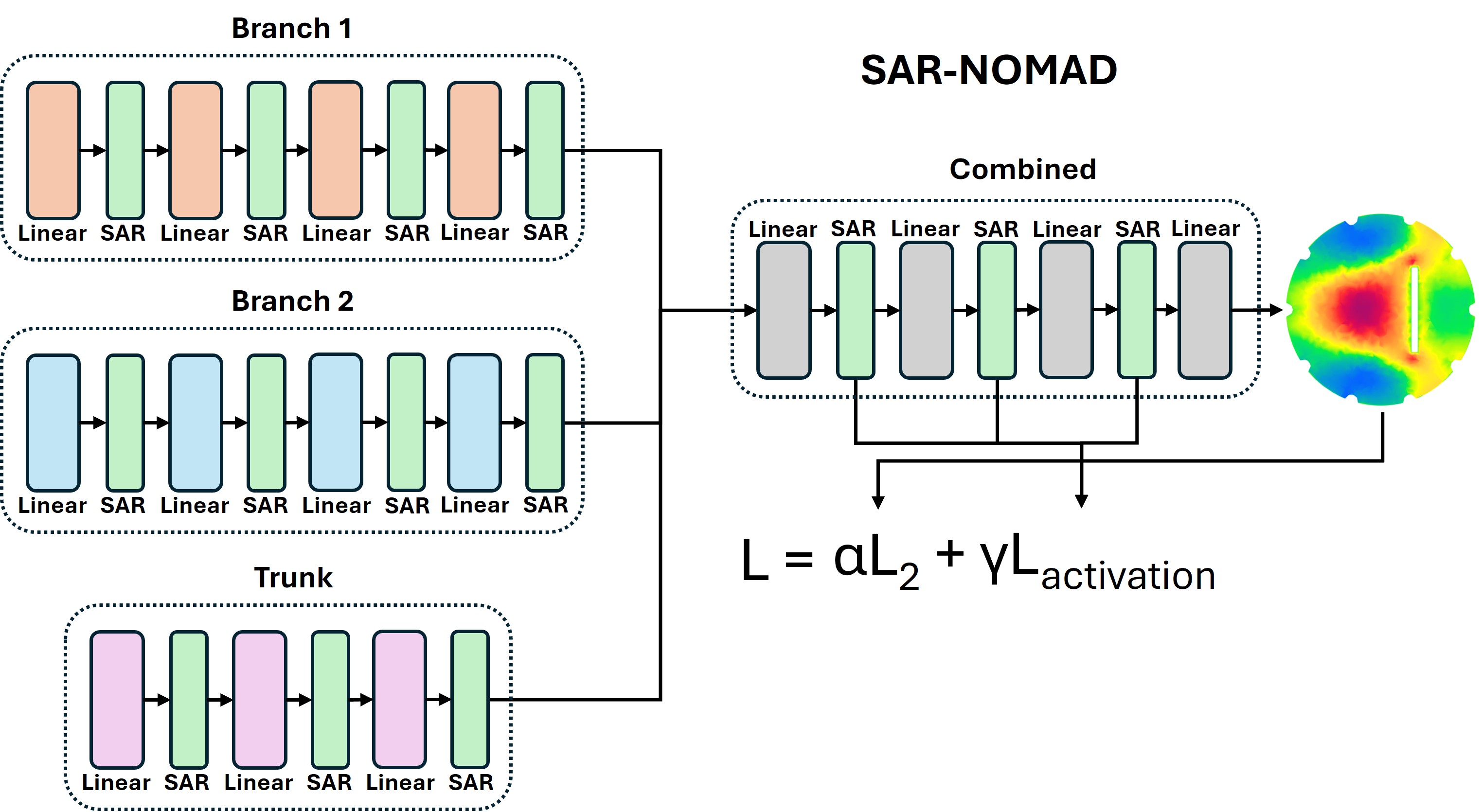}
    \caption{
    \textbf{Sparse-Activation-ReLU Nonlinear Manifold Decoder for Operator Learning}
    Architecture of SAR-NOMAD specifically for the 2D Heat Exchanger. In green, exists every SAR layer which naturally replaces the ReLU layers. Since every computation before SAR is a simple linear layer, we do not include a threshold parameter which is redundant in the presence of the bias term. As shown, the full-field output provides the $L_2$ error for the total objective function while the individual sparse activation outputs of the ReLU (from all SAR layers in the branch, trunk and combined networks not fully depicted in the figure) with the Hoyer/$L_1$ loss term provide the control on energy efficiency.
    }
    \label{fig:sar_nomad}
\end{figure}

\begin{figure}[htbp]
    \centering
    \includegraphics[width=1.05\textwidth]{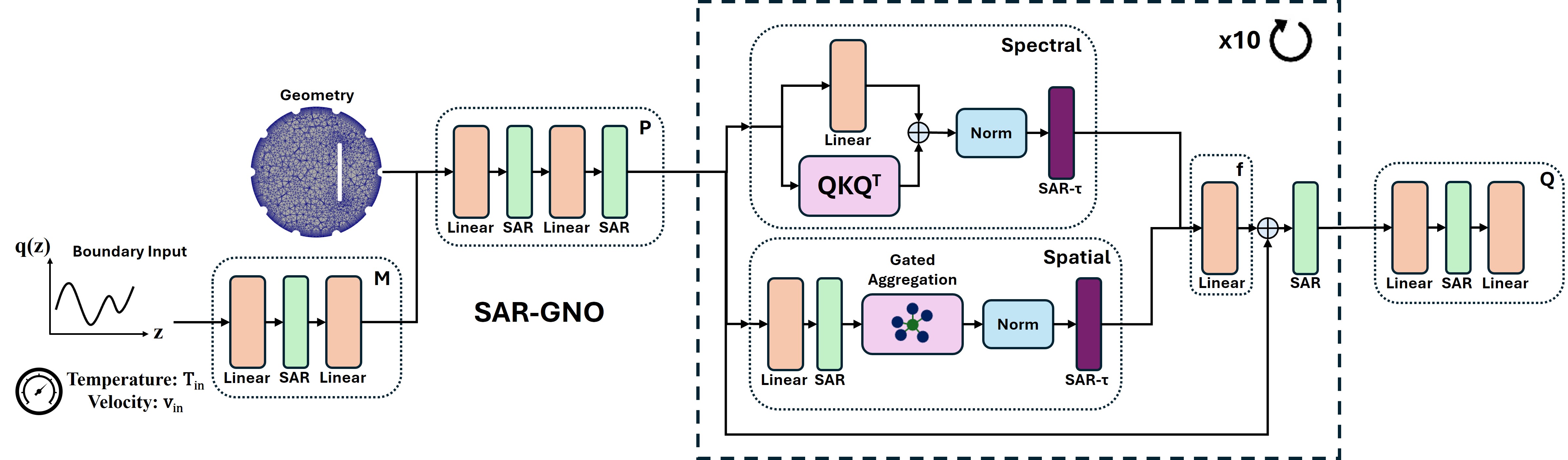}
    \caption{
    \textbf{Sparse-Activation-ReLU Graph Neural Operator}
    Architecture of SAR-GNO for the 2D Heat Exchanger. In green, exists every SAR layer included in the neural operator which does not utilize the optional threshold parameter. The green SAR layers replace either identity mappings or ReLU layers. Components in dark purple represent SAR with the optional threshold parameters $\boldsymbol{\tau}$ included for improved spiking control since they follow a normalization layer and non-linear computational layers. These threshold layers replace an identity mapping (spatial) and GeLU (spectral). SAR-GNO takes in boundary input with the input embedding mapping $M$ and combines it with the geometry coordinates to produce an input for the latent projection mapping $P$. Subsequent spectral-spatial blocks (10 layers total) provide global and local analysis that is combined through a collaboration layer $f$. Final a downlift layer $Q$ provides the final full-field reconstructed multi-physics output.
    }
    \label{fig:sar_gno}
\end{figure}

\subsection{ANN-to-Neuromorphic Conversion for SAR Neural Operators}
\label{sec:conversion}

To evaluate the proposed SAR framework, we integrate the layer and activation sparsity regularization into both the Nonlinear Manifold Decoder for Operator Learning (NOMAD) \cite{seidman2022nomadnonlinearmanifolddecoders} and Virtual Irregular Real-Time Sparse Operator (VIRSO) \cite{howes2026realtimesensinginaccessiblephysical} architectures, resulting in SAR-NOMAD (Figure \ref{fig:sar_nomad}) and SAR-GNO (Figure \ref{fig:sar_gno}). The trunk-branch structure of NOMAD naturally utilizes ReLU activations throughout the network, providing a straightforward transition to the proposed SAR implementation shown in Figure \ref{fig:sar_nomad}. We additionally investigate VIRSO, which employed GeLU activations within the architecture. To maintain as much consistency with the original VIRSO design, Sparse-Activation-ReLU layers are inserted at locations where VSNs were previously utilized shown in Figure \ref{fig:sar_gno} in green, followed by a GeLU activation to preserve the original activation behavior as closely as possible.

For conventional linear layers, which comprise the entirety of the NOMAD architecture, the SAR formulation is implemented as just a ReLU with the bias parameters providing a threshold-like dynamic. Within VS-GNO, certain layers do not contain bias parameters, specifically spectral/spatial convolutions followed by normalization layers. In these cases, optional subtraction parameter introduced in Eq.~\ref{eq:relu_sparse} is utilized to provide additional flexibility and improve training dynamics.

Activation sparsity regularization is applied to the outputs of each SAR layer. Specifically, the activations immediately following the ReLU operation are utilized when computing the sparsity penalty. For a network containing $L$ SAR layers, the sparsity contribution from each layer is computed independently and then summed to obtain a global measure of network sparsity. This produces a direct analogue to the energy-accuracy tradeoff provided by the VSN framework, where increased sparsity corresponds to reduced neuron activity and, consequently, reduced communication within the network.

The final training objective combines prediction accuracy and activation sparsity into a single loss function depicted in Figure \ref{fig:sar_nomad},

\begin{equation}
\mathcal{L}=
\alpha \cdot\mathcal{L}_{L2}
+
\gamma \cdot\sum_{l=1}^{L}\mathcal{L}_{\mathrm{activation}}^{(l)},
\label{eq:total_loss}
\end{equation}

where $\mathcal{L}_{L2}$ denotes the reconstruction error, $\mathcal{L}_{\mathrm{activation}}^{(l)}$ represents either the $L_1$ or Hoyer activation sparsity regularization term computed at layer $l$ (this could be a layer in a branch network from a trunk-branch model or a spectral block from a graph operator) and averaged over the input batch, $\alpha$ controls the emphasis placed on predictive accuracy, and $\gamma$ determines the strength of sparsity regularization. The $\mathcal{L}_{\mathrm{activation}}^{(l)}$ term, which is defined over the features for each grid point, might also be averaged over the evaluation domain as well as the batch, especially in the trunk and combined networks for NOMAD. Increasing $\gamma$ encourages greater activation sparsity and therefore lower neuron activity throughout the network. The sparsity hyperparameter $\gamma$ can be made unique for specific sparsity layers which is explored below.

Training is performed entirely within the ANN framework using the proposed SAR layers and sparsity regularization terms. Following training, the learned model can be transferred directly to a neuromorphic implementation operating with a threshold of zero. Under this formulation, neuron spiking activity becomes directly equivalent to the sparsity pattern learned during ANN optimization, allowing the resulting neuromorphic system to reproduce the same communication behavior observed during training. With the single-step framework, SAR-based layers not only reduce the latency of neuromorphic-based inference, but the computational overhead during training, providing an overall faster alternative to large STS VSN-based neural operators.

\begin{figure}[htbp]
    \centering
    \includegraphics[width=1\textwidth]{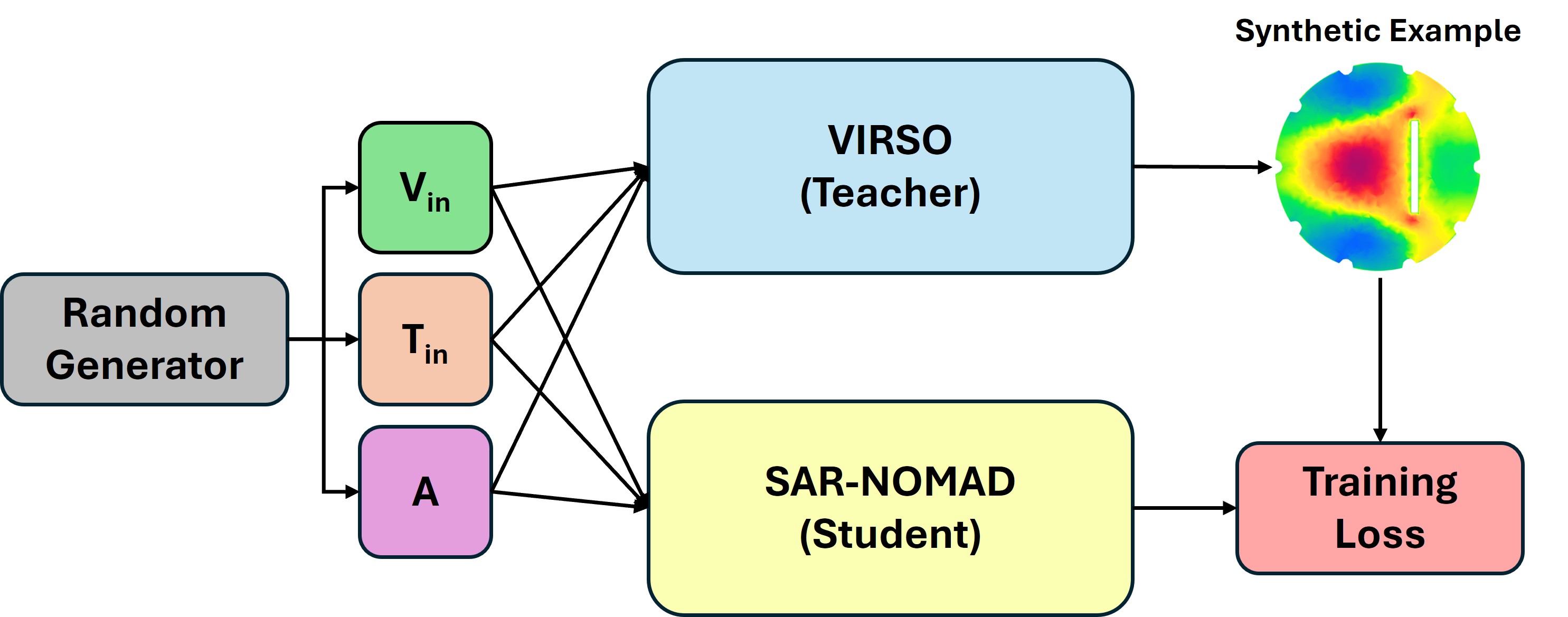}
    \caption{
    \textbf{Synthetic Distillation}
    We show the synthetic distillation framework utilized for neuromorphic virtual sensing. A graph-based VIRSO model, not native for neuromorphic hardware due to high connectivity and difficult integration, generates synthetic Heat Exchanger examples by randomly sampling input parameters for the Heat Exchanger dataset. These synthetic examples are compared against SAR-NOMAD's predictions, a more neuromorphic friendly model, allowing for improved $L_2$ error performance in SAR-NOMAD, ideally keeping efficiency the same.
    }
    \label{fig:synth_distillation}
\end{figure}

\subsection{Model Synthetic Distillation}
\label{sec:distillation}
Increasing sparsity within the proposed framework introduces a natural tradeoff with reconstruction accuracy, as higher sparsity restricts the amount of information that can be transmitted through the network. This effect is particularly pronounced in regimes with limited training data, which is a common constraint in virtual sensing applications where data generation via finite element methods or similar high-fidelity solvers is computationally expensive and time-consuming. Under such conditions, the reduced availability of training samples may prevent the model from learning an efficient communication strategy that is robust under high sparsity constraints. Also, sophisticated model designs, such as FNO \cite{DBLP:journals/corr/abs-2010-08895}, Wavelet Neural Operator (WNO) \cite{tripura2022waveletneuraloperatorneural}, and VIRSO \cite{howes2026realtimesensinginaccessiblephysical} that can better handle limited data, might present challenges when implemented within existing neuromorphic hardware (e.g. Loihi 2 \cite{9605018}) or other edge-deployable hardware that handle our SAR framework but exist with strict memory constraints, so we are further limited with less sophisticated, more neuromorphic/edge-friendly designs such as trunk-branch networks which present with simple FCN layers and independent grid point evaluation.

This behavior motivates the following unique neural operator framework: a non-neuromorphic-native or non-edge-friendly neural operator first learns an accurate approximation of the nonlinear operator mapping and subsequently transfers this knowledge to a more edge-deployable neural operator architecture, thereby improving its performance beyond what can be achieved through direct training alone.

To address this limitation, we employ knowledge distillation, a widely used technique in machine learning for transferring knowledge from a large teacher model to a more compact student model \cite{KANG2024108001} which has seen initial introduction within neural operators \cite{wan2026spectralinspiredoperatorlearninglimitedistill1, Chen2026-nndistill2}. In particular, we investigate synthetic distillation \cite{KANG2024108001, Zhou_2023}, where additional training data is generated using a pretrained teacher model. In this setting, a large and complex neural operator that is not directly suitable for sparse or neuromorphic deployment is used to produce synthetic field outputs corresponding to inputs sampled from a known distribution.

Given that the input distribution for the target application is assumed to be known, new training samples are generated by drawing inputs from this distribution and evaluating them using a high-capacity teacher model. In this work, the teacher model is the graph-based neural operator VIRSO without sparsity regularization. VIRSO's architecture renders it difficult to implement within edge-deployable hardware, but it can be used to efficiently generate corresponding solution fields for generated synthetic inputs, which are then incorporated into the training dataset of the sparsified neural operator that better aligns with hardware integration.

For this study, VIRSO-generated samples are used to augment the training data for the SAR-NOMAD implementation depicted in Figure \ref{fig:synth_distillation} where the main parameters that define the Heat Exchanger input is randomly generated and fed through both the teacher (VIRSO) and student (SAR-NOMAD) models defining a training loss to improve SAR-NOMAD's performance. While this procedure introduces additional computational overhead during training, it enables improved coverage of the input space and provides richer supervision for learning under sparsity constraints with no overhead during inference. Further work is required to reduce the cost of synthetic data generation and training or to explore alternative distillation strategies; however, the focus of this paper is to demonstrate the transition from a large, complex, and non-neuromorphic-friendly model to a sparse architecture that is compatible with neuromorphic or other edge-device deployment. This framework is intended to improve the ability of the sparse model to approximate the underlying nonlinear mapping between input conditions and output fields under limited-data, high-sparsity regimes.

\subsection{Bringing Activation-Regularized ReLU to Variable Spiking Operators}
\label{sec:hoyer_for_variable}
\paragraph{Activation-ReLU Loss Term:} Although the SAR framework is motivated by the observation that many regression-based neural operators with a VSN-type neuron operate in the single spike time step (1 STS) regime to reduce latency, where the temporal dynamics of the VSN become largely unnecessary, this does not imply that the original VSN formulation cannot benefit from multiple spike time steps. On the contrary, allowing neurons to accumulate information over time introduces temporal dynamics that can improve representational capacity. In particular, for transient prediction problems, the persistent membrane memory of the VSN may better integrate temporal information and improve long-term forecasting performance. Moreover, VSN based operators provide flexibility in its signals, allowing negative output. As a result, it would be also prudent to explore techniques to improve the performance VSN-based operators without the need for high latency, allowing for memory dynamics and better signal expressiveness. Motivated by these observations and beliefs, we additionally investigate improvements to the training methodology of Variable Spiking Neural Operators while preserving their underlying neuron dynamics.

Rather than proposing an alternative to surrogate-gradient optimization for neurons with persistent membrane memory, which remains an important direction for future work, we instead reconsider the sparsity objective employed during training. Specifically, we hypothesize that the original spike-percentage regularization term can be replaced with a more informative ReLU-defined, activation-based objective term similar to the SAR layer that takes difference between the current memory (before any reset at time step $t$) similar to the input $\mathbf{z}$ from SAR and the trainable threshold $\boldsymbol{\Theta}$ (similar the parameter $\boldsymbol{\tau}$ from SAR). Instead of operating on binary spike events, we define a continuous pre-spike activation corresponding to the membrane potential above the firing threshold,

\begin{equation}
\mathbf{a}^{(t)}
=\text{ReLU}\left(
\beta\mathbf{M}^{(t-1)}
+\mathbf{z}^{(t)}
-\boldsymbol{\Theta}
\right)\in \mathbb{R}^d,
\label{eq:vsn_activation}
\end{equation}

where $\beta\mathbf{M}^{(t-1)}+\mathbf{z}^{(t)}$ is the membrane potential at time $t$ immediately before thresholding and any potential reset. If a neuron $i$ does not fire, $a_i^{(t)}=\mathbf{0}$. Otherwise, $a_i^{(t)}$ measures the amount by which the membrane potential exceeds the firing threshold before the membrane is reset. This continuous activation therefore contains substantially more information than the corresponding binary spike event.

The Hoyer sparsity measure is then applied directly to the activations defined in Eq.~\ref{eq:vsn_activation}. Since zero-valued activations correspond to neurons whose membrane potentials never exceed the firing threshold, encouraging sparsity in these activations naturally promotes reduced spiking activity. Unlike the original spike-percentage regularization, which is computed solely from binary firing events, the proposed objective exploits the continuous activation values above threshold, providing richer optimization information while maintaining the same sparsity objective. During training, the original spike-percentage regularization term in Eq.~\ref{eq:vsn_loss} is therefore replaced by the activation-based sparsity objective following the same formulation as Eq.~\ref{eq:total_loss}, with the Hoyer loss computed from the activations in Eq.~\ref{eq:vsn_activation} and averaged over the spike time steps. The effectiveness of this activation-based regularization strategy is evaluated in Section~\ref{sec:activation_variable_spiking}.

\paragraph{ReLU Activations Towards Neighbor Thresholding in Spatial Graph Aggregation:} In addition to replacing activation-regularized ReLU layers with the VSN to improve the training of low-latency models, we extend the same thresholding concept to the graph spatial gating mechanism within variable spiking graph operators. Within VIRSO \cite{howes2026realtimesensinginaccessiblephysical}, each edge (neighbor) connection is weighted through a gating function, $\gamma(u,v)$, which is computed as a function of the source node $u$, destination node $v$, the associated edge attributes, and their Lipschitz positional encoding. The learned gate is intended to adaptively construct the graph by assigning larger weights to more informative neighbors while suppressing less relevant connections.

For all previous experiments, the graph connectivity was constructed using a fixed neighborhood size of $k=30$. Consequently, every node aggregates information from 30 neighboring nodes regardless of their relative importance. Although this dense aggregation improves representational capacity, it also increases computational cost and inference latency, particularly for future neuromorphic implementations where every neighbor corresponds to an additional synaptic event. To alleviate this burden, we introduce a thresholded graph gating mechanism that selectively removes low-importance neighbors.

Specifically, we define a threshold parameter, $\tau$, which is constrained to lie within the interval $[0,1]$. The threshold is subtracted from the learned gate value, and the resulting quantity is passed through a ReLU activation,

\begin{equation}
\hat{\gamma}(u,v) = \mathrm{ReLU}\left(\gamma(u,v)-\tau\right) \quad \tau \leftarrow \mathrm{clip}(\tau,0,1).
\label{eq:threshold_gate}
\end{equation}

Only neighbors with gate values exceeding the threshold contribute to the spatial aggregation since a gate weight of zero represents no addition to the output. The resulting thresholded gated aggregation output for the next convolution layer $\ell$ can therefore be expressed as

\begin{equation}
\mathbf{h}_{u}^{(\ell)}
=
\sum_{v\in\mathcal{N}(u)}
\hat{\gamma}(u,v)\mathbf{W}\mathbf{h}_{v}^{(\ell-1)},
\label{eq:thresholded_aggregation}
\end{equation}

where $\mathbf{W}$ denotes the linear weights applied to the spatial aggregation input before summation.

During neuromorphic deployment, the gating network itself is removed, and the learned thresholded gate values are treated as fixed synaptic weights under the assumption of a static computational graph. With our gating threshold technique, the spatial aggregation computation needed during neuromorphic deployed would then be significantly reduced since we only use a fraction of the original gating weight count before thresholding and forcing particular edge weights to zero. If the underlying graph topology changes, the gating module must be recomputed; however, the proposed thresholding formulation should remain applicable which is a potential area of further investigation.

We investigate two thresholding strategies. The first employs a single trainable threshold shared across all graph edges. The second predicts an edge-dependent threshold by introducing an additional output from the gating network, allowing each edge to learn its own adaptive threshold value.

To explicitly control the number of active neighbors, we introduce an additional sparsity loss weighted by the hyperparameter $\epsilon$. This loss is defined as the average activation of the thresholded gate values over all samples, nodes, and neighboring edges,

\begin{equation}
\mathcal{L}_{\mathrm{edge}}
=\frac{1}{N}
\sum_{i=1}^{N}
\frac{1}{|\mathcal{V}|}
\sum_{u\in\mathcal{V}}
\sum_{v\in\mathcal{N}(u)}
\hat{\gamma}_{i}(u,v),
\label{eq:edge_loss}
\end{equation}

where $N$ is the batch size and $\mathcal{V}$ denotes the set of graph nodes. Minimizing this objective encourages the network to reduce the number of active neighbors while preserving only the most informative graph connections.

The edge sparsity loss is incorporated into the overall optimization objective with the $\mathcal{L}_{L2}$ reconstruction error and spiking percentage error $\mathcal{L}_{\text{spike}}$ as

\begin{equation}
\mathcal{L}_{\mathrm{total}}= \alpha\cdot\mathcal{L}_{L2}+\gamma\cdot\mathcal{L}_{\text{spike}}+
\epsilon\cdot\mathcal{L}_{\mathrm{edge}}.
\label{eq:total_loss_edge}
\end{equation}

The weighting parameter $\epsilon$ 
provides explicit control over the trade-off between reconstruction accuracy and the number of active graph edges, enabling the model to learn computationally efficient graph representations while maintaining predictive performance. This technique, which we demonstrate below, outperforms training on a graph with smaller neighbor/k value most likely because it leaves the spectral layer untouched by the reduced edge count and allows the option to focus on farther neighbors than only the k closes nodes. Instead of generating a separate graph with smaller nodes, we reduce memory requirements and provide flexibility in neighbor choices. Theoretically, this concept can applied to the original VIRSO graph network and any SAR-based graph models which we have seen success in compared to lower k graph alternatives, but we chose to focus on the VSN implementation because of the emphasis of spiking in this paper and the increased usage of the spatial by the VSN-based graph model over the SAR alternative which forces the spatial layer to minimal communication. We found that SAR-based models are best utilized in the spectral-only form, completely dropping the unnecessary spatial block. In all fairness, this is also true for the VSN implementation for our chosen benchmark which shows better regression performance with the spectral layer, but we present the option for spatial aggregation with improved computation in case local calibration is needed for future applications.

\subsection{Problem Formulation}

Let $\mathcal{Y} \subset \mathbb{R}^d$ denote the $d$-dimensional spatial domain of the physical system for which a solution field is sought. The objective of neural operators is to learn the underlying nonlinear mapping between input boundary conditions and the corresponding physical solution fields through the following operator formulation:

\begin{equation}
\mathcal{G}: \mathcal{U} \rightarrow \mathcal{S},
\qquad
\mathcal{G}(\mathbf{u})(\mathbf{x}) = \mathbf{s}(\mathbf{x}),
\qquad
\mathbf{x}\in\mathcal{Y},
\label{eq:general_framework}
\end{equation}

where $\mathcal{G}$ denotes a nonlinear operator acting between functional spaces. The input is given by $\mathbf{u}=[u_1,\ldots,u_b]\in\mathcal{U}=\prod_{i=1}^{b}\mathcal{F}$, consisting of $b$ potentially multi-modal input components. The input space is defined such that each $u_i$ may correspond either to a scalar quantity in $\mathbb{R}$ or a functional input belonging to a space such as $L^2(D')$ defined over an alternative domain. The resulting output field is represented by $\mathbf{s}(\mathbf{x})\in\mathbb{R}^{k}$, describing $k$ physical quantities at location $\mathbf{x}$ within the domain $\mathcal{Y}$. Consequently, the output space is given by $\mathcal{S}\subset L^2(\mathcal{Y};\mathbb{R}^{k})$.

Kernel-based neural operators, such as VIRSO, approximate the mapping $\mathcal{G}$ through a sequence of nonlinear integral transformations inspired by Green's function formulations \cite{JMLR:v24:21-1524}. The resulting iterative update can be expressed as

\begin{equation}
\label{eq:kernel_int}
\mathbf{v}_{\ell+1}(\mathbf{x}) = \sigma\left(
W\mathbf{v}_{\ell}(\mathbf{x})
+
\int_{\mathcal{Y}}
\mathcal{K}_{\phi}(\mathbf{x},\mathbf{z})
\mathbf{v}_{\ell}(\mathbf{z})
d\mathbf{z}
\right),
\end{equation}

where $\mathbf{v}_\ell$ and $\mathbf{v}_{\ell+1}$ denote latent function representations evaluated at spatial location $\mathbf{x}\in\mathcal{Y}$ for layer indices $\ell=0,\ldots,L$. The nonlinear activation function $\sigma$ may correspond to common choices such as ReLU or Sigmoid. The term $W\mathbf{v}_{\ell}(\mathbf{x})$ provides a learnable residual mapping, while $\mathcal{K}_{\phi}$ denotes a parameterized kernel responsible for propagating information across the spatial domain through the integral operator. Collectively, these components enable the network to iteratively approximate the target nonlinear operator over the geometry of interest. 

VIRSO utilizes the Graph Fourier Transform to lift the input $\mathbf{v}_{\ell}(\mathbf{x})$ to the spectral domain to allow for the multiplication of a directly parameterized kernel that is optimized to approximate the convolution integral. VIRSO also utilizes a gated point-wise spatial aggregation as a local approximation of the integral. Combined with a projection mapping $f$, the spectral and spatial approximations provide both global and local analysis of the provided input while avoiding the previous scalability concerns. With VIRSO, the boundary input is lifted, with a mapping $M$, to a latent dimension and then copied to each evaluation coordinate. Then, the entire grid with feature vectors equal to coordinates and latent condition representations is fed through a projection layer $P$, to further lift the input into the defined intermediate dimension, and the subsequent convolution-integral layers. A final downlift layer $Q$ maps the grid's intermediate feature vectors to the final output channels. This encoder-decoder structure processes the entire grid and evaluation points at once.

Trunk-branch architectures such as NOMAD provide a different approximation to Equation \ref{eq:general_framework}. Multiple branch encoder networks (typically fully-connected) are utilized to lift the boundary inputs (profile and inlet values) to individual embeddings $b^i(u_i)\in\mathbb{R}^p$ for boundary input $i$. In addition, a trunk network is utilized to process the input queries (where we evaluate our solution). It learns a set of $p$ basis functions, $t(\mathbf{y}) \in \mathbb{R}^p$, that are optimized for the problem and evaluation coordinate. One or more evaluation points can be provided to the trunk branch, allowing for more natural handling of unseen evaluation coordinates. Typically, models such as the Multi-Input Operator Network (MIONet) \cite{jin2022mionetlearningmultipleinputoperators} compute an element-wise dot product between branches, $b(u^1,...,u^k) = b^1(u^1) \odot ... \odot b^k(u^k)$ for k boundary inputs, and a subsequent dot product with the trunk basis output, $t(\mathbf{y}) \in \mathbb{R}^p$, to finalize an output solution for the given trunk evaluation points. NOMAD recognizes that solutions might live on a low-dimensional nonlinear submanifold. This prompts NOMAD to map the individual branch embeddings and trunk basis (usually concatenated) to the final solution for each evaluation point with a nonlinear network (typically fully-connected) called a "combined" network.

\subsection{2D Heat Exchanger and Training Details}

To evaluate the proposed sparsity framework, we consider the steady-state flow analysis of a two-dimensional cross section extracted from a three-dimensional heat exchanger featuring enhanced heat transfer through a highly irregular dimpled surface and wavy tape insert geometry \cite{AHMED2024104583}. The benchmark follows the boundary-to-field reconstruction framework defined in Equation~\ref{eq:general_framework}, where pressure and velocity fields are reconstructed at 3,977 evaluation nodes from only two scalar inlet conditions and a discretized heat flux profile.

The heat exchanger benchmark represents a challenging evaluation with a high degree of geometric irregularity, complex vortex physics, and a four-component coupled output field. The geometry consists of a dimpled cylindrical channel with a wavy tape insert (Figure~\ref{fig:2d_hx}a), whose complex surface topology removes some rotational and reflective symmetries that could be exploited. The wavy insert generates large recirculation regions and secondary vortex structures that significantly increase the complexity of the underlying flow physics and challenge operator learning architectures that lack sufficient local resolution capability \cite{AHMED2024104583}.

The input consists of two scalar inlet conditions shown in Figure \ref{fig:2d_hx}, temperature $T_{in}$ and axial velocity $v_{in}$, together with a 100-point discretization of an axial wall heat flux profile defined by the following equation $A\sin(\pi x/H)$, where $H$ is the height of the original 3D heat exchanger and $A$ is a provided amplitude input. The output consists of four physical fields evaluated at 3,977 spatial locations on a two-dimensional axial cross section $\mathcal{Y}_{hx}$: pressure $p(z,y)$ and the three velocity components $u_x(z,y)$, $u_y(z,y)$, and $u_z(z,y)$, resulting in the sparse-to-dense reconstruction ratio of $\frac{3,977 \times 4}{102}
\approx 156:1$. The corresponding operator mapping can be expressed as

\begin{equation}
    \mathcal{G}_{hx}: (\mathbb{R})^2 \times L^2(\mathbb{R})
    \rightarrow
    (L^2(\mathcal{Y}_{hx}))^4,
    \qquad
    \mathcal{Y}_{hx} \subset \mathbb{R}^{2}.
\end{equation}

In addition to the velocity components, we compute the resulting velocity magnitude from our predicted values and provide the loss in magnitude reconstruction as a regularization term during training and as a fifth performance metric during inference among the other four output channels. The velocity magnitude regularization term was fixed at $\lambda_{mag}=0.1$ for all experiments during training. The term is defined by the following equation: 

\begin{equation}
    e_{\text{mag}}
    = \lambda_{mag}\frac{\left\|\hat{u}_x^2 + \hat{u}_y^2 + \hat{u}_z^2
      - u^2\right\|_2}{\|u^2\|_2},
\end{equation}

Training, validation, and test data were generated using ANSYS Fluent \cite{AnsysFluent2024}, resulting in 988 training examples, 248 validation examples, and 310 testing examples. Due to the computational cost associated with generating high-fidelity computational fluid dynamics solutions for complex geometries, this benchmark additionally serves as a representative virtual sensing application where limited training data motivates the use of synthetic distillation techniques. We utilize a VIRSO model with 6 layers, width of 64, and 64 spectral modes as our teacher model to generate 1000, 2000, 4000, and 8000 synthetic examples that are added to the original 988 examples in our training dataset. The same validation and test sets for the original performance are utilized for performance evaluation of our synthetic distillation framework. We were able to generate synthetic inputs for our teacher model utilizing the data distributions employed for our ANSYS generated data, which are described below:

\begin{equation}
A \sim \mathcal{U}(18, 22)\,[\mathrm{kW/m^2}], \quad 
T_{in} \sim \mathcal{U}(263.7, 322.3)\,[\mathrm{^\circ C}], \quad 
v_{in} \sim \mathcal{U}(4.05, 4.95)\,[\mathrm{m/s}].
\label{eq:distributions}
\end{equation}

Two SAR-based neural operator architectures were evaluated within this work: SAR-NOMAD and SAR-GNO shown in Figures \ref{fig:sar_nomad} and \ref{fig:sar_gno}. SAR-NOMAD follows the original trunk-branch formulation consisting of two branch networks, a trunk network, and a final combination network. Each branch network and the combined branch contains four FCN layers while the trunk network contains three. SAR layers are inserted after every linear layer except the final output layer in the combined network. A hidden dimension of 256 was utilized for all sub-networks, representing an input of $3*256$ for the combined network. This architecture provides a natural integration of the proposed sparsity framework due to its existing reliance on ReLU activations throughout the network. The optional subtraction parameter $\boldsymbol{\tau}$ was not utilized since the bias parameters of the linear layers provide a natural threshold-like dynamic after the application of linear weights.

For SAR-GNO, both a full and spectral-only configuration are investigated. Each model contains ten operator layers with a latent width of 64 and 100 spectral modes. The spectral-only model removes the spatial interaction block while retaining the spectral operator components. SAR layers are incorporated throughout the architecture, including the input embedding layers, projection layers, downlift layers, spectral operator blocks, and spatial operator blocks. Within the spectral and spatial layers which are unlike the straightforward linear layers and have normalization blocks, we utilized the optional subtraction parameter $\boldsymbol{\tau}$ to enhance filtering by mimicking a threshold like behavior. To preserve behavior consistent with the original VIRSO architecture, each sparsity layer is implemented as a ReLU filtering operation followed by a GeLU activation.

For the sparsity loss term, we add the batch-wise averaged activation loss for every sparsity layer in SAR-NOMAD and SAR-GNO, multiply our $\gamma$ hyperparameter, and then add this term to the L2 loss term to define our entire objective function.

All sparsity activation regularization strategies discussed in Sections \ref{sec:sparsification} and \ref{sec:conversion} are applied identically across both architectures. For each SAR layer, activations immediately following the ReLU operation are collected and utilized in the sparsity regularization term. The resulting sparsity penalties are summed across all layers and incorporated into the final loss function. Training is performed entirely within the ANN framework. As discussed before, conversion to a neuromorphic implementation is simple when operating with a zero-threshold spiking mechanism. This formulation allows us to compare directly the efficiency performance of SAR layers compared to traditional spiking layers such as VSN and LIF since the sparsity percentage (essentially binary count of which dimension is zero or nonzero) is identical in comparison to the spiking percentage.

SAR-NOMAD and SAR-GNO were trained using the Adam optimizer with an initial learning rate of $10^{-3}$ and a batch size of 16. Learning rate scheduling was performed using a step decay strategy with a step size of 40 epochs and a decay factor of 0.5. Early stopping was employed with a patience of 40 epochs with all results presented representing the best relative L2 error with the validation dataset. It should be noted that all graph-based models (VIRSO, VS-GNO, SAR-GNO), unless specified, operated on a graph generated from the Heat Exchanger grid using the KNN algorithm with a k value of $30$ nearest neighbors. Moreover, the graph Laplacian utilized for the spectral block was chosen to be distance-weighted.

SAR-NOMAD was trained for a maximum of 500 epochs with a weight decay coefficient of $10^{-5}$ while SAR-GNO was trained for a maximum of 200 epochs with a weight decay coefficient of $10^{-3}$. Unless otherwise stated, all remaining training procedures, optimization settings, and evaluation methodologies were identical between SAR-NOMAD and SAR-GNO. In this paper, we set the $L_2$ norm hyperparmaeter $\alpha$ to 1 in Equation \ref{eq:total_loss} while we vary the $\gamma$ parameter to explore different levels of sparsity and accuracy and demonstrate the controllability of our framework. We explore both the  and Hoyer loss terms and their effectiveness towards accurate and energy-efficient neural operators. Success is not defined by a specific $L_2$ error threshold but providing better accuracy at similar or improved efficiency compared to the VSN and demonstrating success in the controlling the accuracy-energy tradeoff without severe reconstruction error degradation. As mentioned, the relative $L_2$ error, defined below with $n$ interior nodes and output channel $o$, 

\begin{equation}
    e_{\text{rel}}
    = \frac{\|\hat{\mathbf{s}}_o - \mathbf{s}_o\|_2}
           {\|\mathbf{s}_o\|_2}
    = \frac{\sqrt{\sum_{p=1}^{n}
      \left(\hat{s}_o(\mathbf{x}_p) - s_o(\mathbf{x}_p)\right)^2}}
           {\sqrt{\sum_{p=1}^{n} s_o(\mathbf{x}_p)^2}},
\end{equation}

was utilized as the accuracy-based loss term for training and for performance evaluation during testing. All training and evaluation were performed (with spiking simulated by Snntorch \cite{snntorch}) using an NVIDIA GH200 hardware provided through the DeltaAI cluster and an NVIDIA H200/A100 with the Delta cluster at the National Center for Supercomputing Applications (NCSA) \cite{Delta}. The resulting experiments provide a challenging assessment of SAR neural operators on a highly irregular, multi-output reconstruction problem characterized by extreme boundary-to-field mapping requirements and complex nonlinear flow physics. Lastly, as previously mentioned, we explore the performance of SAR-NOMAD and VS-NOMAD under synthetic distillation in Section \ref{sec:distillation_results} and Figure \ref{fig:synth_distillation}. Using trained weights from the VIRSO architecture previously optimized under the same training hyperparameters (6 spectral-spatial layers, 64 modes, and a width of 64), we synthetically generated 2D Heat Exchanger samples then used them, alongside the original data used to train VIRSO, to improve the convergence of SAR-NOMAD with a gamma value equal to $0.001$, $0.005$, and $0.01$.

In addition to SAR-based neural operators, we also present results from a variable spiking NOMAD (VS-NOMAD) and variable spiking GNO (VS-GNO) \cite{howes2026neuroscienceinspiredgraphoperators} that implement the VSN to provide as a spiking benchmark and demonstrate the enhanced training of the activation sparsity layer over the variable spiking implementation. The VS-GNO results utilized the same previously explored graph architecture \cite{howes2026neuroscienceinspiredgraphoperators} and training hyperparameters as its SAR-GNO counterpart. The VS-NOMAD was also implemented with the same parameters as its SAR counterpart. For activation functions, the VS-NOMAD uses ReLU while the VS-GNO utilizes GeLU layers. The VSN was implemented with unique thresholds and leakage parameters along the hidden feature dimension (256 for VS-NOMAD and 100 for VS-GNO). We explored two surrogate gradient functions implemented by Snntorch \cite{snntorch}: fast sigmoid and arctangent. We also explored three different spiking time steps for the VS-NOMAD: 1, 10, and 20 STS to provide analysis of the spiking temporal dynamics and two different time steps (1 and 10 STS) for the spectral-only VS-GNO (full version could not fit on the H200's available memory). The loss function utilized for the VSN-based neural operators is regularized by the spiking percentages which is averaged over each layer within a sub-component (branch, spectral-block, etc.) and then summed and multiplied by the spiking hyperparameter $\gamma$ which provides the same energy-accuracy tradeoff as the sparsity regularization term. We also further explore in Section \ref{sec:activation_variable_spiking} replacing the percentage-based regularization term for variable spiking with an activation-inspired Hoyer loss term utilizing a ReLU function applied to the difference between neuron memory and threshold at each spike pass. We compare the resulting reconstruction error and spiking percentages with the original VSN framework for various different gamma values. Moreover, we utilize the presented full VS-GNO implementations for the 2D Heat Exchanger and experiment with our gated-aggregation threshold technique in Section \ref{sec:hoyer_for_variable}. We utilize the same $\gamma$ choises for the 1 STS full VS-GNO as well as two values for our edge-count hyperparameter $\epsilon$: $0.01$ and $0.001$. We utilized the two presented gate threshold definitions and also present VS-GNO results with KNN graphs having 5, 10, and 15 neighbors with a distance-weighted graph Laplacian to demonstrate the superior performance and flexibility of our presented filter.

Lastly, we also include the Leaky-Integrate-Fire neuron (LIF) as another surrogate gradient based alternative that essentially replaces the VSN (and the activation function it utilizes), defining the LIF-NOMAD and LIF-GNO neural operators. We explore direct and rate based encoding for our boundary and grid input with the latter generated by Snntorch's \textit{spikegen.rate} function \cite{snntorch}. For the LIF-NOMAD, the direct encoding results utilized the same $\gamma$ parameters as VSN with 1, 10, and 20 STS while the rate encoding results were performed with 10 and 20 STS. LIF-GNO (spectral and full) explored 1 and 10 STS for direct encoding and 10 STS for rate encoding with $\gamma$ value of $0$ and $0.5$. We only trained the spectral LIF-GNO for 10 STS due to memory constraints of the H200. For LIF-NOMAD and LIF-GNO, the arctangent surrogate gradient function was utilized. In addition, all spiking percentages presented for the LIF-based models are the result of the original calculated percentage divided by $1.7$ in order to match the energy efficiency of the VSN which, due to its variable communication, does consume more energy based on a SpiNNaker2 analysis \cite{garg2023neuroscienceinspiredscientificmachinepart2}. This correction is admittedly crude and might present an exaggeration of the LIF's energy reduction, but without physical hardware results it is difficult to implement. Further integration on physical neuromorphic hardware itself is needed to truly test the energy efficiency performance of SAR, VSN, or LIF based models. 

\subsection{Lid Driven Cavity and Training Details}

To further test the performance of the activation-based sparsity regularization, the Lid-Driven Cavity (LDC) benchmark was utilized. Unlike conventional field-to-field benchmarks, the LDC problem requires reconstructing high-dimensional spatial fields from a low-dimensional temporal forcing signal while operating on a structured computational mesh.

The computational domain consists of a two-dimensional square cavity, depicted in Figure \ref{fig:2d_ldc}a, in which fluid motion is generated by a moving upper boundary. Rather than prescribing a constant lid velocity, the network receives a time-varying lid velocity profile, $V(t)$, sampled over 90 discrete time steps, as the input shown in Figure \ref{fig:2d_ldc}a,b. The target output consists of three coupled flow quantities evaluated over 4,225 interior spatial nodes: the pressure field $p(x,y)$, the velocity magnitude $|\mathbf{v}(x,y)|$, and the turbulent kinetic energy field $k(x,y)$. This corresponds to reconstructing 12,675 output quantities from 90 input values, resulting in an approximate reconstruction ratio of $141{:}1$.

The underlying fluid dynamics are governed by the incompressible Reynolds Averaged Navier Stokes (RANS) equations together with a standard $k$-$\varepsilon$ turbulence closure,

\begin{equation}
\nabla \cdot \boldsymbol{v} = 0,
\end{equation}
\begin{equation}
\frac{\partial \boldsymbol{v}}{\partial t} + 
(\boldsymbol{v} \cdot \nabla)\boldsymbol{v} = 
-\nabla p + \nabla \cdot \left[ \left( \nu + \nu_t \right) 
\left( \nabla \boldsymbol{v} + \nabla \boldsymbol{v}^\top 
\right) \right],
\end{equation}
\begin{equation}
\frac{\partial k}{\partial t} + \boldsymbol{v} \cdot 
\nabla k = P_k - \varepsilon + \nabla \cdot 
\left[ \left( \nu + \frac{\nu_t}{\sigma_k} \right) 
\nabla k \right].
\end{equation}

The objective of the neural operator is to approximate the nonlinear operator

\begin{equation}
\mathcal{G}_{ldc}: L^2(\mathbb{R}_{\geq0}) \to 
(L^2(\mathcal{Y}_{ldc}))^3, \quad 
\mathcal{Y}_{ldc} \subset \mathbb{R}^2.
\end{equation}

Unless otherwise stated, all models are trained using a batch size of 16 for a maximum of 500 epochs with the Adam optimizer. The initial learning rate is set to $5\times10^{-4}$ and is reduced by a factor of 0.5 every 20 epochs using a StepLR scheduler. A weight decay of $10^{-3}$ is employed for regularization, and all inputs and outputs are normalized using range normalization. Early stopping is performed using a validation patience of 20 epochs to mitigate overfitting with all results presented representing the best relative L2 error with the validation dataset. All training was performed on the Delta supercomputer at NCSA, utilizing NVIDIA A100 and H200 GPUs.

To evaluate the proposed sparsity framework, experiments are performed using both the proposed SAR-NOMAD architecture and the original VS-NOMAD architecture. Since the preceding Heat Exchanger experiments demonstrate (shown below) that the Hoyer activation sparsity regularization consistently provides a superior accuracy-sparsity tradeoff compared to the $L_1$ formulation, only the Hoyer loss term is considered for SAR-NOMAD throughout the remainder of this work. For comparison, VS-NOMAD is evaluated using both the fast sigmoid and arctangent surrogate gradient functions under the 1, 10, 20, and 30 spike step operating regimes. We also implement LIF-NOMAD with direct and rate encoding for the same time steps (1, 10, 20, 30 for direct and 10, 20, 30 for rate). Similar to the Heat Exchanger, we divided all calculated percentages from the LIF models by $1.7$ to ideally match the efficiency of variable based alternatives.

To ensure a fair comparison, all models employ an identical architecture consisting of a hidden width of 256 with three trunk FCN layers, four branch FCN layers, and four combined FCN layers. Since the Lid-Driven Cavity benchmark is parameterized solely by the temporal lid velocity profile, only a single branch network is required to encode the input forcing signal. Consistent with the 2D Heat Exchanger experiments, the reconstruction weighting parameter is fixed at $\alpha=1$ for all models, while the sparsity weighting coefficient $\gamma$ is varied to investigate the resulting accuracy-efficiency tradeoff. For SAR-NOMAD, $\gamma$ controls the Hoyer activation sparsity regularization, whereas for VS-NOMAD it controls the spike-percentage regularization objective. Increasing $\gamma$ therefore encourages reduced communication activity in both formulations, enabling a direct comparison of reconstruction accuracy as network sparsity increases. 


\section{Results}

\begin{figure}[htbp]
    \centering
    \includegraphics[width=1\textwidth]{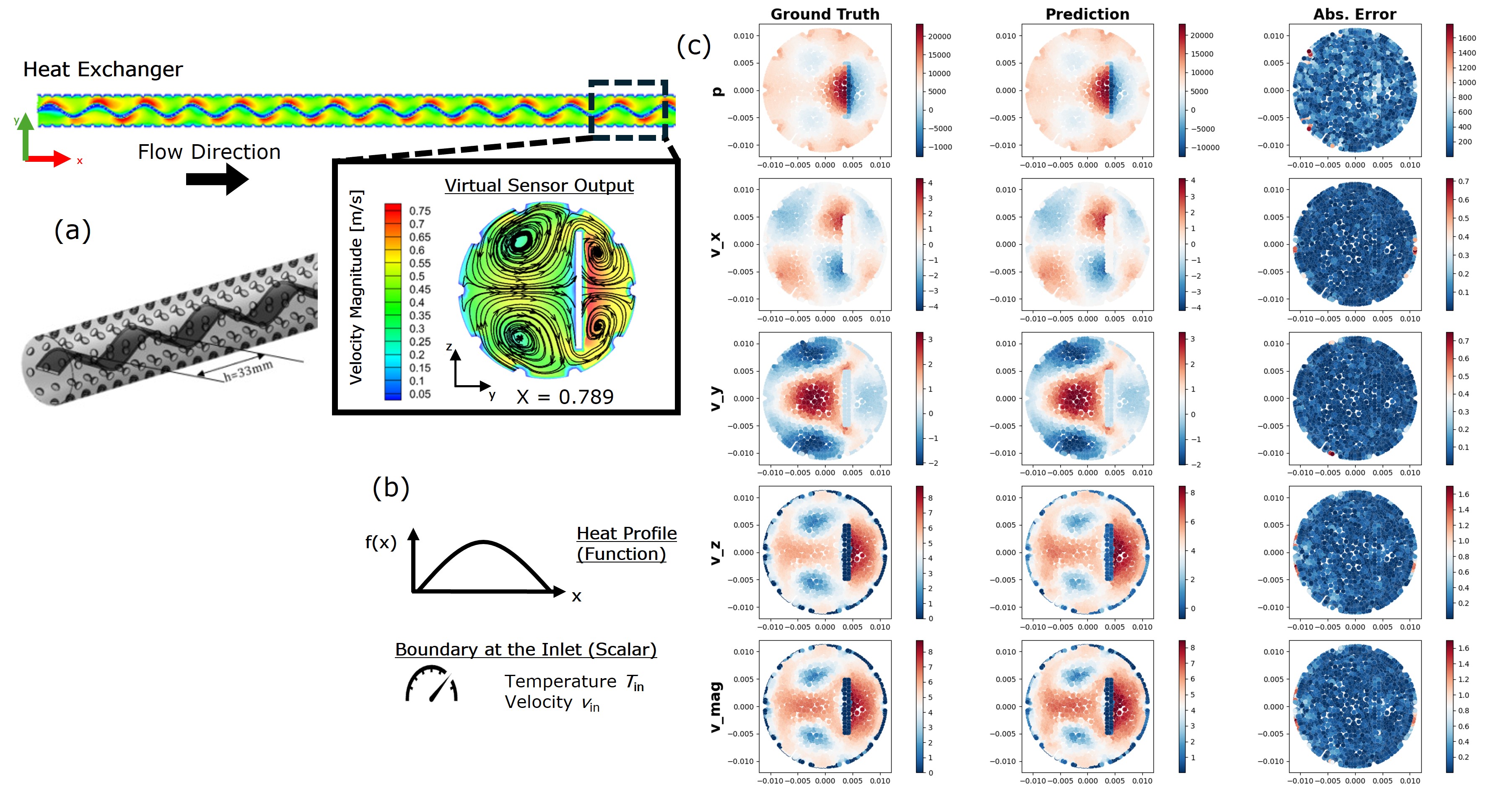}
    \caption{
    \textbf{2D Heat Exchanger} \textbf{(a)} We define the 2D Heat Exchanger geometry which is a cross-sectional slice of a 3D grid at axial position $x=0.789$ with a dimpled wall surface and tape insert utilized for better heat transfer, creating large vortex flow behavior and an irregular geometry with reduced symmetry \textbf{(b)} The boundary input utilized for our 2D Heat Exchanger. It includes a wall heat profile defined along the axial direction of the original 3D geometry and two inlet quantities: temperature and axial velocity/speed. \textbf{(c)} The 50th percentile test example for the SAR-NOMAD model with the Hoyer loss term and $\gamma=0.005$. We see that the main physics and vortex behavior is closely captured with mostly stochastic absolute error behavior and only small absolute error hotspots located around the wall surface.
    }
    \label{fig:2d_hx}
\end{figure}

\subsection{Heat Exchanger Performance (NOMAD)}

\begin{table*}[t]
\centering
\caption{$L_2$ errors and spiking percentages for SAR-NOMAD ($L_1$ and Hoyer loss terms) and VS/LIF-NOMAD (fast sigmoid and arctanget surrogate gradients) with the 2D Heat Exchanger.}
\makebox[\textwidth][c]{\begin{tabular}{c|cccccc|ccccc}
\hline
 & \multicolumn{6}{c|}{Mean Relative L2 Errors (\%)} & \multicolumn{5}{c}{Mean Spiking Percentages (\%)} \\
Spiking Model
& $p$
& $v_z$
& $v_y$
& $v_x$
& $v_{\mathrm{mag}}$
& Mean
& Trunk
& B1
& B2
& Comb
& Mean \\
\hline
Base Model & 0.70 & 1.18 & 1.11 & 0.69 & 1.345 & 1.00 & 42.39 & 18.76 & 22.88 & 40.90 &  31.23\\
\hline
SAR $L_1$ $\gamma=0.0005$ & 0.63 & 0.93 & 0.94 & 0.63 & 1.31 & 0.89 & 32.91 & 0.10 & 5.63 & 40.00 & 19.66 \\
SAR $L_1$ $\gamma=0.0005$* & 0.69 & 1.02 & 1.03 & 0.72 & 1.43 & 0.98 & 27.94 & 0.29 & 8.34 & 37.13 & 18.43 \\
SAR $L_1$ $\gamma=0.001$ & 0.64 & 0.85 & 0.98 & 0.65 & 1.32 & 0.89 & 30.33 & 0.10 & 4.37 & 37.40 & 18.05 \\
SAR $L_1$ $\gamma=0.0015$ & 0.65 & 1.03 & 1.02 & 0.65 & 1.34 & 0.94 & 27.32 & 0. & 3.69 & 36.45 & 16.87 \\
SAR $L_1$ $\gamma=0.005$ & 9.96 & 5.79 & 5.71 & 5.46 & 10.42 & 7.47 & 20.75 & 0. & 0.01 & 30.78 & 12.89 \\
\hline
SAR Hoyer $\gamma=0.0001$ & 0.79 & 1.26 & 1.20 & 0.88 & 1.52 & \textbf{1.13} & 35.12 & 1.02 & 5.64 & 30.18 & \textbf{17.99} \\
SAR Hoyer $\gamma=0.0005$ & 1.47 & 3.18 & 2.81 & 2.07 & 2.60 & \textbf{2.43} & 24.20 & 1.05 & 2.92 & 19.79 & \textbf{11.99} \\
SAR Hoyer $\gamma=0.0005$* & 3.62 & 6.89 & 6.87 & 4.65 & 5.79 & \textbf{5.56} & 13.38 & 4.77 & 2.83 & 8.26 & \textbf{7.31} \\
SAR Hoyer $\gamma=0.001$ & 2.14 & 4.42 & 3.55 & 2.39 & 2.96 & \textbf{3.09} & 17.81 & 1.86 & 2.07 & 14.09 & \textbf{8.96} \\
SAR Hoyer $\gamma=0.005$ & 4.11 & 6.57 & 5.82 & 4.84 & 5.72 & \textbf{5.41} & 8.91 & 0.60 & 0.83 & 9.15 & \textbf{4.87} \\
\hline
VS Fast Sig. 1 STS $\gamma=0$ & 37.23 & 38.56 & 33.21 & 34.15 & 41.40 & 36.91 & 6.15 & 7.71 & 15.27 & 4.29 & 8.36 \\
VS Fast Sig. 1 STS $\gamma=10^{-6}$ & 49.61 & 52.06 & 34.79 & 38.87 & 52.67 & 45.60 & 9.61 & 4.52 & 11.51 & 1.33 & 6.74 \\
VS Fast Sig. 10 STS $\gamma=0$ & 15.59 & 26.36 & 21.74 & 22.00 & 25.96 & 22.33 & 15.13 & 6.83 & 16.96 & 5.10 & 11.01 \\
VS Fast Sig. 10 STS $\gamma=10^{-6}$ & 20.39 & 27.20 & 22.64 & 24.64 & 28.09 & 24.59 & 15.57 & 5.76 & 17.02 & 3.47 & 10.45 \\
VS Fast Sig. 20 STS $\gamma=0$ & 14.58 & 25.85 & 20.06 & 19.70 & 23.70 & 20.78 & 14.40 & 10.04 & 17.36 & 4.90 & 11.68 \\
VS Fast Sig. 20 STS $\gamma=10^{-6}$ & 22.51 & 26.04 & 26.52 & 27.43 & 31.92 & 26.88 & 15.40 & 7.52 & 16.29 & 3.52 & 10.68 \\

\hline
VS Arctan. 1 STS $\gamma=0$ & 5.09 & 11.35 & 11.25 & 11.24 & 11.18 & 10.02 & 17.08 & 16.64 & 14.55 & 7.39 & 13.92 \\
VS Arctan. 1 STS $\gamma=10^{-6}$ & 49.37 & 17.17 & 17.22 & 38.38 & 51.57 & 34.74 & 13.01 & 1.27 & 13.14 & 2.45 & 7.47 \\
VS Arctan. 10 STS $\gamma=0$ & 4.25 & 9.93 & 8.79 & 8.21 & 8.49 & 7.93 & 20.72 & 14.02 & 18.21 & 7.69 & 15.16 \\
VS Arctan. 10 STS $\gamma=10^{-6}$ & 12.72 & 18.13 & 19.05 & 18.62 & 20.48 & 17.58 & 19.35 & 11.43 & 16.58 & 3.17 & 12.63 \\
VS Arctan. 20 STS $\gamma=0$ & 3.38 & 9.51 & 9.49 & 8.86 & 8.52 & 7.95 & 21.30 & 7.55 & 16.66 & 8.53 & 13.51 \\
VS Arctan. 20 STS $\gamma=10^{-6}$ & 7.18 & 14.10 & 14.24 & 12.22 & 13.35 & 12.22 & 20.06 & 5.86 & 14.44 & 3.32 & 10.92 \\
\hline
LIF Arctan. 1 STS DE $\gamma=0$ & 6.23 & 16.57 & 14.40 & 14.86 & 15.67 & 13.54 & 20.43 & 22.40 & 20.54 & 21.63 & 21.25 \\
LIF Arctan. 1 STS DE $\gamma=10^{-6}$ & 7.03 & 19.42 & 17.19 & 18.22 & 19.03 & 16.18 & 15.23 & 0. & 9.52 & 21.33 & 11.52 \\
LIF Arctan. 10 STS DE $\gamma=0$ & 3.36 & 11.34 & 10.39 & 9.50 & 8.50 & 8.62 & 16.71 & 8.86 & 9.83 & 22.59 & 14.50 \\
LIF Arctan. 10 STS DE $\gamma=10^{-6}$ & 3.49 & 10.87 & 9.93 & 9.26 & 8.50 & 8.41 & 16.57 & 8.12 & 9.40 & 22.45 & 14.14 \\
LIF Arctan. 20 STS DE $\gamma=0$ & 3.29 & 11.64 & 11.11 & 10.24 & 9.67 & 9.19 & 15.26 & 7.98 & 9.26 & 18.69 & 12.80\\
LIF Arctan. 20 STS DE $\gamma=10^{-6}$ & 3.16 & 10.56 & 10.35 & 9.45 & 8.46 & 8.40 & 16.52 & 8.17 & 9.91 & 21.24 & 13.96 \\
\hline
LIF Arctan. 10 STS RE $\gamma=0$ & 55.39 & 107.42 & 99.72 & 41.36 & 57.88 & 72.35 & 8.84 & 3.21 & 5.85 & 9.04 & 6.74 \\
LIF Arctan. 10 STS RE $\gamma=10^{-6}$ & 54.90 & 109.83 & 101.43 & 42.08 & 58.73 & 73.39 & 9.17 & 2.82 & 4.80 & 8.86 & 6.41 \\
LIF Arctan. 20 STS RE $\gamma=0$ & 55.36 & 99.52 & 83.36 & 41.55 & 57.46 & 67.45 & 8.92 & 6.40 & 5.78 & 8.60 & 7.43 \\
LIF Arctan. 20 STS RE $\gamma=10^{-6}$ & 55.88 & 98.60 & 86.77 & 41.59 & 56.98 & 67.96 & 8.50 & 3.41 & 5.37 & 8.43 & 6.43 \\
\hline
\end{tabular}}
\label{tab:sparsity_results}
\caption*{\textit{*Training loss utilized 3$\gamma$ for Trunk and Comb terms and $\gamma$/3 for Branch terms}}
\end{table*}

\paragraph{SAR-NOMAD Results:} Table \ref{tab:sparsity_results} summarizes the accuracy of the different loss terms for SAR-NOMAD ($L_1$ and Hoyer) together with different surrogate gradient functions for VS-NOMAD on the 2D Heat Exchanger benchmark. For SAR-NOMAD the reported spiking percentage is equivalent to the spike activity obtained after neuromorphic conversion using a zero activation threshold, making the comparison to VSN/LIF spiking results easier. As a result, we will discuss the sparsity of SAR layers in terms of spiking based off the previously discussed neuromorphic formulation. The baseline NOMAD model achieves approximately 1\% average relative $L_2$ error across the output channels. Despite being trained without any explicit sparsity regularization, the baseline architecture exhibits nearly 30\% average spiking across the network components when converted to an event-driven representation. This indicates that the original ReLU-based architecture is inherently sparse and that a substantial amount of unnecessary computation can be eliminated through neuromorphic.

Table \ref{tab:sparsity_results} compares two activation sparsity regularization strategies, namely the $L_1$ activation loss and the Hoyer sparsity loss, over varying values of $\gamma$ while holding $\alpha=1$. Both approaches successfully reduce network spiking relative to the baseline model, resulting in more computationally efficient inference. As expected, increasing $\gamma$ decreases the spiking activity while increasing the reconstruction error. However, the manner in which this tradeoff occurs differs significantly between the two regularization strategies.

For the $L_1$ activation loss, the reduction in spiking occurs gradually while the average relative $L_2$ error remains below approximately 1\% until $\gamma=0.005$. At this point, the reconstruction error increases abruptly to over 7\% while the average spiking percentage remains approximately 13\%. Examination of the component-wise spiking statistics in Table \ref{tab:spiking_percentages} explains this behavior. The baseline model exhibits the largest spiking percentages within the trunk and combined sub-networks, indicating that these components perform the majority of the informative computation. The $L_1$ loss primarily suppresses activity within the branch networks while leaving the trunk and combined networks largely unaffected. Eventually, both branch networks approach zero spiking activity, effectively collapsing into constant bias terms because only the final layer bias remains active. Consequently, the operator loses its ability to communicate boundary-condition information through the branch pathways, resulting in the dramatic increase in reconstruction error observed at $\gamma=0.005$.

In contrast, the Hoyer sparsity loss exhibits a considerably more favorable sparsity-accuracy tradeoff. As $\gamma$ increases from $10^{-4}$ to $0.005$, the average relative L2 error increases gradually from approximately 1\% to approximately 5\%, while the average spiking percentage decreases to below 5\%. Notably, the Hoyer regularization simultaneously achieves lower spiking activity and lower reconstruction error than the $L_1$ formulation at $\gamma=0.005$ especially. Furthermore, the Hoyer loss substantially reduces activity within the trunk and combined sub-networks, decreasing their spiking percentages by more than 20\% across the explored regularization strengths. This behavior explains why the $L_1$ loss maintains low reconstruction error for smaller $\gamma$ values; although the branch sub-networks are heavily suppressed, the majority of the information continues to be transmitted through the trunk and combined sub-networks. However, this also makes the $L_1$ regularization more difficult to tune since the computationally dominant sub-networks remain relatively active.

The superior behavior of the Hoyer loss can be understood by considering the objective that it optimizes. Unlike the $L_1$ norm, which penalizes only the overall activation magnitude, the Hoyer loss explicitly promotes sparse activation patterns. For example, the activation vectors $[0.5,0.5,0.5,0.5]$ and $[1,0,1,0]$ both have an $L_1$ norm equal to 2 despite the latter producing significantly fewer spikes. Under the Hoyer metric, these vectors have values of 4 and 2, respectively, thereby favoring the genuinely sparse representation. Although the Hoyer loss produces slightly larger reconstruction errors for some operating points, it provides significantly better control over the energy-accuracy tradeoff. Since the acceptable reconstruction error is ultimately application dependent, controllable sparsity/spiking dynamics are more valuable than simply minimizing $L_2$ error. Furthermore, as demonstrated later through synthetic distillation, the Hoyer-based models can recover additional accuracy while maintaining similarly low spiking percentages.

Additional evidence for the superiority of the Hoyer formulation is obtained by independently weighting the activation penalties for each NOMAD component. Rather than assigning an identical coefficient of $\gamma=5\times10^{-4}$ to every sub-network, the trunk and combined sub-networks were weighted using $3\gamma$, while the branch sub-networks were assigned $\gamma/3$ to more aggressively suppress the dominant communication pathways. Under this modified weighting, the $L_1$ formulation reduced trunk and combined spiking by only 15\% and 7\%, respectively. In contrast, the Hoyer formulation reduced trunk and combined spiking by 45\% and 58\%, respectively. These results further demonstrate that the Hoyer loss directly targets sparsity rather than simply reducing activation magnitudes.

In Figure \ref{fig:2d_hx}c, we see the 50th percentile performance of SAR-NOMAD with the Hoyer loss and $\gamma = 0.005$. The main vortices caused by the 3D geometry's wavy tape insert are shown in high detail. The absolute error is mostly stochastic in nautre with only small hotspots located on the wall surface. This visualization shows that with SAR-based layers, we can still accurately simulate the governing physics while providing highly efficient computation. 

\begin{table*}[t]
\centering
\caption{Per-layer spiking percentages for each network component for Sparse-NOMAD and VS-NOMAD with the 2D Heat Exchanger.}
\makebox[\textwidth][c]{%
\begin{tabular}{c|c c c c}
\hline
Spiking Model & Trunk (\%) & Branch 1 (\%) & Branch 2 (\%) & Comb (\%)\\
\hline

Base Model &
$[47.97, 40.92, 38.27]$ &
$[15.61, 17.56, 14.92, 26.94]$ &
$[49.35, 6.34, 11.79, 24.04]$ &
$[35.23, 39.34, 48.14]$ \\
\hline
SAR $L_1$ $\gamma=0.0005$ &
$[46.50, 18.85, 33.39]$ &
$[0., 0., 0., 0.39]$ &
$[4.75, 0.81, 7.97, 9.00]$ &
$[32.02, 38.47, 49.50]$ \\

SAR $L_1$ $\gamma=0.0005$* &
$[[44.13, 8.78, 30.90]$ &
$[0., 0., 0.78, 0.39]$ &
$[12.44, 0.70, 14.42, 5.83]$ &
$[30.29, 36.19, 44.91]$ \\

SAR $L_1$ $\gamma=0.001$ &
$[45.20, 13.99, 31.80]$ &
$[0., 0., 0., 0.39]$ &
$[2.76, 0.63, 5.51, 8.60]$ &
$[31.69, 36.10, 44.41]$ \\

SAR $L_1$ $\gamma=0.0015$ &
$[44.23, 9.36, 28.35]$ &
$[0., 0., 0., 0.]$ &
$[1.27, 0.96, 4.61, 7.93]$ &
$[31.03, 36.67, 41.66]$ \\

SAR $L_1$ $\gamma=0.005$ &
$[36.86, 2.01, 23.37]$ &
$[0., 0., 0., 0.]$ &
$[0.05, 0., 0., 0.]$ &
$[29.29, 31.34, 31.72]$ \\
\hline
SAR Hoyer $\gamma=0.0001$ &
$[47.42, 28.44, 29.50]$ &
$[0.79, 1.18, 0.67, 1.43]$ &
$[14.13, 2.41, 2.40, 3.64]$ &
$[28.42, 28.35, 33.78]$ \\

SAR Hoyer $\gamma=0.0005$ &
$[43.22, 13.14, 16.26]$ &
$[0.95, 0.78, 0.90, 1.56]$ &
$[5.61, 2.18, 1.62, 2.28]$ &
$[16.39, 20.74, 22.24]$ \\

SAR Hoyer $\gamma=0.0005$* &
$[29.11, 3.03, 8.02]$ &
$[8.55, 3.90, 3.12, 3.52]$ &
$[5.54, 2.15, 1.24, 2.42]$ &
$[5.13, 9.62, 10.02]$ \\

SAR Hoyer $\gamma=0.001$ &
$[37.18, 6.25, 10.00]$ &
$[2.73, 1.56, 1.56, 1.56]$ &
$[2.52, 1.11, 1.85, 2.82]$ &
$[11.35, 13.06, 17.86]$ \\

SAR Hoyer $\gamma=0.005$ &
$[17.85, 3.89, 5.00]$ &
$[0.82, 0.78, 0.39, 0.39]$ &
$[1.02, 0.39, 0.39, 1.51]$ &
$[8.49, 9.56, 9.41]$ \\

\hline


VS Arctan. 1 STS $\gamma=0$ &
$[33.65, 9.78, 7.80]$ &
$[21.96, 8.46, 17.39, 18.75]$ &
$[34.85, 14.13, 7.04, 2.19]$ &
$[4.88, 5.97, 11.34]$ \\
VS Arctan. 1 STS $\gamma=10^{-6}$ &
$[32.35, 5.11, 1.57]$ &
$[0., 0., 3.52, 1.56]$ &
$[42.39, 6.64, 2.34, 1.17]]$ &
$[0.74, 3.89, 2.73]$ \\
VS Arctan. 10 STS $\gamma=0$ &
$[34.98, 10.53, 16.67]$ &
$[29.67, 15.45, 7.03, 3.91]$ &
$[54.19, 9.84, 6.67, 2.14]$ &
$[4.45, 8.42, 10.19]$ \\
VS Arctan. 10 STS $\gamma=10^{-6}$ &
$[37.73, 12.08, 8.24]$ &
$[23.05, 12.50, 7.42, 2.73]$ &
$[49.17, 7.38, 8.59, 1.17]$ &
$[0.91, 5.25, 3.33]$ \\
VS Arctan. 20 STS $\gamma=0$ &
$[36.42, 12.90, 14.58]$ &
$[15.29, 9.06, 2.73, 3.12]$ &
$[43.78, 12.61, 8.47, 1.76]$ &
$[5.39, 8.73, 11.48]$ \\
VS Arctan. 20 STS $\gamma=10^{-6}$ &
$[37.69, 12.83, 9.67]$ &
$[10.55, 7.42, 2.34, 3.12]$ &
$[40.53, 8.50, 6.58, 2.13]$ &
$[1.19, 4.94, 3.85]$ \\
\hline
\end{tabular}}
\label{tab:spiking_percentages}
\caption*{\textit{*Training loss utilized 3$\gamma$ for Trunk and Comb terms and $\gamma$/3 for Branch terms}}
\end{table*}

Table \ref{tab:spiking_percentages} provides the layer-wise spiking percentages for the SAR-NOMAD models and the 2D Heat Exchanger. For the trunk network and the second branch network, which process the spatial coordinates and inlet boundary conditions, the largest spiking activity generally occurs within the first and final hidden layers. The elevated activity in the first layer is expected because this layer lifts a low-dimensional input into the higher-dimensional latent representation, requiring denser communication to preserve representational capacity. Increased activity in the final layer, which is also observed within the combined network, may indicate progressively stronger feature selection as the network approaches the output representation. Although this interpretation remains speculative, it suggests that different layers may exhibit distinct functional roles within the sparse operator and warrants further investigation.

More importantly, the layer-wise analysis highlights the tendency of the $L_1$ regularization to completely suppress the branch sub-networks. At $\gamma=0.005$, both branches contain layers with zero spiking activity, causing each branch to collapse into a constant bias term independent of the input boundary conditions. In contrast, the Hoyer regularization maintains nonzero activity throughout the branch sub-networks. The minimum observed spiking percentage of approximately 0.39\% corresponds to roughly one active neuron out of 256, allowing boundary-condition information to continue propagating through the network while substantially reducing computational cost. This preservation of sparse yet meaningful communication provides another explanation for the improved accuracy obtained using the Hoyer formulation.

\begin{table*}[t]
\centering
\caption{Average fractional feature-spiking entropy for each evaluation node for each network component of Sparse-NOMAD and VS-NOMAD with the 2D Heat Exchanger.}
\makebox[\textwidth][c]{\begin{tabular}{c|c|c|c|c}
\hline
Spiking Model & Trunk & Branch 1 & Branch 2 & Comb \\
\hline

Base Model &
$[0.87, 0.84, 0.83]$ &
$[0.67, 0.69, 0.67, 0.76]$ &
$[0.97, 0.66, 0.75, 0.80]$ &
$[0.81, 0.83, 0.87]$ \\
\hline
SAR $L_1$ $\gamma=0.0005$ &
$[0.86, 0.69, 0.80]$ &
$[0., 0., 0., 0.]$ &
$[0.59, 0.20, 0.61, 0.62]$ &
$[0.80, 0.83, 0.88]$ \\

SAR $L_1$ $\gamma=0.0005$* &
$[0.85, 0.55, 0.79]$ &
$[0., 0., 0.13, 0.]$ &
$[0.75, 0.24, 0.70, 0.58]$ &
$[0.79, 0.82, 0.86]$ \\

SAR $L_1$ $\gamma=0.001$ &
$[0.86, 0.64, 0.80]$ &
$[0., 0., 0., 0.]$ &
$[0.48, 0.16, 0.52, 0.60]$ &
$[0.80, 0.82, 0.86]$ \\

SAR $L_1$ $\gamma=0.0015$ &
$[0.85, 0.56, 0.77]$ &
$[0., 0., 0., 0.]$ &
$[0.36, 0.19, 0.49, 0.57]$ &
$[0.79, 0.82, 0.85]$ \\

SAR $L_1$ $\gamma=0.005$ &
$[0.82, 0.29, 0.73]$ &
$[0., 0., 0., 0.]$ &
$[0.64, 0., 0., 0.]$ &
$[0.78, 0.79, 0.79]$ \\
\hline
SAR Hoyer $\gamma=0.0001$ &
$[0.87, 0.77, 0.78]$ &
$[0.13, 0.20, 0.17, 0.28]$ &
$[0.77, 0.42, 0.41, 0.52]$ &
$[0.77, 0.77, 0.81]$ \\

SAR Hoyer $\gamma=0.0005$ &
$[0.85, 0.63, 0.67]$ &
$[0.19, 0.13, 0.18, 0.25]$ &
$[0.61, 0.35, 0.33, 0.37]$ &
$[0.67, 0.72, 0.73]$ \\

SAR Hoyer $\gamma=0.0005$* &
$[0.78, 0.36, 0.53]$ &
$[0.56, 0.42, 0.38, 0.40]$ &
$[0.58, 0.32, 0.24, 0.35]$ &
$[0.46, 0.58, 0.59]$ \\

SAR Hoyer $\gamma=0.001$ &
$[0.82, 0.49, 0.58]$ &
$[0.36, 0.25, 0.25, 0.25]$ &
$[0.47, 0.25, 0.37, 0.41]$ &
$[0.61, 0.63, 0.69]$ \\

SAR Hoyer $\gamma=0.005$ &
$[0.69, 0.41, 0.45]$ &
$[0.16, 0.13, 0., 0.]$ &
$[0.32, 0.03, 0., 0.25]$ &
$[0.56, 0.57, 0.58]$\\

\hline
VS Arctan. 1 STS $\gamma=0$ &
$[0.80, 0.58, 0.54]$ &
$[0.74, 0.57, 0.69, 0.70]$ &
$[0.90, 0.65, 0.52, 0.32]$ &
$[0.46, 0.49, 0.61]$ \\
VS Arctan. 1 STS $\gamma=10^{-6}$ &
$[0.80, 0.46, 0.24]$ &
$[0., 0., 0.40, 0.25]$ &
$[0.87, 0.51, 0.32, 0.20]$ &
$[0.11, 0.41, 0.35]$ \\
VS Arctan. 10 STS $\gamma=0$ &
$[0.82, 0.60, 0.69]$ &
$[0.78, 0.67, 0.52, 0.42]$ &
$[0.97, 0.60, 0.51, 0.32]$ &
$[0.45, 0.57, 0.59]$ \\
VS Arctan. 10 STS $\gamma=10^{-6}$ &
$[0.83, 0.62, 0.53]$ &
$[0.74, 0.63, 0.53, 0.35]$ &
$[0.93, 0.53, 0.56, 0.20]$ &
$[0.20, 0.49, 0.41]$ \\
VS Arctan. 20 STS $\gamma=0$ &
$[0.82, 0.60, 0.69]$ &
$[0.78, 0.67, 0.52, 0.42]$ &
$[0.95, 0.63, 0.57, 0.31]$ &
$[0.48, 0.57, 0.62]$ \\
VS Arctan. 20 STS $\gamma=10^{-6}$ &
$[0.84, 0.63, 0.58]$ &
$[0.59, 0.53, 0.32, 0.38]$ &
$[0.93, 0.57, 0.54, 0.31]$ &
$[0.27, 0.48, 0.44]$ \\
\hline
\end{tabular}}
\label{tab:entropy_vectors}
\caption*{\textit{*Training loss utilized 3$\gamma$ for Trunk and Comb terms and $\gamma$/3 for Branch terms}}
\end{table*}

Finally, Table \ref{tab:entropy_vectors} presents an entropy-based analysis of feature utilization during spiking inference on the 2D Heat Exchanger. For each layer, spike counts are accumulated across the entire test dataset to produce a feature-wise spike count vector where for each feature we have the number of spikes that this feature neuron emits. Normalizing by the total number of spikes along the feature vector yields a probability distribution,

\begin{equation}
p_i^\ell=\frac{s_i^\ell}{\sum_{j=1}^{d}s_j^\ell},
\end{equation}

where $s_i$ denotes the total spike count of feature $i$ in layer $\ell$ and $d$ is the feature dimension for layer $\ell$. The corresponding spiking entropy for layer $\ell$ is

\begin{equation}
H^\ell=-\sum_{i=1}^{d}p_i^\ell\log(p_i^\ell),
\end{equation}

which is normalized by the maximum possible entropy,

\begin{equation}
H^\ell_{\mathrm{norm}}=\frac{H^\ell}{\log(d)}.
\end{equation}

For the trunk and combined sub-networks, the normalized entropy $H^\ell$ for a feature vector in layer $\ell$ is also averaged over all node points evaluated in that layer (3,977 for the 2D Heat Exchanger). A normalized entropy approaching one indicates that communication is distributed across nearly the entire feature space, whereas values approaching zero indicate that the network communicates through only a small subset of intermediate latent features.

\begin{figure}[htbp]
    \centering
    \includegraphics[width=1\textwidth]{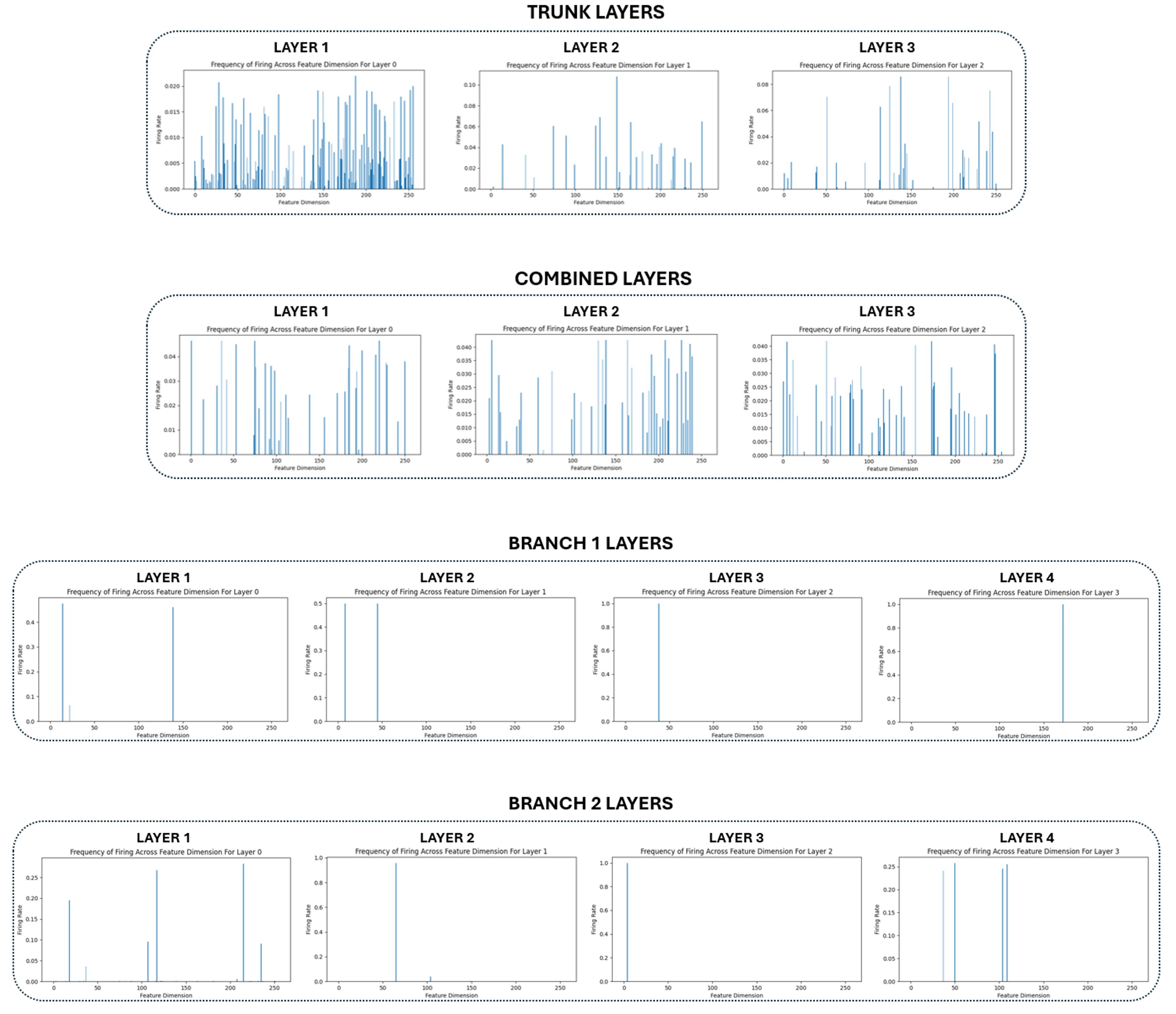}
    \caption{
    \textbf{Spiking Activity Results} 
    We depict the probability/frequency of different feature dimensions spiking for the SAR-NOMAD model with the Hoyer loss and $\gamma = 0.005$. For the branch layers, the sparsity regularization forces only a small subset, as a low as one dimension, to fire and communication information, indicating that the model is forced to collapse it feature dimensions in order to provide low spiking. The trunk and combined subnetworks have a lot more activity across the entire dimension.
    }
    \label{fig:sparsity_entropy}
\end{figure}

The entropy analysis reveals a consistent trend across all SAR-NOMAD models. As $\gamma$ increases and the overall spiking percentage decreases, the normalized entropy also decreases, indicating that the model increasingly relies on a smaller subset of latent features for communication. This behavior is particularly evident within the first branch network. Under $L_1$ regularization, the entropy rapidly decreases to zero as the branch sub-networks completely collapse, particularly at $\gamma=0.005$, where both branches lose all expressive capacity. Conversely, the Hoyer formulation generally maintains nonzero entropy within the branch sub-networks, with occasional zero entries only at $\gamma=0.005$, corresponding to communication through a single active feature. This behavior again aligns with the improved reconstruction accuracy observed for the Hoyer models. The entropy of the Hoyer results still show a collapse of layer dimensions to a small subset with the branch networks further indicated by the spiking activity results in Figure \ref{fig:sparsity_entropy} for the Hoyer run with $\gamma=0.005$. 

The entropy analysis presents both opportunities and challenges. Low entropy suggests considerable potential for structured network pruning since many latent features, and as a result synapse weights, are rarely utilized. In fact, for the branch networks in the Hoyer results could easily be eliminated to the single feature and single vector of synapse weights utilized based off our test run, saving a lot of memory. However, it also indicates that the model may be minimizing spike activity by reducing the effective latent dimensionality rather than learning richer threshold-based dynamics distributed across the full feature space. Although preliminary experiments explored entropy as an additional regularization objective, limited improvements were observed, indicating that further investigation is necessary.

\paragraph{VS/LIF-NOMAD Results:} In addition to SAR-NOMAD, we investigate the VS-NOMAD and LIF-NOMAD architectures using the spiking percentage loss introduced in \cite{garg2023neuroscienceinspiredscientificmachinepart2}. Two surrogate gradient functions, fast sigmoid and arctangent, are evaluated together for the VSN implementation with spike time step (STS) values of 1, 10, and 20 and regularization coefficients of $\gamma=0$ and $10^{-6}$ to assess the effect of introducing temporal dynamics into VS-NOMAD. For the LIF-NOMAD model, we explore the arctangent function with direct encoding for 1, 10, and 20 STS and rate encoding for 10 and 20 STS.

As expected, increasing $\gamma$ reduces the average spiking percentage (except for 20 STS LIF-NOMAD with direct encoding) while increasing the reconstruction error, as summarized in Table \ref{tab:sparsity_results}. A clear distinction emerges between the surrogate gradient functions for the VSN implementation. Fast sigmoid produces lower average spiking percentages, ranging from approximately 6-12\%, but suffers from substantially larger mean reconstruction errors ranging from 20-46\%. Conversely, arctangent produces significantly lower reconstruction errors between approximately 8-35\% while exhibiting higher spiking percentages between approximately 7.5-15\%. Although fast sigmoid achieves greater computational efficiency, its severe degradation in reconstruction accuracy makes arctangent the preferable surrogate function, particularly since it still achieves substantially lower spiking activity than the baseline NOMAD model. Comparing VS-NOMAD and LIF-NOMAD, we find that the rate-encoded implementations with reduced precision for input and neuron-neuron communication for LIF-NOMAD resulted in severe error degradation compared to the far better reconstruction performance with the VS-NOMAD with mean $L_2$ errors over 67\% even though the equivalent spiking percentages are reduced. For example, VS-NOMAD with arctangent for 20 STS and $\gamma=10^{-6}$ does have around 1.7 times more spiking that the most efficient rate-encoded LIF implementation but over 5 times improved $L_2$ error. Contrastingly, the results with the LIF-NOMAD model utilizing direct encoding indicate a different story due to the improved precision of the model input. Although VS-NOMAD with $\gamma = 0$ for 10 and 20 STS resulted with lower $L_2$ error, the overall performance of the LIF-NOMAD with direct encoding is at least similar to arctangent VS-NOMAD with similar error and similar or lower spiking percentage. Directly encoded LIF-NOMAD presents no clear performance distinction from VS-NOMAD unlike its rate-encoded counterpart which utilizes reduced precision in its input.

Increasing the number of spike time steps generally improves reconstruction accuracy (not for LIF-NOMAD, directly encoded for 10 to 20 STS), suggesting that temporal dynamics enrich the latent representation and improve the reconstruction of the underlying physics. However, the average spiking percentage also increases because additional spike time steps allow greater membrane potential accumulation and consequently more spike events. With inference on an NVIDIA A100, we find that 1, 10, 20, and 100 STS result in an average latency of 7, 57, 114, and 556 milliseconds (approximately linear scaling). With applications requiring real time performance and potentially multiple auto-regressive calculations for long-term forecasting, a low STS with lower latency might be desirable in terms of edge-deployment and integration of energy-efficient neural operators. In addition, for the 2D Heat Exchanger, we were unable to train an STS of 30 due to memory constraints for a single H200, further demonstrating the difficulty of dealing with high spike step dynamics and the benefits of reducing spiking to the single step regime.

Despite the performance of LIF and VSN implementations for NOMAD, the Hoyer/$L_1$-based SAR-NOMAD provides significantly improved reconstruction and efficiency results over the surrogate gradient alternatives. The best-reconstruction-performing VS/LIF configuration employs the arctangent surrogate and VSN with ten spike time steps and $\gamma=0$, producing approximately 8\% average relative $L_2$ error with roughly 15\% average spiking. The lowest spiking VS/LIF result is the 10 STS LIF-NOMAD with rate encoding which presented with 73.39\% average $L_2$ error and 6.41\% spiking. In comparison, the Hoyer-based SAR-NOMAD model at $\gamma=0.005$ achieves approximately 5.41\% average relative $L_2$ error while requiring only approximately 4.87\% average spiking and only a single spike step. Not only does the SAR based model perform with better power consumption and reconstruction error than all VS/LIF-NOMAD results, it also significantly reduces the latency that might be required by VS/LIF-NOMAD to achieve similar performance. This performance gap is likely attributable to the surrogate gradient mismatch that commonly limits the optimization of spiking neural operators. In conclusion, SAR-NOMAD provides an efficient alternative that avoids the long spike trains and surrogate gradient training while providing desired energy consumption performance. SAR-NOMAD is guaranteed to be the closest to the desired speed for real-time monitoring with only a single spike step, making it an ideal candidate for efficient virtual sensing and a benchmark towards proposed spiking models in the future.

An additional observation is that VS-NOMAD primarily suppresses activity within the first branch network and the combined network rather than within the second branch network like SAR-NOMAD and LIF-NOMAD. Since the combined network integrates information from all preceding sub-networks before projecting to the nonlinear output manifold, excessive suppression of this component may explain the larger degradation in reconstruction accuracy compared to SAR-NOMAD and LIF-NOMAD with direct encoding.

Table \ref{tab:spiking_percentages} further illustrates the layer-wise spiking activity for VS-NOMAD using the arctangent surrogate gradient, the best performing function. Similar to SAR-NOMAD, the highest activity occurs within the first hidden layer of both branch networks and the trunk network due to the lifting of low-dimensional inputs into the latent space. However, unlike the Hoyer-based SAR-NOMAD models, the VS-NOMAD architecture permits complete collapse of the first branch network for the one-STS configuration with $\gamma=10^{-6}$, where zero spiking is observed within the first hidden layer for fast sigmoid and arctangent. Consequently, this branch reduces to a constant term independent of the heat profile input. The Hoyer-based SAR-NOMAD models, which have no gradient mismatch avoid this collapse completely.

Finally, Table \ref{tab:entropy_vectors} presents the same entropy-based analysis for arctangent VS-NOMAD. Consistent with the SAR-NOMAD observations, reduced spiking activity corresponds to lower normalized entropy, indicating that the network communicates through a progressively smaller subset of latent features as spike activity is suppressed. Although entropy for VS-NOMAD is higher than the SAR models, the spiking percentages of the variable spiking results is higher. All indications show that with lower spiking both VS-NOMAD and SAR-NOMAD tend to reduce entropy and reduce the scale of the feature dimensions utilized for inference.

Overall, the results for the 2D Heat Exchanger strongly demonstrate the potential for SAR-based spiking as an alternative for surrogate gradient trained VSN/LIF operators. Not only does it provide improved accuracy and spiking performance, the ReLU layer also removes the need for high STS and keeps latency down closer to the desired real-time properties for virtual sensing applications. At the very least, as improvements in training for traditional spiking neural operators occur, the SAR layer can provide a gold-standard benchmark in terms of efficiency, latency, and accuracy reconstruction.

\subsection{Heat Exchanger Performance (VIRSO)}

\begin{table*}[t]
\centering
\caption{$L_2$ errors and spiking percentages between the spectral-only SAR-GNO and spectral-only VS/LIF-GNO with the 2D Heat Exchanger \cite{howes2026neuroscienceinspiredgraphoperators}. (Thresh.) corresponds to the inclusion of the threshold parameter for SAR and an additional GeLU layer. (No Thresh.) corresponds to removing the threshold and keeping the GeLU layer while (Thresh. No GeLU) represents the removal of the GeLU function while keeping the threshold.}
\makebox[\textwidth][c]{\begin{tabular}{c|c|ccccccc}
\hline
  & & \multicolumn{7}{c}{Mean Spiking Percentages (\%)} \\
Spiking Model
& Mean Rel. $L_2$ (\%)
& M
& P
& Spectral
& Spatial
& f
& Q
& Mean\\
\hline
Arctan. LIF-GNO $\gamma=0$ DE 1 STS & 10.93 & 25.51 & 31.87 & 21.52 & - & - & 14.85 & 23.44\\
Arctan. LIF-GNO $\gamma=0.5$ DE 1 STS & 10.85 & 0. & 1.17 & 3.37 & - & - & 9.19 & 3.43\\
Arctan. LIF-GNO $\gamma=0.5$ DE 10 STS & 9.71 & 0. & 1.17 & 3.80 & - & - & 8.42 & 3.35\\
Arctan. LIF-GNO $\gamma=0$ RE 10 STS & 21.00 & 7.81 & 6.44 & 18.47 & - & - & 11.35 & 11.01\\
Arctan. LIF-GNO $\gamma=0.5$ RE 10 STS & 16.39 & 0. & 0. & 0.94 & - & - & 6.54 & 1.87\\
\hline
Fast Sig. VS-GNO $\gamma=0$ 1 STS& 0.40 & 10.53 & 14.64 & 25.69 & - & - & 15.84 & 16.68\\
Fast Sig. VS-GNO $\gamma=0.5$ 1 STS& 0.49 & 0.78 & 2.17 & 20.06 & - & - & 8.81 & 7.96\\
Fast Sig. VS-GNO $\gamma=0.5$ 10 STS& 0.45 & 2.34 & 1.70 & 20.73 & - & - & 9.47 & 8.56\\
\hline
Arctan. VS-GNO $\gamma=0$ 1 STS & 0.36 & 8.64 & 9.41 & 27.42 & - & - & 13.03 & 14.62\\
Arctan. VS-GNO $\gamma=0.5$ 1 STS & 0.47 & 0.39 & 3.75 & 13.10 & - & - & 4.62 & 5.47\\
Arctan. VS-GNO $\gamma=0.5$ 10 STS & 0.48 & 1.56 & 3.30 & 13.98 & - & - & 4.67 & 5.88\\
\hline
SAR-GNO $\gamma=0.001$ (Thresh.) & \textbf{0.45} & 3.95 & 8.85 & 11.04 & - & - & 24.54 & \textbf{12.10}\\
SAR-GNO $\gamma=0.003$ (Thresh.) & \textbf{0.54} & 0.39 & 1.57 & 6.67 & - & - & 16.24 & \textbf{6.22}\\
\hline
SAR-GNO $\gamma=0.001$ (No Thresh.) & 0.56 & 4.18 & 7.45 & 16.50 & - & - & 24.57 & 13.18\\
SAR-GNO $\gamma=0.003$ (No Thresh.) & 6.32 & 0. & 0. & 8.42 & - & - & 15.72 & 6.03\\
\hline
SAR-GNO $\gamma=0.001$ (Thresh. No GeLU) & 0.41 & 3.17 & 11.62 & 13.53 & - & - & 30.20 & 14.63\\
SAR-GNO $\gamma=0.003$ (Thresh. No GeLU) & 0.46 & 2.55 & 4.56 & 9.51 & - & - & 24.21 & 10.21\\
\hline
\end{tabular}}
\label{tab:sparsity_results_graph_spectral}
\end{table*}

\begin{table*}[t]
\centering
\caption{$L_2$ errors and spiking percentages between the full SAR-GNO architecture and the Full VS/LIF-GNO architecture with the 2D Heat Exchanger \cite{howes2026neuroscienceinspiredgraphoperators}. (Thresh.) corresponds to the inclusion of the threshold parameter for SAR and an additional GeLU layer. (No Thresh.) corresponds to removing the threshold and keeping the GeLU layer while (Thresh. No GeLU) represents the removal of the GeLU function while keeping the threshold.}
\makebox[\textwidth][c]{\begin{tabular}{c|c|ccccccc}
\hline
  & & \multicolumn{7}{c}{Mean Spiking Percentages (\%)} \\
Spiking Model
& Mean Rel. $L_2$ (\%)
& M
& P
& Spectral
& Spatial
& f
& Q
& Mean\\
\hline
Arctan. LIF-GNO $\gamma=0$ DE 1 STS & 14.80 & 8.65 & 13.92 & 22.27 & 18.84 & 242.04 & 15.04 & 53.46\\
Arctan. LIF-GNO $\gamma=0.5$ DE 1 STS & 17.07 & 0.76 & 2.98 & 20.78 & 4.25 & 71.40 & 9.14 & 18.22\\
\hline
Fast Sig. VS-GNO $\gamma=0$ 1 STS& 1.09 & 22.52 & 16.44 & 25.04 & 38.39 & 14.91 & 30.24 & 24.59\\
Fast Sig. VS-GNO $\gamma=0.5$ 1 STS & 0.83 & 0.39 & 5.68 & 26.67 & 2.69 & 12.16 & 11.35 & 9.82\\
\hline
Arctan. VS-GNO $\gamma=0$ 1 STS & 2.32 & 28.90 & 34.37 & 21.80 & 19.62 & 21.33 & 13.47 & 23.25\\
Arctan. VS-GNO $\gamma=0.5$ 1 STS & 5.72 & 13.72 & 23.29 & 25.19 & 9.02 & 8.92 & 10.74 & 15.15\\
\hline
SAR-GNO $\gamma=0.01$ (Thresh.) & \textbf{1.17} & 0.83 & 2.14 & 4.73 & 0.60 & 13.18 & 17.48 & \textbf{6.49} \\
SAR-GNO $\gamma=0.05$ (Thresh.) & \textbf{4.05} & 0.39 & 1.07 & 1.59 & 0.70 & 3.56 & 7.53 & \textbf{2.47} \\
\hline
SAR-GNO $\gamma=0.01$ (No Thresh.) & 1.08 & 0.78 & 1.00 & 9.64 & 1.00 & 15.68 & 20.03 & 8.02 \\
SAR-GNO $\gamma=0.05$ (No Thresh.) & 2.35 & 0.39 & 1.00 & 3.21 & 1.00 & 6.37 & 13.51 & 4.25 \\
\hline
SAR-GNO $\gamma=0.01$ (Thresh. No GeLU) & 0.95 & 1.55 & 1.52 & 6.63 & 0.50 & 15.32 & 23.72 & 8.21\\
SAR-GNO $\gamma=0.05$ (Thresh. No GeLU) & 3.23 & 0.39 & 1.00 & 2.31 & 0.80 & 4.57 & 13.78 & 3.81 \\
\hline
\end{tabular}}
\label{tab:sparsity_results_graph_full}
\end{table*}

Following the success of SAR layer within NOMAD, we extend the approach to the graph-based VIRSO architecture by inserting SAR layers followed by GeLU activations, consistent with the original VIRSO design, resulting in the proposed SAR-GNO architecture. We evaluate SAR-GNO on the 2D Heat Exchanger benchmark and directly compare its performance against the previously proposed VS-GNO using the VSN formulation \cite{howes2026neuroscienceinspiredgraphoperators} and LIF-GNO which utilizes the traditional LIF neuron. Consistent with the VS-GNO study, both the spectral-only and full VIRSO architectures are investigated for VSN, LIF, and SAR implementations. Because SAR-NOMAD existed with already built in bias terms within its linear layers, we chose not to include the optional threshold $\boldsymbol{\tau}$. SAR-GNO has non-linear layers with normalization blocks such as the output of the spectral block and the spatial aggregation. With those layers, we decided to add the threshold which we test its performance in this section.

Tables \ref{tab:sparsity_results_graph_spectral} and \ref{tab:sparsity_results_graph_full} summarize the average relative L2 error across all output channels together with the average spiking percentages for each component of the original VIRSO architecture for the spectral-only and full version of LIF/VS-GNO and SAR-GNO. For the spectral-only architecture (Table \ref{tab:sparsity_results_graph_spectral}), the previously proposed VS-GNO achieved strong performance compared to LIF-GNO, obtaining average relative $L_2$ errors below 0.5\% while reducing the average spiking percentage to approximately 5.5\% for the arctangent implementation and 8\% for fast sigmoid. For the graph architecture, the VSN demonstrates far improved performance over the LIF implementation unlike the VS-NOMAD results. The LIF-GNO did achieve lower spiking with the 1 STS directly encoded model with $\gamma=0.5$ having 3.43\% spiking and the 10 STS, $\gamma=0.5$ rate encoded model having 1.87\% spiking, but the error is over 20 times higher, representing severe performance degradation. The VSN, with its variable communication most likely naturally aligns with the hybrid SNN/ANN structure of VS-GNO, provides far better reconstruction performance while also still significantly reducing spiking albeit potentially not as low as the LIF implementations although the 1.7 correction factor is not as accurate for a hybrid architecture unlike the purely spiking NOMAD model since there exists standard downstream computations that don't have reduced energy between VSN or LIF since they would be identical. The concept of the 1.7 different originated from analysis of how much energy a LIF/VSN neuron needs for computation \cite{garg2023neuroscienceinspiredscientificmachinepart2} which is the most relevant for a completely spiking model such as VS/LIF-NOMAD. Further integration is needed to understand the true difference in energy consumption.

For the full-version of VS-GNO, we see even more improved performance over the LIF-GNO implementation which has, at best, twice as less spiking but over 20 times worse reconstruction performance, strongly indicating the superior communication enrichment from the Variable Spiking Neuron.

With the SAR layer, we see similar or slightly improved performance over the VSN implementation. For the spectral-only architecture, at $\gamma=0.003$, SAR-GNO with a threshold parameter $\boldsymbol{\tau}$ and added GeLU layer achieves an average relative $L_2$ error below 0.6\% while maintaining an average spiking percentage around 6.22\%. This spiking is lower than the fast sigmoid VSN implementations but the SAR-GNO's reconstruction performance is slightly worse (0.05\% higher than the lowest spiking fast sigmoid VSN result), concluding with an overall similar performance between SAR-GNO and VS-GNO. For the arctangent implementation of VS-GNO, we have slightly better $L_2$ (at most 0.18\%) and spiking (at most 0.75\%) than the SAR-GNO implementation. In addition, for both fast sigmoid and arctangent, the 10 STS implementation for VS-GNO resulted in similar $L_2$ error with worse spiking percentages indicating that the 1 STS VS-GNO architecture with no memory temporal dynamics was the preferred option. Comparing both a step SAR layer and the 1 STS VS-GNO, the results potentially indicate that the spectral-only VSN implementation was able to avoid surrogate gradient instability and provided slightly better communication enrichment with the ability to permit negative signals unlike the SAR layer and its dependence on the ReLU function. Despite the comparison, the SAR implementation with the spectral-only architecture is essentially the same, and the reliance on ReLU only provides a slight difference in performance. Although further investigation into sparsity-inducing layers that allow negative signals might be warranted, the magnitude of the improved communication's impact is not clear, and in the context of surrogate gradient functions, might be surpassed by the training instability.

For the full graph architecture, the SAR-GNO provides slightly improved results over VS-GNO. For the fast arctangent implementation of VS-GNO, as gamma increases the spiking is reduced but the mean relative $L_2$ error degrades to over 5\%. Meanwhile, SAR provides a reconstruction performance of at most 4\% with spiking below 3\% and far below the arctangent spiking performance. In contrast, the fast sigmoid VSN implementation, somewhat surprisingly, provides low $L_2$ error with a spiking performance as low as 9.82\% with 0.83\% error, avoiding the same degradation that the arctangent implementation experienced. While SAR-GNO provides spiking below 3\%, the resulting L2 is slightly higher reaching at most 4\%. Despite this, we see that at $\gamma=0.01$ with the threshold and attached GeLU layer, the full SAR-GNO architecture with threshold parameters achieves an average relative $L_2$ error of only 1.17\% with spiking 6.49\% resulting in a combined Error-Energy score (described in more detail in the next section) of 7.59 which is slightly lower than the Error-Energy score of 8.15 for the fast sigmoid VS-GNO. Based on these results, SAR-GNO does seem to provide slightly improved performance over VS-GNO, although the overall reconstruction performance remains relatively close between the two approaches. More importantly, these results demonstrate that the reliance on ReLU within the SAR layer does not severely degrade performance compared to the GeLU activation used with VS-GNO. In contrast, the use of surrogate gradients in VS-GNO can introduce additional instability and training degradation, as demonstrated by the performance degradation observed with the arctangent implementation. Thus, SAR provides a competitive alternative that avoids surrogate-gradient training while maintaining comparable or slightly improved reconstruction performance.

We also examine the impact of the threshold parameter for the full and spectral-only version of SAR-GNO as well as the inclusion of the subsequent GeLU layer in Tables \ref{tab:sparsity_results_graph_spectral} and \ref{tab:sparsity_results_graph_full}. For the spectral-only model, removing the threshold for the $\gamma=0.003$ implementation significantly worsens $L_2$ error with similar spiking performance. We can see that the removal of the threshold parameter allows the spectral spiking to increase forcing the model to collapse the input embedding spiking (M) to zero which hurts performance. Adding a threshold allows the spectral SAR-GNO to better regulate spiking for the spectral layer and avoid the collapse for the input embedding, resulting in an improved optimization and efficiency performance. This is further emphasized with the performance of spectral-only SAR-GNO at $\gamma=0.001$ which displays a less dramatic difference between no threshold and a threshold but still exhibits slightly better spiking and accuracy when the threshold is included because it is able to significantly reduce the spectral layer percentage due to better spiking control. 

For both gamma values explored with the full graph architecture, adding the threshold slightly worsens relative $L_2$ error compared to the non-threshold version, but can also reduce spiking overall. We see that adding threshold brings down the spectral layer spiking from 9.64\% to 4.73\% for $\gamma=0.01$ and 3.21\% to 1.59\% for $\gamma=0.05$. The spatial layer does observe less dramatic behavior and worse performance with the threshold, but these results still indicate how the optional threshold $\boldsymbol{\tau}$ can be used to improve spiking control for certain layers and potentially reduce overall spiking. 

In addition, we explore ablation analysis with the GeLU activation function. We remove the GeLU layer while keeping the threshold parameters for both spectral-only and full architectures. For the first model version, we find that adding the GeLU layer slightly weakens the L2 error performance (around 0.1\% increase for $\gamma=0.003$) but the spiking percentage shows a significant decrease. This result indicates that for the spectral-only model the addition of the GeLU layer does allow for noticeable improvements in neuron communication and training, resulting in almost similar reconstruction with lower spiking. For the full-version, the performance of the GeLU layer demonstrates similar characteristics. When adding the GeLU layer for $\gamma=0.05$, the L2 error increases by around 0.82\% while the spiking decreases by around 1.34\%. For $\gamma=0.01$, the L2 error between GeLU and no-GeLU differs by only 0.22\% while the spiking with the GeLU layer was lower by 1.72\%, indicating that although the GeLU layer can introduce a small increase in reconstruction error, it consistently reduces the required spiking activity. This suggests that the GeLU layer potentially improves the efficiency of neuron communication while maintaining comparable reconstruction performance, further supporting its use after the ReLU thresholding in the SAR layer depicted in Equation \ref{eq:relu_sparse}. Despite these results, the conclusion is not definitive and further benchmarking and analysis is required.

Overall, SAR-GNO further reinforces the potential advantages of sparse activation regularization with ReLU over surrogate-gradient-based variable spiking formulations. By directly regularizing the activation outputs, SAR-GNO effectively suppresses unnecessary computation while preserving information flow throughout the network. Although the SAR layer's performance is not as strong compared to VS-GNO as seen with the NOMAD implementation, our presented sparsity layer provides similar to slightly better reconstruction performance and efficiency, indicating the success of SAR within the context of non-linear layers/architectures. The inclusion of the threshold and subsequent GeLU does show noticeable improvement in spiking performance with only slightly worse or similar $L_2$ error, demonstrating their potential effectiveness when included in the SAR framework introduced in Equation \ref{eq:relu_sparse}. Overall, the reliance of ReLU has not demonstrated severe performance degradation within the context of SAR-GNO.

Finally, it is worth reemphasizing that implementing 10 STS did not drastically change performance while worsening computational efficiency compared to its 1 STS counterpart. Unlike the improvements observed for VS-NOMAD, the introduction of additional temporal dynamics provided little benefit for the graph neural operator architecture, suggesting that temporal spiking dynamics are less effective within the VIRSO framework.

\begin{figure}[htbp]
    \centering
    \includegraphics[width=1\textwidth]{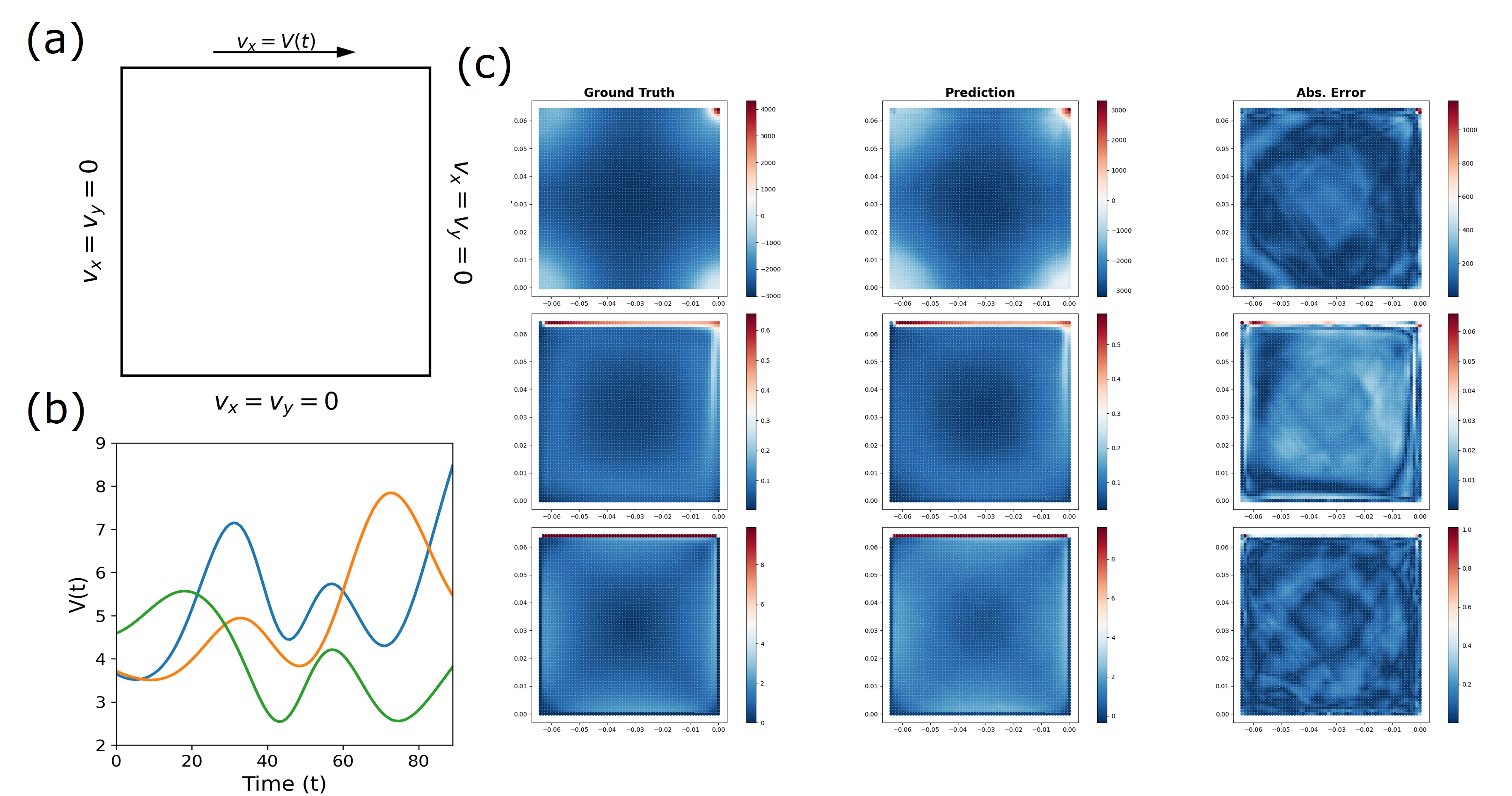}
    \caption{
    \textbf{2D Lid-Driven Cavity} \textbf{(a)} We define the boundary conditions for the Lid-Driven Cavity geometry which shows zero velocity on all faces except for a temporal signal at the top surface. \textbf{(b)} We display example time-varying velocity profiles for the top surface which is defined for 90 time steps. \textbf{(c)} The 50th percentile test example for the SAR-NOMAD model with the Hoyer loss term and $\gamma=0.05$. We see that the main vortex and boundary behavior is closely captured as well as the corner eddies with mostly stochastic absolute error behavior. We do see small hotspots of absolute error located at the top surface and corners indicating some difficulty in capturing surface physics with a regular discretized grid.
    }
    \label{fig:2d_ldc}
\end{figure}

\subsection{Lid Driven Cavity (NOMAD)}

To further demonstrate the improved convergence behavior of SAR-based operators relative to surrogate gradient-based VSN and LIF neurons, both as a potential neuromorphic alternative mechanism and as a benchmark for future development, we evaluated both approaches on the LDC dataset. Consistent with the 2D Heat Exchanger study, Hoyer-based SAR-NOMAD models were trained using an $\alpha$ value of 1 while varying the sparsity coefficient $\gamma$. In parallel, VS-NOMAD models employing both the fast sigmoid and arctangent surrogate gradients were trained across multiple values of $\gamma$ and STS. We also utilized LIF-NOMAD implementations with the arctangent gradient function utilizing direct encoding with 1, 10, 20, and 30 STS and rate encoding with 10, 20, 30 STS. A summary of the resulting performance is provided in Table \ref{tab:sparsity_results_ldc}.

As expected, the baseline model without explicit spiking regularization already exhibits a degree of inherent sparsity across its sub-networks (almost 40\% average spiking). The VS-NOMAD models reduce the average spiking activity to as low as 14.78\% using the arctangent surrogate gradient and 15.02\% using the fast sigmoid surrogate gradient, corresponding to a reduction of more than half of the baseline spiking activity. However, this reduction is accompanied by a substantial degradation in reconstruction accuracy, with the lowest observed mean relative $L_2$ error remaining above 13\%. The spiking behavior of VS-NOMAD is similar to what is observed for the 2D Heat Exchanger dataset. For the arctangent surrogate gradient, increasing $\gamma$ generally produces lower spiking activity. The fast sigmoid also shows this behavior except for the implementation with 10 STS which exhibits increased spiking at $\gamma = 10^{-5}$ compared to lower gamma. Contrastingly, increasing STS does not result in the exact same behavior with the 2D Heat Exchanger. From 20 to 30 STS for the Fast Sigmoid, the L2 error increases from 20.44\% to 31.15\% for $\gamma=10^{-5}$ and 12.86\% to 14.83\% for $\gamma=0$ while increasing or keeping the same spiking percentage. Fast Sigmoid also exhibits similar behavior at different spike time transitions. Although the arctangent function is more aligned with our expectations, the VS-NOMAD results indicate that the spiking dynamic behavior of VS-NOMAD is not guaranteed to improve performance with higher STS potentially due to its high dependence on the surrogate gradient formulation. In addition, we were unable to train an STS of 40 due to memory constraints with a single H200 further indicating the difficulties of training with high spike time neural operators for virtual sensing.

Suprisingly, the LIF-NOMAD results for direct encoding slightly outperformed the VS-NOMAD implementation. Similar to VSN, higher STS improved $L_2$ performance while generally increasing spiking slighlty but the increase in gamma produced less predictable results with 1 STS and 20 STS seeing improved $L_2$ error with higher gamma. Either way, with 30 STS and $\gamma=0$, the directly-encoded LIF-NOMAD had an $L_2$ error of 8.76\% with spiking around 10.32\% which is improved in both reconstruction and effeciency over all VSN implementations. Of course the rate-encoded, with far reduced precision in its input, presents sever error degradation similar to the Heat Exchanger. Overall, the LIF-NOMAD with direct-encoding provides a more competitive output than the VS-NOMAD unlike the VS/LIF-GNO results in the previous section which might have benefited with the variable communication more due to the SNN-ANN hybrid architecture design.

Despite the previous results, the SAR layer with its variable communication outperforms the VSN and LIF implementations. Similar to the Heat Exchanger application, SAR-NOMAD exhibits a strong, consistent, and predictable sparsity-accuracy tradeoff. As $\gamma$ increases from 0.001 to 0.05, the average spiking activity decreases monotonically from approximately 19\% to 4\%, while the mean relative $L_2$ error increases gradually from approximately 1\% to 10\%. Although reconstruction error increases with stronger sparsity regularization, SAR-NOMAD models with $\gamma > 0.005$ outperform every evaluated VS-NOMAD configuration in both reconstruction accuracy and spiking efficiency. In comparsion with LIF-NOMAD, the LIF neuron implemention with direct encoding does present slightly lower L2 error at 30 STS than the $\gamma=0.05$ SAR result but at the cost of higher spiking and significantly higher latency. Overally, SAR-NOMAD provides improved performance over all LIF results (even outperforming the spiking percentages of the rate-encoded model) in terms of error, spiking, and latency. Similar to the results obtained for the 2D Heat Exchanger, these findings further establish SAR-based spiking as a low-latency, high-accuracy alternative to VSN-based networks while simultaneously providing a practical gold-standard benchmark for future spiking neural operator development due to its single spike-step operation. In Figure \ref{fig:2d_ldc}c, we see the 50th percentile performance of SAR-NOMAD with the Hoyer loss term at $\gamma = 0.05$. The main vortex and corner behaviors are captured accurately, further indicating that SAR is capable of providing accurate reconstruction with high energy-efficiency.

Because SAR-NOMAD and VS/LIF-NOMAD share the same underlying network architecture and differ only in their spiking formulation, their performance can be directly compared using a simple Latency-Error-Energy (LEE) metric similar to the energy-delay product in concept but including reconstruction performance: an important pillar of efficient virtual sensing. The proposed LEE score is defined as the product of the latency measured in spike time steps or STS, the mean relative $L_2$ reconstruction error (\%), and the average spiking percentage across all sub-networks (\%), with lower values indicating superior overall performance under the assumption that latency, accuracy, and energy efficiency are equally weighted. To further isolate the tradeoffs between these quantities, three additional metrics are considered: the Latency-Error (LErr), Error-Energy (EE), and Latency-Energy (LEn) scores. While latency and energy consumption could ultimately be expressed in physical units following hardware implementation, the identical network architecture and implementation of both SAR-NOMAD and VS-NOMAD allow STS and average spiking percentage to serve as appropriate proxies for latency and energy consumption in the present software-based evaluation. 

\textbf{It should be noted that for LIF-based architectures the reduced computation made possible by the binary signaling between neurons might improve latency for similar spike step counts, making the presented LEE score limited in its latency comparison between the SAR and LIF layers. Physical hardware installation is required, though limited devices exist currently that can approach variable communication, and as  a result, this direction is being actively explored. We still believe that combined with the improved error and/or spiking, our SAR layer is preferred over LIF and VSN alternatives.} Moreover, although equal weighting is assumed in these metrics, application-specific requirements may prioritize one objective over another. For example, safety-critical applications may favor lower reconstruction error, whereas energy-constrained deployments may prioritize reduced energy consumption.

Using these metrics, Tables \ref{tab:performance_scores_hx} and \ref{tab:performance_scores_ldc} clearly demonstrate that the selected SAR-NOMAD configuration outperforms every evaluated VS/LIF-NOMAD model across all four performance metrics. For the comprehensive LEE score, the selected SAR-NOMAD model achieves a score of 26.35 for the 2D Heat Exchanger, representing at least a fivefold improvement over the best-performing VS/LIF-NOMAD configuration. Similarly, for the Lid Driven Cavity dataset, SAR-NOMAD achieves an LEE score of 37.96, corresponding to an improvement of also more than fivefold compared to the best VS/LIF-NOMAD model. SAR-NOMAD also consistently achieves superior performance across the individual LErr, EE, and LEn metrics, demonstrating improvements not only in reconstruction accuracy but also in computational efficiency while maintaining the low-latency advantages of single-step spiking. Given that real-time prediction, and ultimately faster-than-real-time prediction, is essential for providing timely information to system operators, latency represents a critical performance metric. Consequently, the combination of low latency, high reconstruction accuracy, and improved energy efficiency makes SAR-NOMAD and the SAR layer in general a highly effective candidate for efficient virtual sensing and field reconstruction.

\begin{table*}[t]
\centering
\caption{$L_2$ errors and spiking percentages for Hoyer-based SAR-NOMAD and VS/LIF-NOMAD (Fast Sigmoid and Arctanget surrogate gradients) with the Lid Driven Cavity.}
\makebox[\textwidth][c]{\begin{tabular}{c|cccc|cccc}
\hline
 & \multicolumn{4}{c|}{Mean Relative L2 Errors (\%)} & \multicolumn{4}{c}{Mean Spiking Percentages (\%)} \\
Spiking Model
& $p$
& $v$
& $k$
& Mean
& Trunk
& Branch
& Comb
& Mean \\
\hline
Base Model & 0.89 & 1.69 & 1.05 & 1.21 & 45.35 & 31.41 & 37.92 & 38.23\\
\hline
SAR Hoyer $\gamma=0.001$ & 1.03 & 1.93 & 1.10 & \textbf{1.35} & 26.21 & 7.24 & 24.95 & \textbf{19.47}\\
SAR Hoyer $\gamma=0.005$ & 1.69 & 3.40 & 1.81 & \textbf{2.30} & 16.59 & 3.83 & 16.79 & \textbf{12.40}\\
SAR Hoyer $\gamma=0.01$ & 2.88 & 5.36 & 2.88 & \textbf{3.71} & 13.03 & 2.74 & 12.28 & \textbf{9.35}\\
SAR Hoyer $\gamma=0.05$ & 5.19 & 16.10 & 8.15 & \textbf{9.81} & 5.90 & 0.74 & 4.98 & \textbf{3.87}\\
\hline
VS Fast Sig. 1 STS $\gamma=0$ & 6.65 & 14.02 & 22.56 & 14.41 & 31.45 & 35.72 & 12.63 & 26.60\\
VS Fast Sig. 1 STS $\gamma=10^{-5}$ & 30.02 & 45.16 & 42.82 & 39.33 & 19.55 & 14.98 & 2.67 & 12.40\\
VS Fast Sig. 10 STS $\gamma=0$ & 12.25 & 20.66 & 25.79 & 19.57 & 28.93 & 10.22 & 9.28 & 16.14\\
VS Fast Sig. 10 STS $\gamma=10^{-5}$ & 12.27 & 27.84 & 23.91 & 21.34 & 30.08 & 19.54 & 7.12 & 18.91\\
VS Fast Sig. 20 STS $\gamma=0$ & 6.36 & 15.05 & 17.17 & 12.86 & 33.20 & 16.32 & 10.58 & 20.03\\
VS Fast Sig. 20 STS $\gamma=10^{-5}$ & 13.21 & 23.13 & 24.99 & 20.44 & 30.42 & 22.58 & 7.97 & 20.32\\
VS Fast Sig. 30 STS $\gamma=0$ & 7.95 & 16.57 & 19.97 & 14.83 & 42.40 & 27.33 & 17.26 & 29.00\\
VS Fast Sig. 30 STS $\gamma=10^{-5}$ & 15.54 & 37.28 & 40.64 & 31.15 & 29.92 & 12.14 & 9.29 & 17.12\\
\hline
VS Arctan. 1 STS $\gamma=0$ & 10.05 & 11.52 & 21.28 & 14.28 & 33.89 & 18.08 & 7.96 & 19.98\\
VS Arctan. 1 STS $\gamma=10^{-5}$ & 19.92 & 33.47 & 45.75 & 33.05 & 18.63 & 23.24 & 2.48 & 14.78\\
VS Arctan. 10 STS $\gamma=0$ & 5.92 & 17.29 & 23.64 & 15.62 & 35.57 & 31.60 & 14.85 & 27.34\\
VS Arctan. 10 STS $\gamma=10^{-5}$ & 10.01 & 32.46 & 27.71 & 23.39 & 28.49 & 16.04 & 9.02 & 17.85\\
VS Arctan. 20 STS $\gamma=0$ & 6.56 & 13.02 & 21.81 & 13.80 & 35.19 & 26.86 & 11.24 & 24.43\\
VS Arctan. 20 STS $\gamma=10^{-5}$ & 11.28 & 21.59 & 24.74 & 19.21 & 29.35 & 18.29 & 8.95 & 18.86\\
VS Arctan. 30 STS $\gamma=0$ & 4.32 & 13.50 & 22.52 & 13.45 & 35.53 & 27.46 & 12.81 & 25.27\\
VS Arctan. 30 STS $\gamma=10^{-5}$ & 7.53 & 15.93 & 23.29 & 15.58 & 29.41 & 20.90 & 8.47 & 19.59\\
\hline

LIF Arctan. 1 STS DE $\gamma=0$ & 27.20 & 36.93 & 33.15 & 32.42 & 3.90 & 4.82 & 10.57 & 6.43\\
LIF Arctan. 1 STS DE $\gamma=10^{-5}$ & 26.24 & 35.14 & 29.40 & 30.26 & 3.89 & 8.00 & 10.29 & 7.39\\
LIF Arctan. 10 STS DE $\gamma=0$ & 9.00 & 13.38 & 10.89 & 11.09 & 8.77 & 8.55 & 11.26 & 9.53\\
LIF Arctan. 10 STS DE $\gamma=10^{-5}$ & 9.15 & 12.46 & 12.28 & 11.30 & 8.81 & 10.90 & 11.46 & 10.39\\
LIF Arctan. 20 STS DE $\gamma=0$ & 7.96 & 11.39 & 10.89 & 10.08 & 9.89 & 8.85 & 11.33 & 10.02\\
LIF Arctan. 20 STS DE $\gamma=10^{-5}$ & 7.42 & 11.69 & 10.08 & 9.73 & 9.59 & 9.16 & 10.76 & 9.84\\
LIF Arctan. 30 STS DE $\gamma=0$ & 8.24 & 9.80 & 8.25 & 8.76 & 9.84 & 9.41 & 11.71 & 10.32\\
LIF Arctan. 30 STS DE $\gamma=10^{-5}$ & 8.05 & 10.46 & 9.02 & 9.18 & 9.89 & 9.34 & 11.23 & 10.15\\
\hline
LIF Arctan. 10 STS RE $\gamma=0$ & 30.83 & 56.32 & 63.97 & 50.37 & 5.12 & 5.23 & 4.93 & 5.09\\
LIF Arctan. 10 STS RE $\gamma=10^{-5}$ & 29.99 & 55.86 & 63.70 & 49.85 & 5.03 & 4.43 & 4.74 & 4.73\\
LIF Arctan. 20 STS RE $\gamma=0$ & 26.20 & 54.18 & 62.08 & 47.49 & 7.79 & 6.12 & 4.27 & 6.06\\
LIF Arctan. 20 STS RE $\gamma=10^{-5}$ & 23.10 & 50.26 & 60.79 & 44.71 & 7.01 & 5.98 & 4.53 & 5.84\\
LIF Arctan. 30 STS RE $\gamma=0$ & 21.10 & 47.90 & 57.10 & 42.04 & 7.89 & 6.02 & 4.51 & 6.14\\
LIF Arctan. 30 STS RE $\gamma=10^{-5}$ & 20.76 & 48.10 & 57.82 & 42.23 & 7.68 & 5.41 & 4.51 & 5.87\\
\hline
\end{tabular}}
\label{tab:sparsity_results_ldc}
\end{table*}

\begin{table*}[t]
\centering
\caption{LEE, LErr, EE, and LEn scores for the Hoyer-Based SAR-NOMAD with $\gamma=0.005$ and various VS/LIF-NOMAD results for the 2D Heat Exchanger.}
\makebox[\textwidth][c]{\begin{tabular}
{c|cccc}
\hline
 & \multicolumn{4}{c}{Performance Scores}  \\
Spiking Model
& LEE
& LErr
& EE
& LEn\\
\hline
SAR Hoyer $\gamma=0.005$ & \textbf{26.35} & \textbf{5.41} & \textbf{26.35} & \textbf{4.87} \\
\hline
VS Arctan. 1 STS $\gamma=0$ & 139.48 & 10.02 & 139.48 & 13.92 \\
VS Arctan. 1 STS $\gamma=10^{-6}$ & 259.51 & 34.74 & 259.51 & 7.47\\
VS Arctan. 10 STS $\gamma=0$ & 1202.19 & 79.30 & 120.22 & 151.60\\
VS Arctan. 10 STS $\gamma=10^{-6}$ & 2220.35 & 175.80 & 222.04 & 126.3\\
VS Arctan. 20 STS $\gamma=0$ & 2148.09 & 159.00 & 107.40 & 270.20\\
VS Arctan. 20 STS $\gamma=10^{-6}$ & 2668.8 & 244.40 & 133.44 & 218.40\\
\hline
LIF Arctan. 1 STS DE $\gamma=0$ & 287.73 & 13.54 & 287.73 & 21.25 \\
LIF Arctan. 1 STS DE $\gamma=10^{-6}$ & 186.39 & 16.18 & 186.39 & 11.52\\
LIF Arctan. 10 STS DE $\gamma=0$ & 1249.90 & 86.20 & 124.99 & 145.00\\
LIF Arctan. 10 STS DE $\gamma=10^{-6}$ & 1189.17 & 84.10 & 118.92 & 141.40\\
LIF Arctan. 20 STS DE $\gamma=0$ & 2352.64 & 183.80 & 117.63 & 256.00\\
LIF Arctan. 20 STS DE $\gamma=10^{-6}$ & 2345.28 & 168.00 & 117.26 & 279.20\\
\hline
LIF Arctan. 10 STS RE $\gamma=0$ & 4876.39 & 723.50 & 487.64 & 67.40\\
LIF Arctan. 10 STS RE $\gamma=10^{-6}$ & 4704.30 & 733.90 & 470.43 & 64.10\\
LIF Arctan. 20 STS RE $\gamma=0$ & 10023.07 & 1349.00 & 501.15 & 148.60\\
LIF Arctan. 20 STS RE $\gamma=10^{-6}$ & 8739.66 & 1359.20 & 436.98 & 128.60\\
\hline
VS Fast Sig. 1 STS $\gamma=0$ & 308.57 & 36.91 & 308.57 & 8.36\\
VS Fast Sig. 1 STS $\gamma=10^{-6}$ & 307.34 & 45.60 & 307.34 & 6.74 \\
VS Fast Sig. 10 STS $\gamma=0$ & 2458.53 & 223.30 & 245.85 & 110.10 \\
VS Fast Sig. 10 STS $\gamma=10^{-6}$ & 2569.66 & 245.90 & 256.97  & 104.50 \\
VS Fast Sig. 20 STS $\gamma=0$ & 4854.21 & 415.60 & 242.71 & 233.60 \\
VS Fast Sig. 20 STS $\gamma=10^{-6}$ & 5741.57 & 537.60 & 287.08 & 213.6 \\
\hline
\end{tabular}}
\label{tab:performance_scores_hx}
\end{table*}

\begin{table*}[t]
\centering
\caption{LEE, LErr, EE, and LEn scores for the Hoyer-Based SAR-NOMAD with $\gamma=0.005$ and various VS-NOMAD results for the Lid Driven Cavity.}
\makebox[\textwidth][c]{\begin{tabular}
{c|cccc}
\hline
 & \multicolumn{4}{c}{Performance Scores}  \\
Spiking Model
& LEE
& LErr
& EE
& LEn\\
\hline
SAR Hoyer $\gamma=0.05$ & \textbf{37.96} & \textbf{9.81} & \textbf{37.96} & \textbf{3.87} \\
\hline
VS Arctan. 1 STS $\gamma=0$ & 285.31 & 14.28 & 285.31 & 19.98 \\
VS Arctan. 1 STS $\gamma=10^{-5}$ & 488.48 & 33.05 & 488.48 & 14.78\\
VS Arctan. 10 STS $\gamma=0$ & 4270.51 & 156.20 & 427.05 & 273.40\\
VS Arctan. 10 STS $\gamma=10^{-5}$ & 4175.12 & 233.90 & 417.51 & 178.50\\
VS Arctan. 20 STS $\gamma=0$ & 6742.68 & 276.00 & 337.13 & 488.60\\
VS Arctan. 20 STS $\gamma=10^{-5}$ & 7246.01 & 384.20 & 362.30 & 377.20\\
VS Arctan. 30 STS $\gamma=0$ & 10196.45 & 403.50 & 339.88 & 758.10\\
VS Arctan. 30 STS $\gamma=10^{-5}$ & 9156.37 & 467.40 & 305.21 & 587.70\\
\hline
LIF Arctan. 1 STS DE $\gamma=0$ & 208.46 & 32.42 & 208.46 & 6.43 \\
LIF Arctan. 1 STS DE $\gamma=10^{-6}$ & 223.62 & 30.26 & 223.62 & 7.39\\
LIF Arctan. 10 STS DE $\gamma=0$ & 1056.88 & 110.90 & 105.69 & 95.30\\
LIF Arctan. 10 STS DE $\gamma=10^{-6}$ & 1174.07 & 113.00 & 117.41 & 103.90\\
LIF Arctan. 20 STS DE $\gamma=0$ & 2020.03 & 201.60 & 101.00 & 200.40\\
LIF Arctan. 20 STS DE $\gamma=10^{-6}$ & 1914.86 & 194.60 & 95.74 & 196.80\\
LIF Arctan. 30 STS DE $\gamma=0$ & 2712.10 & 262.80 & 90.40 & 309.60\\
LIF Arctan. 30 STS DE $\gamma=10^{-6}$ & 2795.31 & 275.40 & 93.18 & 304.50\\
\hline
LIF Arctan. 10 STS RE $\gamma=0$ & 2563.83 & 503.70 & 256.38 & 50.90\\
LIF Arctan. 10 STS RE $\gamma=10^{-6}$ & 2357.91 & 498.5 & 235.79 & 47.30\\
LIF Arctan. 20 STS RE $\gamma=0$ & 5755.79 & 949.80 & 287.79 & 121.20\\
LIF Arctan. 20 STS RE $\gamma=10^{-6}$ & 5222.13 & 894.20 & 261.11 & 116.80\\
LIF Arctan. 30 STS RE $\gamma=0$ & 7743.77 & 1261.20 & 258.13 & 184.20\\
LIF Arctan. 30 STS RE $\gamma=10^{-6}$ & 7436.70 & 1266.90 & 247.89 & 176.10\\
\hline
VS Fast Sig. 1 STS $\gamma=0$ & 383.31 & 14.41 & 383.31 & 26.60\\
VS Fast Sig. 1 STS $\gamma=10^{-5}$ & 487.69 & 39.33 & 487.69 & 12.40 \\
VS Fast Sig. 10 STS $\gamma=0$ & 3158.60 & 195.70 & 315.86 & 161.40 \\
VS Fast Sig. 10 STS $\gamma=10^{-5}$ & 4035.39 & 213.40 & 403.54 & 189.10 \\
VS Fast Sig. 20 STS $\gamma=0$ & 5151.72 & 257.20 & 257.59 & 400.60 \\
VS Fast Sig. 20 STS $\gamma=10^{-5}$ & 8306.82 & 408.80 & 415.34 & 406.40 \\
VS Fast Sig. 30 STS $\gamma=0$ & 12902.10 & 444.9 & 430.07 & 870.00 \\
VS Fast Sig. 30 STS $\gamma=10^{-5}$ & 15998.64 & 934.50 & 533.29 & 513.60 \\
\hline
\end{tabular}}
\label{tab:performance_scores_ldc}
\end{table*}

\subsection{Synthetic Distillation Performance}
\label{sec:distillation_results}

As discussed in Section \ref{sec:distillation}, within virtual sensing, and particularly spatial-temporal field reconstruction, the availability of training data is frequently limited for several reasons. For experimentally acquired datasets, measurements may be restricted due to security considerations or may be prohibitively expensive to obtain. Furthermore, when the desired reconstruction cannot be directly measured, high-fidelity computational methods, such as the Finite Element Method (FEM), are required to generate training data, resulting in substantial computational cost. Consequently, neural operators for virtual sensing are often trained using relatively small datasets.

Advanced neural operator architectures, such as VIRSO and Geo-FNO, can effectively approximate nonlinear operator mappings under limited data conditions due to their strong inductive biases, sophisticated spectral representations, and ability to operate on irregular geometries. However, these architectures require convolution over the entire computational grid, making them less suitable for deployment on edge devices and neuromorphic hardware. In contrast, NOMAD and other trunk-branch neural operator architectures consist primarily of fully connected network (FCN) layers and independently evaluate each spatial location. While this structure is well suited for neuromorphic implementation, it lacks the enriched feature extraction and communication dynamics present in convolution-based architectures. 

This behavior is reflected in the performance difference between VS-GNO and VS-NOMAD on the 2D Heat Exchanger dataset, despite both models employing the same Variable Spiking Neuron (VSN). At $\gamma = 0.5$, VS-GNO achieves both a lower mean relative $L_2$ error and lower average spiking activity than the majority of VS-NOMAD configurations. These observations bring us towards synthetic distillation, as discussed in Section \ref{sec:distillation}. The large graph-based VIRSO model provides new data examples, based on the distributions in Equation \ref{eq:distributions} and the model's outputs, for SAR-NOMAD which results in better optimization and convergence while ideally preserving the energy-efficiency of the model architecture.

To establish the feasibility of this framework, the synthetic input-output pairs from VIRSO are then combined with the original training dataset and used to train a SAR-NOMAD model with the Hoyer sparsity loss and $\gamma = 0.001$, $0.005$, and $0.01$. Tables \ref{tab:synthetic_results1}, \ref{tab:synthetic_results2}, and \ref{tab:synthetic_results3} summarize the resulting performance using the same test dataset employed throughout the previous experiments.

For all gamma examples, as we add more synthetic examples, the mean relative $L_2$ error decreases with either similar or improved spiking efficiency. With the addition of 8,000 synthetic training examples, the mean relative $L_2$ error is reduced by approximately a factor of two for both $\gamma = 0.001$ and $0.005$, while the average spiking activity decreases by approximately 1\% for $\gamma = 0.001$ and $0.6\%$ for $\gamma = 0.005$. These results indicate that additional knowledge generated by the teacher model enables SAR-NOMAD to maintain its energy-efficient sparse representations while learning significantly more accurate field reconstructions. Consequently, the overall Latency-Error-Energy (LEE) score for the $\gamma=0.005$ result is reduced by more than a factor of two to 11.34.

For $\gamma = 0.01$ in Table~\ref{tab:synthetic_results3}, the baseline model trained without synthetic data exhibits substantial performance degradation, with the relative $L_2$ error increasing to nearly $19\%$ while reducing the average spiking activity to below $2\%$. Introducing 1,000 synthetic training samples dramatically improves reconstruction accuracy, reducing the $L_2$ error to below $5\%$ while increasing the average spiking activity by only $1.3\%$. As additional synthetic samples are incorporated, the model continues to follow the same accuracy trend, with 8,000 synthetic samples further reducing the mean $L_2$ error by approximately a factor of two relative to the 1,000-sample case while requiring only a slight increase in spiking activity. Although the 8,000-sample model exhibits higher average spiking than the original $\gamma = 0.01$ baseline, the substantial improvement in reconstruction accuracy demonstrates the effectiveness of our initial synthetic distillation approach for SAR-NOMAD. In contrast, applying synthetic distillation to VS-NOMAD using the 10 STS, $\gamma = 10^{-6}$, arctangent configuration yielded minimal improvements in either $L_2$ error or average spiking percentage. These results suggest that extending synthetic distillation to variable-spiking architectures is considerably more challenging, potentially due to a mismatch between the surrogate gradient used during optimization and the behavior of the synthesized training data.

However, generating 8,000 synthetic samples increases the effective training dataset to nearly nine times its original size. Although this straightforward synthetic data generation strategy substantially reduces the dependence on expensive experimental measurements or computationally intensive high-fidelity simulations, it still introduces a considerable computational burden during training because a large quantity of synthetic data is required when random input sampling is employed. Future work should therefore investigate more sophisticated knowledge distillation strategies that transfer information more efficiently while requiring fewer synthetic examples. Nevertheless, the present results demonstrate the viability of a distillation framework specifically tailored for spiking neural operators, wherein a teacher model that is not inherently suitable for neuromorphic deployment learns from limited ground-truth data and subsequently generates synthetic training data that enables a neuromorphic-friendly student model to achieve substantially better performance than training on the original dataset alone.

\begin{table*}[t]
\centering
\caption{$L_2$ errors and spiking percentages for SAR-NOMAD and Hoyer loss term at $\gamma = 0.001$ for different synthetically generated training example counts with the 2D Heat Exchanger.}
\makebox[\textwidth][c]{\begin{tabular}{c|cccccc|ccccc}
\hline
 & \multicolumn{6}{c|}{Mean Relative L2 Errors (\%)} & \multicolumn{5}{c}{Mean Spiking Percentages (\%)} \\
Spiking Model
& $p$
& $v_z$
& $v_y$
& $v_x$
& $v_{\mathrm{mag}}$
& Mean
& Trunk
& B1
& B2
& Comb
& Mean \\
\hline
SAR Hoyer (Original $\gamma=0.001$) & 2.14 & 4.42 & 3.55 & 2.39 & 2.96 & 3.09 & 17.81 & 1.86 & 2.07 & 14.09 & 8.96 \\
SAR Hoyer (Samples = 1000) & 1.83 & 4.08 & 2.96 & 2.48 & 2.84 & 2.84 & 17.55 & 1.95 & 2.22 & 13.45 & 8.79 \\
SAR Hoyer (Samples = 2000) & 1.58 & 2.78 & 2.41 & 2.12 & 2.65 & 2.31 & 17.28 & 1.56 & 1.71 & 13.36 & 8.48 \\
SAR Hoyer (Samples = 4000) & 1.43 & 2.45 & 1.90 & 1.46 & 2.14 & 1.87 & 16.92 & 0.97 & 1.91 & 13.13 & 8.23 \\
SAR Hoyer (Samples = 8000) & 1.17 & 1.70 & 1.56 & 1.24 & 1.89 & 1.51 & 15.91 & 1.07 & 1.23 & 12.68 & 7.72 \\
\hline
\end{tabular}}
\label{tab:synthetic_results1}
\end{table*}

\begin{table*}[t]
\centering
\caption{$L_2$ errors and spiking percentages for SAR-NOMAD and Hoyer loss term at $\gamma = 0.005$ for different synthetically generated training example counts with the 2D Heat Exchanger.}
\makebox[\textwidth][c]{\begin{tabular}{c|cccccc|ccccc}
\hline
 & \multicolumn{6}{c|}{Mean Relative L2 Errors (\%)} & \multicolumn{5}{c}{Mean Spiking Percentages (\%)} \\
Spiking Model
& $p$
& $v_z$
& $v_y$
& $v_x$
& $v_{\mathrm{mag}}$
& Mean
& Trunk
& B1
& B2
& Comb
& Mean \\
\hline
SAR Hoyer (Original $\gamma=0.005$) & 4.11 & 6.57 & 5.82 & 4.84 & 5.72 & 5.41 & 8.91 & 0.60 & 0.83 & 9.15 & 4.87 \\
SAR Hoyer (Samples = 1000) & 3.11 & 5.63 & 4.84 & 3.51 & 4.42 & 4.30 & 8.55 & 0.39 & 0.75 & 8.98 & 4.67 \\
SAR Hoyer (Samples = 2000) & 2.92 & 4.34 & 3.76 & 2.90 & 3.56 & 3.50 & 8.77 & 0.49 & 0.71 & 8.67 & 4.66 \\
SAR Hoyer (Samples = 4000) & 2.61 & 3.75 & 3.39 & 2.70 & 3.40 & 3.17 & 8.23 & 0.39 & 0.68 & 8.88 & 4.55 \\
SAR Hoyer (Samples = 8000) & 2.20 & 3.06 & 2.92 & 2.19 & 3.02 & 2.68 & 7.97 & 0.39 & 0.58 & 7.99 & 4.23 \\
\hline
\end{tabular}}
\label{tab:synthetic_results2}
\end{table*}

\begin{table*}[t]
\centering
\caption{$L_2$ errors and spiking percentages for SAR-NOMAD and Hoyer loss term at $\gamma = 0.01$ for different synthetically generated training example counts with the 2D Heat Exchanger.}
\makebox[\textwidth][c]{\begin{tabular}{c|cccccc|ccccc}
\hline
 & \multicolumn{6}{c|}{Mean Relative L2 Errors (\%)} & \multicolumn{5}{c}{Mean Spiking Percentages (\%)} \\
Spiking Model
& $p$
& $v_z$
& $v_y$
& $v_x$
& $v_{\mathrm{mag}}$
& Mean
& Trunk
& B1
& B2
& Comb
& Mean \\
\hline
SAR Hoyer (Original $\gamma=0.01$) & 8.24 & 11.22 & 11.80 & 29.21 & 36.17 & 19.33 & 3.31 & 0.49 & 0.62 & 2.96 & 1.85 \\
SAR Hoyer (Samples = 1000) & 4.05 & 6.31 & 5.05 & 4.08 & 5.26 & 4.95 & 6.15 & 0.39 & 0.48 & 5.57 & 3.15 \\
SAR Hoyer (Samples = 2000) & 3.35 & 5.31 & 3.98 & 3.20 & 4.26 & 4.02 & 6.25 & 0.39 & 0.48 & 6.28 & 3.35 \\
SAR Hoyer (Samples = 4000) & 2.89 & 4.25 & 3.32 & 2.80 & 3.85 & 3.42 & 6.33 & 0.39 & 0.57 & 5.76 & 3.26 \\
SAR Hoyer (Samples = 8000) & 2.52 & 3.43 & 2.96 & 2.34 & 3.26 & 2.90 & 6.46 & 0.39 & 0.48 & 5.76 & 3.27 \\
\hline
\end{tabular}}
\label{tab:synthetic_results3}
\end{table*}

\subsection{Sparse Activation ReLU towards VS-GNO}
\label{sec:activation_variable_spiking}

\paragraph{Improving The VSN Loss Term}

In order to preserve the dynamic spiking behavior of the VSN, enabling richer feature representations and improved modeling of transient physics, we investigate an alternative regularization strategy rather than applying SAR-based neural operators directly. Specifically, we introduce an activation-based regularization loss inspired by SAR, described in Section~\ref{sec:hoyer_for_variable} and Equation~\ref{eq:vsn_activation}. Unlike the original percentage-based spiking loss, this formulation employs the Hoyer loss to regularize the activation strength of the spiking neurons, providing a more informative measure of spiking behavior. To evaluate the effectiveness of this approach, we compare the original VS-NOMAD with Activation Regularized VS-NOMAD (ARVS-NOMAD) on both the 2D Heat Exchanger and Lid-Driven Cavity datasets using 1 and 10 STS configurations and an arctangent surrogate gradient across a range of $\gamma$ values, as summarized in Tables ~\ref{tab:activ_vsn_results} and \ref{tab:activ_vsn_results_ldc}.

For the 1 STS experiments on the 2D Heat Exchanger, ARVS-NOMAD consistently achieves substantially lower mean relative $L_2$ error than the original VS-NOMAD. The highest-performing ARVS-NOMAD configuration reduces the mean $L_2$ error to below $9\%$, whereas all VS-NOMAD configurations remain above $30\%$. Furthermore, the spiking activity of ARVS-NOMAD is comparable to or lower than that of VS-NOMAD, with the $\gamma = 5 \cdot 10^{-7}$ configuration achieving both lower spiking and lower reconstruction error than every VS-NOMAD result. Similar improvements are observed for the 10 STS experiments, where ARVS-NOMAD consistently produces lower mean $L_2$ errors than the original VS-NOMAD. Although the average spiking activity is generally slightly higher, a direct comparison of identical $\gamma$ values is not entirely appropriate because the percentage-based and activation-based regularization terms differ in scale and magnitude. Nevertheless, ARVS-NOMAD is able to reduce the average spiking activity to below $6.52\%$ while achieving an $L_2$ error significantly lower than the $16.05\%$ obtained by the lowest-spiking VS-NOMAD configuration. Interestingly, the reconstruction error temporarily improves as $\gamma$ increases from $5 \cdot 10^{-8}$ to $10^{-6}$, suggesting that a moderate amount of activation regularization improves optimization by encouraging more efficient utilization of the learned feature representations.

The Lid-Driven Cavity results exhibit similar trends. Across both the 1 and 10 STS configurations, ARVS-NOMAD consistently achieves mean relative $L_2$ errors of approximately $11\%$, whereas the corresponding VS-NOMAD models remain above $23\%$. For the 1 STS configuration, ARVS-NOMAD also reduces the average spiking activity, providing improvements in both reconstruction accuracy and energy efficiency. For the 10 STS configuration with $\gamma = 10^{-4}$, ARVS-NOMAD achieves a substantial reduction in $L_2$ error compared to VS-NOMAD while maintaining nearly identical spiking activity.

The results on both the Heat Exchanger and Lid-Driven Cavity datasets demonstrate the improved learning dynamics enabled by the proposed activation-based regularization. By leveraging the superior sparsity-promoting properties of the Hoyer loss, as demonstrated in our comparison between the $L_1$ and Hoyer regularizers, ARVS-NOMAD regularizes the magnitude of neuron activations rather than treating spikes as purely binary events, as in the original percentage-based loss. Specifically, the proposed regularization operates on the difference between the pre-reset membrane potential and the firing threshold, thereby encoding richer information about spike intensity. This formulation preferentially penalizes neurons with consistently large membrane potentials/low thresholds that are likely to spike continuously, encouraging more adaptive threshold behavior and improved feature utilization. Consequently, this targeted regularization produces both lower reconstruction errors and reduced spiking activity.

Despite these improvements, ARVS-NOMAD still underperforms the SAR-NOMAD architectures because the underlying optimization remains dependent on surrogate gradients, resulting in an inherent gradient mismatch. Consequently, these results primarily demonstrate the effectiveness of combining the SAR-inspired activation encoding with Hoyer regularization for developing energy-efficient neural operators, rather than resolving the optimization challenges associated with variable-spiking neural operators. Future work should therefore investigate alternative training algorithms that eliminate or mitigate the reliance on surrogate gradients.

\begin{table*}[t]
\centering
\caption{$L_2$ errors and spiking percentages for Activation-Regularized VS-NOMAD and Percentage-Regularized VS-NOMAD with the 2D Heat Exchanger trained using the Arctangent surrogate gradient function.}
\makebox[\textwidth][c]{\begin{tabular}{c|cccccc|ccccc}
\hline
 & \multicolumn{6}{c|}{Mean Relative L2 Errors (\%)} & \multicolumn{5}{c}{Mean Spiking Percentages (\%)} \\
Spiking Model
& $p$
& $v_z$
& $v_y$
& $v_x$
& $v_{\mathrm{mag}}$
& Mean
& Trunk
& B1
& B2
& Comb
& Mean \\
\hline
Spiking Fraction 1 STS $\gamma=10^{-8}$ & 20.03 & 26.50 & 22.18 & 29.71 & 34.23 & 26.53 & 16.31 & 11.13 & 15.19 & 1.85 & 11.12 \\
Spiking Fraction 1 STS $\gamma=5\cdot10^{-8}$ & 34.23 & 41.71 & 37.61 & 35.45 & 44.12 & 38.62 & 16.60 & 10.45 & 14.31 & 3.28 & 11.16 \\
Spiking Fraction 1 STS $\gamma=10^{-7}$ & 13.63 & 22.86 & 22.97 & 34.31 & 41.66 & 27.08 & 14.27 & 8.33 & 13.80 & 2.58 & 7.25\\
Spiking Fraction 1 STS $\gamma=5\cdot10^{-7}$ & 36.84 & 32.49 & 30.87 & 35.39 & 44.92 & 36.10 & 14.19 & 7.03 & 13.94 & 2.14 & 9.33 \\
\hline
Activation 1 STS $\gamma=10^{-8}$ & 9.23 & 18.70 & 17.36 & 15.80 & 16.64 & 15.55 & 15.24 & 12.01 & 15.01 & 4.94 & 11.8 \\
Activation 1 STS $\gamma=5\cdot10^{-8}$ & 4.72 & 11.54 & 9.52 & 8.32 & 9.36 & 8.69 & 17.59 & 0.39 & 16.19 & 8.35 & 10.63 \\
Activation 1 STS $\gamma=10^{-7}$ & 12.18 & 12.27 & 11.16 & 10.67 & 14.50 & 12.16 & 15.74 & 0.20 & 1.00 & 7.84 & 6.20 \\
Activation 1 STS $\gamma=5\cdot10^{-7}$ & 12.63 & 12.13 & 12.49 & 10.28 & 14.25 & 12.36 & 14.79 & 0.39 & 0. & 8.31 & 5.87 \\
\hline
Spiking Fraction 10 STS $\gamma=10^{-8}$ & 5.85 & 12.35 & 11.10 & 9.90 & 11.24 & 10.09 & 18.38 & 12.84 & 14.10 & 4.65 & 12.49 \\
Spiking Fraction 10 STS $\gamma=5\cdot10^{-8}$ & 6.80 & 13.28 & 13.78 & 11.72 & 13.22 & 11.76 & 18.45 & 9.46 & 15.06 & 4.44 & 11.85 \\
Spiking Fraction 10 STS $\gamma=10^{-7}$ & 7.87 & 14.84 & 15.30 & 13.54 & 15.14 & 13.34 & 18.74 & 13.67 & 15.50 & 3.69 & 12.90 \\
Spiking Fraction 10 STS $\gamma=5\cdot10^{-7}$ & 7.75 & 14.34 & 13.93 & 14.08 & 15.67 & 13.15 & 20.69 & 12.47 & 15.10 & 3.68 & 12.99 \\
Spiking Fraction 10 STS $\gamma=10^{-6}$ & 12.72 & 18.13 & 19.05 & 18.62 & 20.48 & 17.58 & 19.35 & 11.43 & 16.58 & 3.17 & 12.63 \\
Spiking Fraction 10 STS $\gamma=5\cdot10^{-6}$ & 9.61 & 14.42 & 15.33 & 17.73 & 23.18 & 16.05 & 16.90 & 5.47 & 13.33 & 2.95 & 9.66 \\
\hline
Activation 10 STS $\gamma=10^{-8}$ & 4.25 & 9.97 & 9.17 & 8.21 & 8.63 & 8.05 & 19.83 & 15.36 & 16.31 & 9.58 & 15.27 \\
Activation 10 STS $\gamma=5\cdot10^{-8}$ & 4.57 & 11.68 & 11.71 & 10.22 & 9.88 & 9.61 & 22.94 & 15.33 & 16.60 & 9.51 & 16.10 \\
Activation 10 STS $\gamma=10^{-7}$ & 4.60 & 11.01 & 9.88 & 9.17 & 9.68 & 8.87 & 22.48 & 15.14 & 16.65 & 10.29 & 16.14 \\
Activation 10 STS $\gamma=5\cdot10^{-7}$ & 3.42 & 9.21 & 8.53 & 6.84 & 7.05 & 7.01 & 22.48 & 10.64 & 16.96 & 9.91 & 15.00 \\
Activation 10 STS $\gamma=10^{-6}$ & 3.66 & 9.32 & 7.75 & 5.73 & 6.41 & 6.57 & 20.90 & 0. & 16.59 & 8.27 & 11.44 \\
Activation 10 STS $\gamma=5\cdot10^{-6}$ & 11.58 & 12.09 & 11.72 & 10.86 & 14.63 & 12.18 & 18.72 & 0.29 & 0.10 & 6.96 & 6.52 \\

\hline
\end{tabular}}
\label{tab:activ_vsn_results}
\end{table*}

\begin{table*}[t]
\centering
\caption{$L_2$ errors and spiking percentages for Activation-Regularized VS-NOMAD and Percentage-Regularized VS-NOMAD with the Lid Driven Cavity trained using the Arctangent surrogate gradient function.}
\makebox[\textwidth][c]{\begin{tabular}{c|cccc|cccc}
\hline
 & \multicolumn{4}{c|}{Mean Relative L2 Errors (\%)} & \multicolumn{4}{c}{Mean Spiking Percentages (\%)} \\
Spiking Model
& $p$
& $v$
& $k$
& Mean
& Trunk
& Branch
& Comb
& Mean \\
\hline
Spiking Fraction 1 STS $\gamma=10^{-5}$ & 19.92 & 33.47 & 45.75 & 33.05 & 18.63 & 23.24 & 2.48 & 14.78\\
Spiking Fraction 10 STS $\gamma=10^{-5}$  & 10.01 & 32.46 & 27.71 & 23.39 & 28.49 & 16.04 & 9.02 & 17.85\\
\hline
Activation 1 STS $\gamma=10^{-5}$ & 4.79 & 9.06 & 19.67 & 11.17 & 21.08 & 8.61 & 7.09 & 12.26\\
Activation 10 STS $\gamma=10^{-5}$ & 4.92 & 9.13 & 19.17 & 11.07 & 31.07 & 19.23 & 10.27 & 20.19\\
Activation 10 STS $\gamma=10^{-4}$ & 5.46 & 9.84 & 19.14 & 11.48 & 28.11 & 13.27 & 9.86 & 17.08\\
\hline
\end{tabular}}
\label{tab:activ_vsn_results_ldc}
\end{table*}

\paragraph{Reducing Spatial Aggregation}

\begin{table*}[t]
\centering
\caption{Average error and spiking with the layer-wise neighbor count and average count for the VS-GNO model on the Heat Exchanger with the arctangent surrogate function utilizing both spatial thresholding gating mechanisms: shared threshold (Exp. 1) and edge-dependent threshold (Exp.2).}
\makebox[\textwidth][c]{\begin{tabular}{c|c|c|c|c}
\hline
Spiking Model
& Mean $L_2$
& Mean Spiking
& Neighbor Count Per Layer
& Avg. Count\\
\hline
 Exp. 1 ($\epsilon = 10^{-2}, \gamma=0$) & \textbf{1.23} & \textbf{19.52} & \textbf{[1.06, 0., 18.74, 0., 19.20, 6.64, 0., 0., 0., 0.]} & \textbf{4.56}\\
 Exp. 1 ($\epsilon = 10^{-3}, \gamma=0$) & 3.01 & 25.39 & [31.73, 24.15, 0., 0., 30.93, 32.25, 0., 31.53, 0., 0.] & 15.05\\
 Exp. 2 ($\epsilon = 10^{-2}, \gamma=0$) & 0.69 & 25.86 & [0., 0., 0., 0., 0., 0., 0., 0., 0., 0.] & 0.\\
 Exp. 2 ($\epsilon = 10^{-3}, \gamma=0$) & 0.66 & 27.79 & [0., 0., 0., 22.03, 0., 0., 0., 0., 0., 0.] & 2.20\\
\hline
 Exp. 1 ($\epsilon = 10^{-2}, \gamma=0.5$) & 10.23 & 19.27 & [33.84, 0., 0., 0., 19.17, 23.06, 0., 0., 0., 0.] & 7.61\\
 Exp. 1 ($\epsilon = 10^{-3}, \gamma=0.5$) & 4.17 & 12.16 & [0., 0., 33.84, 0., 3.12, 19.77, 0., 0., 0., 3.34] & 6.01\\
 Exp. 2 ($\epsilon = 10^{-2}, \gamma=0.5$) & \textbf{0.93} & \textbf{7.74} & \textbf{[0, 0., 0., 0., 1.63, 0., 0., 0., 0., 0.]} & \textbf{0.16}\\
 Exp. 2 ($\epsilon = 10^{-3}, \gamma=0.5$) & \textbf{1.22} & \textbf{8.16} & \textbf{[0., 0., 5.20, 0., 0., 0., 21.42, 0., 30.29, 0.]} & \textbf{5.69}\\
 \hline
 k = 5, $\gamma=0$ & 9.84 & 23.42 & - & 5.87\\
   k = 10, $\gamma=0$ & 8.16 & 28.70 & - & 11.75\\
   k = 15, $\gamma=0$ & 9.95 & 22.14 & - & 17.51\\
 \hline
   k = 5, $\gamma=0.5$ & 13.51 & 23.79 & - & 5.87\\
  k = 10, $\gamma=0.5$ & 14.97 & 15.03 & - & 11.75\\
  k = 15, $\gamma=0.5$ & 3.46 & 10.36 & - & 17.51\\
\end{tabular}}
\label{tab:spatial_gating}
\end{table*}

To demonstrate the potential of ReLU thresholding to reduce spatial aggregation costs and, consequently, the latency of VS-GNO implementations, Table \ref{tab:spatial_gating} presents the full VS-GNO results using the arctangent surrogate gradient with $\gamma=0$ and $\gamma=0.5$ for the two proposed thresholding techniques: a single trainable threshold value (Exp. 1) and an edge-dependent trainable threshold value (Exp. 2). We utilize two $\epsilon$ values (0.01 and 0.001) to demonstrate different levels of emphasis on retaining neighbor connections. For comparison, we also report the full VS-GNO with a reduced number of nearest neighbors within the KNN graph construction utilized for these results. For $\gamma=0$, all thresholding experiments outperform the reduced-neighbor configurations in terms of $L_2$ error while maintaining slightly higher or similar spiking percentages and similar or reduced neighbor counts. For $\epsilon=0.01$ and $\gamma=0$, with the constant trainable threshold (Exp. 1), we achieve an almost sevenfold lower $L_2$ error than the best-performing reduced-neighbor configuration based on KNN construction, while also obtaining lower spiking and an average neighbor count of less than 5. Examining the layer-wise neighbor statistics shows that the thresholding technique effectively removes neighbors in specific layers while maintaining aggregation in others, essentially finding a favorable tradeoff between reconstruction error and the overall need for spatial aggregation. With the edge-dependent thresholding at $\gamma=0$, we observe cases where the neighbor count is reduced to zero, effectively reducing the model to a spectral-only version. Despite this, the resulting models maintain errors below 1\% with similar spiking percentages. For $\gamma=0.5$, the edge-dependent thresholding (Exp. 2) demonstrates the better overall performance, providing greater control over which neighbors are blocked based on their spatial positioning and edge weights. In particular, the $\epsilon=0.001$ result for $\gamma=0.5$ in Exp. 2 achieves an average $L_2$ error of approximately 1\%, a spiking percentage below 9\%, and an average neighbor count lower than all reduced-neighbor alternatives for $\gamma=0.5$. These results suggest that trainable edge thresholding is preferable over simply reducing neighbor count which not only affects the amount of spatial aggregation but also limits the model's ability to selectively determine which neighbors should be removed, while simultaneously affecting the spectral component. Experimenting with post-training thresholding, which essentially entailed finetuning of a constant threshold, resulted in poor performance which makes the proposed trainable ReLU thresholding approach a more favorable alternative for maintaining low $L_2$ error while reducing spatial aggregation. We also experimented the thresholding concept with the full SAR-GNO model but the results demonstrated more error degradation. Since the SAR implementations were shown, in Table \ref{tab:sparsity_results_graph_full}, to have far less spiking (at least 10 times less compared to the arctangent VS-GNO results) for the spatial blocks, the inclusion of the neighbor ReLU-based loss term could have thrown off ideal convergence for the SAR implementation especially since it has been shown with SAR-NOMAD that lower spiking leads to collapse in the feature dimension. This, of course, is purely speculative and further analysis into the spiking behavior of SAR-GNO and the impact of the neighbor thresholding is required. Nevertheless, if spatial aggregation is required for local spatial calibration, the proposed method provides a potential mechanism that does require further exploration to determine its true efficacy. At least in the context of the VS-GNO algorithm, which does rely more on the spatial aggregation in terms of spiking amount, our proposed neighbor "filter" has promising performance towards optimizing the tradeoff between accuracy and the overall number of neighbor connections. Further research is also required to fully understand the impact of the proposed thresholding techniques and the differences between the two thresholding approaches. Also, further applications are required to determine whether spatial aggregation is necessary at all, given that the spectral-only layer achieves better $L_2$ error and that lower neighbor counts correspond to Exp. 1 with $\gamma=0$ and Exp. 2 with $\gamma=0.5$.

\section{Conclusions and Further Work}

In this paper, we addressed several key limitations of previously presented energy-efficient solutions, mainly spiking neural operators and the Variable Spiking Neuron. Owing to surrogate gradient mismatch, the reconstruction performance of single-spike-timestep (1 STS) model tends to degrade in specific circumstances, often requiring additional spike timesteps to recover accuracy. Increasing the number of spike timesteps, however, introduces undesirable latency and significantly increases the computational burden during training. We demonstrate this limitation through memory constraints encountered on a single H200 GPU when training 30 STS and 40 STS models for the 2D Heat Exchanger and Lid-Driven Cavity use cases. To overcome these challenges, we move away from the traditional spiking-based models proposed for energy efficiency and propose an activation-sparsity framework. We present a low-latency, single-step framework based on a Sparse-Activation-ReLU (SAR) layer that avoids any surrogate gradient mismatch entirely and allows for variable communication similar to the VSN while removing the temporal integration that can risk increased latency, operating in a one-step forward pass. The SAR layer provides a alternative solution to the typical spiking implementations. Although other hardware implementations that take advantage of activation sparsity, our energy-efficient models can fit within an ANN-to-neuromorphic framework, utilizing a zero-thresholding event policy, that is unlike traditional ANN-SNN algorithms which rely on rate coding and binary communication and risk precision loss. Although SAR is somewhat more restrictive than standard VSNs, as it only permits positive activations beyond the zero threshold whereas VSNs can utilize both positive and negative thresholds and signals, it consistently outperformed or performed similarly to VSN-based and LIF-based implementations for both the trunk-branch NOMAD architecture and the graph-based VIRSO architecture (which utilized GeLU layers) on the presented benchmark problems.

To quantify the overall tradeoff between reconstruction accuracy, latency, and energy efficiency, we introduced the Latency-Energy-Error (LEE) score, which equally weights latency (number of spike time steps or STS), reconstruction error (mean relative $L_2$ error (\%)), and energy consumption (average spiking (\%)). Using this metric, we found that SAR-NOMAD improved upon VS/LIF-NOMAD by at least a factor of five on the 2D Heat Exchanger and the Lid-Driven Cavity benchmark.

We also introduced analysis based on spiking entropy to quantify how much of a model's feature representation is utilized across the test dataset. Evaluating the Heat Exchanger benchmark, we observed that both SAR-NOMAD and VS-NOMAD exhibit decreasing feature entropy as the target spiking percentage is reduced. For sufficiently small sparsity coefficients, such as $\gamma = 0.005$ for SAR-NOMAD, the entropy approaches zero, indicating that nearly all information is concentrated within a single feature dimension. These results suggest that, in order to satisfy increasingly restrictive sparsity constraints, both SAR and VSN models progressively collapse their latent feature representations to achieve lower activation or spiking rates.

Building upon these findings, we proposed a synthetic distillation framework for neuromorphic/edge device virtual sensing. Synthetic input-output samples were generated using the graph-based VIRSO model and combined with the original ground-truth dataset to train SAR-NOMAD on the 2D Heat Exchanger benchmark. This approach reduced the mean relative $L_2$ error by approximately a factor of two using 8,000 synthetic training samples while maintaining nearly identical or lower average spiking percentages. Consequently, the corresponding LEE score was reduced by approximately a factor of two, further improving SAR-NOMAD's overall performance and increasing its advantage over VS-NOMAD to more than an order of magnitude on the 2D Heat Exchanger benchmark.

Finally, we investigated methods to improve the performance of the VSN implementation which provides potentially beneficially temporal integration, especially in the context of transient dynamics, as well as improved signaling (no ReLU reliance). Rather than replacing the neuron itself, we substituted the original spike-percentage regularization term with a SAR-inspired sparse activation loss. Although these results were not as pronounced as those achieved with SAR-NOMAD, the modified VS-NOMAD models consistently achieved lower reconstruction errors while maintaining comparable or lower spiking percentages on both the 2D Heat Exchanger and Lid-Driven Cavity benchmarks, especially in the 1 STS domain. For the Heat Exchanger with 1 STS and $\gamma = 5\cdot10^{-7}$, the ReLU activation loss implementation provided an $L_2$ error at least 2 times lower than the presented VSN results with a fraction-based spiking loss term while also having an overall lower spiking percentage. These findings suggest that sparse activation regularization provides richer information encoding than the original binary spike-percentage objective while preserving the temporal processing capabilities unique to spiking neural networks. We also explored ReLU-based thresholding within the context of gated neighborhood aggregation of the VS-GNO algorithm for the Heat Exchanger, which showed stronger dependence on the spatial blocks, indicated by higher spiking percentages. Trainable thresholds, either shared or edge-dependent, were able to reduce overall neighbor count with mostly improved L2 error and spiking over lower neighbor constructed graphs or manual post-training thresholding. For $\gamma=0$, the single threshold (Exp. 1) result with $\epsilon=0.01$ provided almost 7 times lower L2 error than the lower k-value results for the same gamma with also lower spiking and overall average neighbor count. Although the true efficacy of the spatial layer is in doubt, we present a methodology to mitigate the neighbor connectivity while still providing accurate reconstruction and low spiking, overall improving the integration of VS-GNO into neuromorphic/edge device hardware and its overall latency. Further research should delve into transient or generally varying graph structures instead of the single, stationary graph topology which might represent only a subset of virtual sensing applications.

As discussed throughout this work, SAR-based neural operators are not intended to replace conventional spiking neural operators, but rather to serve as a complementary low-latency sparsity-based alternative and a strong performance benchmark while improvements in spiking dynamics and optimization continue to emerge. Initial research should explore modifications to SAR that allow the expressiveness seen with the VSN. Currently, the ReLU function filters out negative inputs, and although a subsequent activation function $\sigma$ exists that can be important for sequential/transformer models, using tanh/softmax, the layer has less expression than the VSN which permits negative signals. Despite this limitation, SAR still performs similar or slightly outperforms VSN with the GeLU activation in VS-GNO, indicating that gradient mismatch poses a larger disadvantage than layer expressiveness. Moving beyond SAR, future research should also focus on overcoming the limitations of surrogate-gradient training and developing optimization techniques that preserve the inherent advantages of spiking computation. Ideally, neuron membrane integration, temporal memory, and dynamic spike generation should provide richer feature representations and more effective processing of transient physical phenomena. Spiking neural networks have consistently demonstrated strong performance on time-series problems, making it natural to investigate whether aligning neuron memory dynamics with the temporal behavior of transient physics can further improve virtual sensing performance. In particular, an important research question is whether membrane potentials should be reset after each physical timestep or propagated continuously throughout a transient simulation to better capture long-term temporal dependencies.

ANN-to-SNN conversion remains one of the primary surrogate-free training methodologies and has demonstrated considerable success for LIF-based neural networks. However, existing conversion techniques are fundamentally based on rate-coded communication and therefore do not naturally extend to Variable Spiking Neurons (VSNs), whose graded spike outputs encode continuous-valued information rather than firing rates and binary communication. The proposed SAR neural operator framework provides a alternative solution that is based on sparsity events not standard spiking dynamics but can still enable direct ANN-to-neuromorphic conversion for single-spike-timestep using zero thresholding while avoiding surrogate-gradient optimization and providing variable communication vital for performance on regression based spatial-temporal reconstruction. Nevertheless, this simplification sacrifices the temporal memory dynamics that make VSNs particularly attractive for multi-timestep inference. Consequently, extending surrogate-free optimization techniques such as SAR-NOMAD to incorporate membrane memory and temporal spiking dynamics (and exist not as a simple sparsity filter) represents a promising direction for future investigation. More broadly, this challenge motivates exploration of fundamentally different optimization paradigms that eliminate gradient mismatch altogether, including gradient-free optimization techniques, reinforcement learning, and emerging quantum computing approaches for large-scale search-based training \cite{aenugu2020trainingspikingneuralnetworks, CHEN2022435}.

Beyond training methodology, our entropy-based analysis of sparse neural representations and VSN implementations suggests additional opportunities for improving overall event-based model efficiency and expressiveness. In particular, the branch layers of the Hoyer-regularized SAR-NOMAD model exhibited entropy values approaching zero in several intermediate layers, indicating that the learned representations collapsed onto only one or a few dominant feature dimensions. While this behavior promotes sparsity, it also suggests that the available feature space is not being fully utilized. These observations motivate two complementary research directions. The first is model pruning \cite{cheng2024surveydeepneuralnetwork}, where inactive neurons, unused synapses connections, or even entire model components can be removed to reduce both parameter count and memory requirements. Such compression is especially important for deployment on highly resource-constrained edge platforms, such as neuromorphic devices (Loihi 2 \cite{9605018}), where memory capacity often represents a more significant limitation than computational throughput. The second direction involves developing new regularization strategies that explicitly consider feature entropy in addition to activation sparsity. Encouraging a more balanced utilization of the latent feature space while maintaining equivalent spiking activity could improve reconstruction accuracy by preventing feature collapse and increasing the representational capacity of intermediate neural embeddings.

In parallel with algorithmic advances, continued work toward realistic hardware deployment remains essential. Implementing SAR-based neural operators on edge device hardware that exploits the sparsity, which we have discussed could include neuromorphic chips, will not only demonstrate and answer the question of the practical feasibility of neural operators for energy-efficient virtual sensing but also quantify the latency, throughput, and energy-efficiency improvements predicted by the proposed framework. As discussed previously, the presented LEE score is limited in its ability to measure the true latency comparison between binary-based LIF layers and our presented SAR layer. Hardware deployment will additionally highlight the importance of the synthetic knowledge distillation strategy introduced in this work. As neural operators continue to increase in architectural complexity, direct deployment onto current edge device hardware becomes increasingly difficult due to memory, connectivity, and architectural constraints. Knowledge transfer from large, expressive teacher models to compact trunk-branch student architectures therefore represents a practical pathway for translating state-of-the-art neural operators into deployable edge systems.

Further development of the proposed synthetic distillation framework is also warranted. Although synthetic training data substantially improved reconstruction performance on the limited Heat Exchanger dataset, particularly for SAR-NOMAD, these gains were accompanied by a significant increase in training cost, especially when utilizing 8,000 synthetic samples. Future work should therefore investigate more efficient synthetic data generation strategies capable of achieving comparable performance improvements with significantly fewer generated examples. Likewise, performing knowledge distillation directly on the original training dataset rather than relying exclusively on synthetic samples may further reduce computational overhead while preserving the benefits of teacher-student learning. Finally, the current framework assumes that the input parameter distributions are known a priori, enabling straightforward sampling of synthetic training examples. Extending this methodology to more general functional inputs with unknown or highly complex distributions remains an important open problem, particularly for experimental measurements and real-world datasets where governing parameterizations are unavailable.

Overall, this work represents an important step toward practical, energy-efficient, real-time virtual sensing through the development of sparse neural operators. By introducing surrogate-free activation-sparsity regularization, analyzing sparse feature representations through entropy-based metrics, investigating improved optimization strategies for spiking neural operators, and proposing a synthetic knowledge distillation framework for reducing data requirements, this work advances several key challenges associated with deploying neural operators on low-power edge hardware. Collectively, these contributions help bridge the gap between increasingly sophisticated scientific machine learning architectures and practical hardware implementations capable of accurate, low-latency spatial-temporal physics reconstruction.

\section{Acknowledgments} This research used both the DeltaAI advanced computing and data resource, which is supported by the National Science Foundation (award OAC 2320345) and the State of Illinois, and the Delta advanced computing and data resource which is supported by the National Science Foundation (award OAC 2005572) and the State of Illinois. Delta and DeltaAI are joint efforts of the University of Illinois Urbana-Champaign and its National Center for Supercomputing Applications. This work was also made possible by support from the U.S. Department of Energy's Office of Nuclear Energy, specifically the University Nuclear Leadership Program's Graduate Fellowship. Lastly, LLMs were utilized solely for language modification and structuring.

\bibliographystyle{unsrt}  
\bibliography{references}
\end{document}